\documentclass[nohyperref]{article}

\usepackage{microtype}
\usepackage{graphicx}
\usepackage{subfigure}

\usepackage{booktabs} %
\usepackage{natbib}

\usepackage[colorlinks,citecolor=DarkGreen,linkcolor=FireBrick,urlcolor=FireBrick]{hyperref}

\usepackage{graphicx}
\usepackage{tikz}
\usepackage{tikz-cd}
\usepackage{times}
\usepackage{courier}
\usepackage{bm}
\usepackage{physics}
\usepackage{xcolor}
\usepackage{natbib}
\usepackage{mdframed}
\usepackage{multirow} 
\usepackage{nicefrac}
\usepackage{booktabs}
\usepackage{blindtext}
\usepackage{braket}
\usepackage{amsmath}
\usepackage{amssymb}
\usepackage{mathtools}
\usepackage{titletoc}
\usepackage{enumitem}
\usepackage{float}
\usepackage{colortbl}

\usepackage{graphicx}
\usepackage{tikz}
\usepackage{tikz-cd}

\usepackage{times}
\usepackage{courier}
\usepackage{bm}
\usepackage{physics}
\usepackage{natbib}
\usepackage{mdframed}
\usepackage{nicefrac}
\usepackage{booktabs}
\usepackage{lipsum}
\usepackage{titlesec}
\usepackage{wrapfig,lipsum,booktabs}
\usepackage{authblk}
\usepackage{blindtext}
\usepackage{titletoc}
\usepackage{amsmath}
\usepackage{adjustbox}
\usepackage{titletoc}

\usepackage{authblk}
\definecolor{DarkGreen}{rgb}{0,0.40,0}
\definecolor{FireBrick}{rgb}{0.698,0.133,0.133}

\usepackage{xcolor}
\definecolor{LightCyan}{rgb}{0.88,1,1}

\usepackage{array} %

\newcolumntype{G}{>{\columncolor{green}}c}
\newcolumntype{B}{>{\columncolor{blue}}c}
\definecolor{LightCyan}{rgb}{0.8, 0.9, 1}

\newcolumntype{g}{>{\columncolor{LightCyan}\hspace{0pt}}c}

\usepackage{dsfont} %
\newcommand{\one}{\mathds{1}}

\usepackage[nameinlink,capitalize,noabbrev]{cleveref}

\usepackage{CJK}

\usepackage{array}
\usepackage{makecell}

\def\half{{\frac{1}{2}}}

\newcommand{\bea}{\begin{eqnarray}}
\newcommand{\eea}{\end{eqnarray}}

\usepackage{slashed}

\def\({\left(}
\def\){\right)}
\def\[{\left[}
\def\]{\right]}

\usepackage{dashbox}
\usepackage{xcolor}
\usepackage{colortbl}
\usepackage{arydshln}
\definecolor{lightyellow}{rgb}{1.0, 0.95, 0.7}
\definecolor{blue}{rgb}{0.0, 0.4, 1.0}
\definecolor{Blue}{rgb}{0,0,1}
\definecolor{darkgreen}{rgb}{0,0.40,0}
\definecolor{firebrick}{rgb}{0.698,0.133,0.133}

\definecolor{colorA}{rgb}{1,0,0}
\definecolor{colorB}{rgb}{0,0.3,1}
\definecolor{colorC}{rgb}{0.9,0.8,0.2}
\definecolor{colorD}{rgb}{0,0.65,0}
\definecolor{lesslightgray}{rgb}{0.5,0.5,0.5}
\definecolor{light-gray}{gray}{0.95}

\let\tilde\widetilde
\let\hat\widehat

\newcommand{\calF}{\mathcal{F}}

\newcommand{\calH}{\mathcal{H}}

\newcommand{\calJ}{\mathcal{J}}

\newcommand{\calO}{\mathcal{O}}

\newcommand{\calS}{\mathcal{S}}
\newcommand{\calT}{\mathcal{T}}

\newcommand{\bA}{A}

\newcommand{\bK}{K}

\newcommand{\bP}{P}
\newcommand{\bQ}{Q}
\newcommand{\bR}{R}

\newcommand{\bV}{V}
\newcommand{\bW}{W}
\newcommand{\bX}{X}
\newcommand{\bY}{Y}
\newcommand{\bZ}{Z}
\newcommand{\ba}{a}

\newcommand{\br}{r}
\newcommand{\bv}{v}
\newcommand{\bq}{q}
\newcommand{\bx}{x}
\newcommand{\by}{y}
\newcommand{\bz}{z}
\newcommand{\bxi}{\xi}

\newcommand{\Max}{\mathop{\rm Max}}
\newcommand{\Min}{\mathop{\rm Min}}

\newcommand{\Softmax}{{\rm{Softmax}}}

\newcommand{\lse}{{\rm{lse}}}

\newcommand{\sT}{ \mathsf{T} }

\def\R{\mathbb{R}}

\let\cite\citep 

\usepackage{amsthm}
\makeatletter
\def\th@remark{%
  \thm@headfont{\bfseries}%
  \normalfont %
  \thm@preskip\topsep \divide\thm@preskip\tw@
  \thm@postskip\thm@preskip
}
\makeatother
\usepackage[many]{tcolorbox}
\theoremstyle{definition}
\newtheorem{theorem}{Theorem}[section]
\tcolorboxenvironment{theorem}{
  breakable,
  colback=black!10,
  colframe=white,%
  width=\dimexpr\linewidth+10pt\relax,%
  enlarge left by=-5pt,%
  enlarge right by=-5pt,%
  boxsep=5pt,%
  boxrule=0pt,
  left=0pt,right=0pt,top=0pt,bottom=0pt,
  sharp corners,
  before skip=\topsep,
  after skip=\topsep
}
\newtheorem{lemma}{Lemma}[section]
\tcolorboxenvironment{lemma}{
  breakable,
  colback=black!10,
  colframe=white,%
  width=\dimexpr\linewidth+10pt\relax,%
  enlarge left by=-5pt,%
  enlarge right by=-5pt,%
  boxsep=5pt,%
  boxrule=0pt,
  left=0pt,right=0pt,top=0pt,bottom=0pt,
  sharp corners,
  before skip=\topsep,
  after skip=\topsep
}
\newtheorem{corollary}{Corollary}[theorem]
\tcolorboxenvironment{corollary}{
  breakable,
  colback=black!10,
  colframe=white,%
  width=\dimexpr\linewidth+10pt\relax,%
  enlarge left by=-5pt,%
  enlarge right by=-5pt,%
  boxsep=5pt,%
  boxrule=0pt,
  left=0pt,right=0pt,top=0pt,bottom=0pt,
  sharp corners,
  before skip=\topsep,
  after skip=\topsep
}

\theoremstyle{definition}
\newtheorem{definition}{Definition}[section]
\tcolorboxenvironment{definition}{
  breakable,
  colback=black!10,
  colframe=white,%
  width=\dimexpr\linewidth+10pt\relax,%
  enlarge left by=-5pt,%
  enlarge right by=-5pt,%
  boxsep=5pt,%
  boxrule=0pt,
  left=0pt,right=0pt,top=0pt,bottom=0pt,
  sharp corners,
  before skip=\topsep,
  after skip=\topsep
}
\theoremstyle{remark}
\newtheorem{remark}{Remark}[section]
\newtheorem{assumption}{Assumption}[section]
\tcolorboxenvironment{assumption}{
  breakable,
  colback=black!10,
  colframe=white,%
  width=\dimexpr\linewidth+10pt\relax,%
  enlarge left by=-5pt,%
  enlarge right by=-5pt,%
  boxsep=5pt,%
  boxrule=0pt,
  left=0pt,right=0pt,top=0pt,bottom=0pt,
  sharp corners,
  before skip=\topsep,
  after skip=\topsep
}

\tcolorboxenvironment{problem}{
  breakable,
  colback=black!10,
  colframe=white,%
  width=\dimexpr\linewidth+10pt\relax,%
  enlarge left by=-5pt,%
  enlarge right by=-5pt,%
  boxsep=5pt,%
  boxrule=0pt,
  left=0pt,right=0pt,top=0pt,bottom=0pt,
  sharp corners,
  before skip=\topsep,
  after skip=\topsep
}

\crefname{theorem}{Theorem}{Theorems}
\crefname{proposition}{Proposition}{Propositions}
\crefname{lemma}{Lemma}{Lemmas}
\crefname{corollary}{Corollary}{Corollaries}
\crefname{definition}{Definition}{Definitions}
\crefname{assumption}{Assumption}{Assumptions}
\crefname{remark}{Remark}{Remarks}
\crefname{problem}{Problem}{Problems}
\crefname{property}{Property}{property}

\numberwithin{equation}{section}
\numberwithin{theorem}{section}
\numberwithin{proposition}{section}
\numberwithin{definition}{section}
\numberwithin{lemma}{section}
\numberwithin{assumption}{section}
\numberwithin{remark}{section}

\usepackage{lipsum}

\newcommand*{\annot}[1]{\tag*{\footnotesize{\textcolor{black!50}{\big(#1\big)}}}}

\makeatletter
\let\save@mathaccent\mathaccent
\newcommand*\if@single[3]{%
    \setbox0\hbox{${\mathaccent"0362{#1}}^H$}%
    \setbox2\hbox{${\mathaccent"0362{\kern0pt#1}}^H$}%
    \ifdim\ht0=\ht2 #3\else #2\fi
}
\newcommand*\rel@kern[1]{\kern#1\dimexpr\macc@kerna}
\newcommand*\widebar[1]{\@ifnextchar^{{\wide@bar{#1}{0}}}{\wide@bar{#1}{1}}}
\newcommand*\wide@bar[2]{\if@single{#1}{\wide@bar@{#1}{#2}{1}}{\wide@bar@{#1}{#2}{2}}}
\newcommand*\wide@bar@[3]{%
    \begingroup
    \def\mathaccent##1##2{%
        \let\mathaccent\save@mathaccent
        \if#32 \let\macc@nucleus\first@char \fi
        \setbox\z@\hbox{$\macc@style{\macc@nucleus}_{}$}%
        \setbox\tw@\hbox{$\macc@style{\macc@nucleus}{}_{}$}%
        \dimen@\wd\tw@
        \advance\dimen@-\wd\z@
        \divide\dimen@ 3
        \@tempdima\wd\tw@
        \advance\@tempdima-\scriptspace
        \divide\@tempdima 10
        \advance\dimen@-\@tempdima
        \ifdim\dimen@>\z@ \dimen@0pt\fi
        \rel@kern{0.6}\kern-\dimen@
        \if#31
        \overline{\rel@kern{-0.6}\kern\dimen@\macc@nucleus\rel@kern{0.4}\kern\dimen@}%
        \advance\dimen@0.4\dimexpr\macc@kerna
        \let\final@kern#2%
        \ifdim\dimen@<\z@ \let\final@kern1\fi
        \if\final@kern1 \kern-\dimen@\fi
        \else
        \overline{\rel@kern{-0.6}\kern\dimen@#1}%
        \fi
    }%
    \macc@depth\@ne
    \let\math@bgroup\@empty \let\math@egroup\macc@set@skewchar
    \mathsurround\z@ \frozen@everymath{\mathgroup\macc@group\relax}%
    \macc@set@skewchar\relax
    \let\mathaccentV\macc@nested@a
    \if#31
    \macc@nested@a\relax111{#1}%
    \else
    \def\gobble@till@marker##1\endmarker{}%
    \futurelet\first@char\gobble@till@marker#1\endmarker
    \ifcat\noexpand\first@char A\else
    \def\first@char{}%
    \fi
    \macc@nested@a\relax111{\first@char}%
    \fi
    \endgroup
    }
\makeatother
\let\bar\widebar

\makeatletter
\newcommand*{\redefinesymbolwitharg}[1]{%
  \expandafter\let\csname ltx#1\expandafter\endcsname\csname #1\endcsname
  \@namedef{#1}{\@ifnextchar{^}{\@nameuse{#1@}}{\@nameuse{#1@}^{}}}%
  \expandafter\def\csname #1@\endcsname^##1##2{%
     \csname ltx#1\endcsname\ifx!##1!\else^{##1}\fi\mathopen{}\mathclose\bgroup\left(##2\aftergroup\egroup\right)
     }%
}
\makeatother
\redefinesymbolwitharg{sin}
\redefinesymbolwitharg{cos}

\usepackage{url}
\definecolor{LightCyan}{rgb}{0.8, 0.9, 1}

\newcolumntype{b}{>{\columncolor{LightCyan}\hspace{0pt}}c}

\newcommand*\rot{\rotatebox{90}}

\usepackage{amsmath}
\usepackage{amssymb}
\usepackage{mathtools}
\usepackage{amsthm}

\usepackage[accepted]{icml2024}

\usepackage[textsize=tiny]{todonotes}

\icmltitlerunning{Outlier Efficient Modern Hopfield Model for Large Transformer-Based Models}

\begin{document}

\twocolumn[
\icmltitle{Outlier-Efficient Hopfield Layers for Large Transformer-Based Models
}

\icmlsetsymbol{equal}{*}

\begin{icmlauthorlist}
\icmlauthor{Jerry Yao-Chieh Hu}{equal,nucs}
\icmlauthor{Pei-Hsuan Chang}{equal,yyy}
\icmlauthor{Haozheng Luo}{equal,nucs}
\icmlauthor{Hong-Yu Chen}{yyy}
\icmlauthor{Weijian Li}{nucs}
\icmlauthor{Wei-Po Wang}{yyy}
\icmlauthor{Han Liu}{nucs,nustats}

\end{icmlauthorlist}

\icmlaffiliation{yyy}{Department of Physics, National Taiwan University, Taipei, Taiwan}
\icmlaffiliation{nucs}{Department of Computer Science, Northwestern University, Evanston, USA}
\icmlaffiliation{nustats}{Department of Statistics and Data Science, Northwestern University, Evanston, USA}

\icmlcorrespondingauthor{Jerry Yao-Chieh Hu}{\href{mailto:jhu@u.northwestern.edu}{jhu@u.northwestern.edu}}
\icmlcorrespondingauthor{Pei-Hsuan Chang}{\href{mailto:b09202022@ntu.edu.tw}{b09202022@ntu.edu.tw}}
\icmlcorrespondingauthor{Haozheng Luo}{\href{mailto:robinluo2022@u.northwestern.edu}{robinluo2022@u.northwestern.edu}}
\icmlcorrespondingauthor{Hong-Yu Chen}{\href{mailto:b0976960890@gmail.com}{b0976960890@gmail.com}}
\icmlcorrespondingauthor{Weijian Li}{\href{mailto:weijianli@u.northwestern.edu}{weijianli@u.northwestern.edu}}
\icmlcorrespondingauthor{Wei-Po Wang}{\href{mailto:b09202009@ntu.edu.tw}{b09202009@ntu.edu.tw}}
\icmlcorrespondingauthor{Han Liu}{\href{mailto:hanliu@northwestern.edu}{hanliu@northwestern.edu}}

\icmlkeywords{Machine Learning, ICML}

\vskip 0.3in
]

\printAffiliationsAndNotice{\icmlEqualContribution} %

\setlength{\abovedisplayskip}{4pt}
\setlength{\abovedisplayshortskip}{4pt}
\setlength{\belowdisplayskip}{4pt}
\setlength{\belowdisplayshortskip}{4pt}

\titlespacing*{\section}{0pt}{0pt}{0pt}
\titlespacing*{\subsection}{0pt}{0pt}{0pt}
\titlespacing*{\subsubsection}{0pt}{0pt}{0pt}

\setlist[itemize]{leftmargin=1em, before=\vspace{-0.5em}, after=\vspace{-0.5em}, itemsep=0.1em}
\setlist[enumerate]{leftmargin=1.2em, before=\vspace{-0.5em}, after=\vspace{-0.5em}, itemsep=0.1em}

\begin{abstract}
We introduce an Outlier-Efficient Modern Hopfield Model (termed $\mathtt{OutEffHop}$) and use it to address the outlier inefficiency problem of {training} gigantic transformer-based models. 
Our main contribution is a novel associative memory model facilitating  \textit{outlier-efficient} associative memory retrievals.
Interestingly, this memory model manifests a model-based interpretation of an outlier-efficient attention mechanism ($\Softmax_1$): it is an approximation
of the memory retrieval process of  $\mathtt{OutEffHop}$.
Methodologically, this allows us to introduce novel outlier-efficient Hopfield layers as powerful alternatives to traditional attention mechanisms, with superior post-quantization performance. 
Theoretically, the Outlier-Efficient Modern Hopfield Model retains and improves the desirable properties of standard modern Hopfield models, including fixed point convergence and exponential storage capacity.
Empirically, we demonstrate the efficacy of the proposed model across large-scale transformer-based and Hopfield-based models (including BERT, OPT, ViT, and STanHop-Net), benchmarking against state-of-the-art methods like $\mathtt{Clipped\_Softmax}$ and $\mathtt{Gated\_Attention}$. 
Notably, $\mathtt{OutEffHop}$ achieves an average reduction of 22+\% in average kurtosis and 26+\% in the maximum infinity norm of model outputs across four models.
Code is available at \href{https://github.com/MAGICS-LAB/OutEffHop}{GitHub};
future updates are on \href{https://arxiv.org/abs/2404.03828}{arXiv}.

\end{abstract}

\section{Introduction}
\label{sec:intro}

We address the outlier-inefficient problem in large Transformer-based models by debuting a novel outlier-efficient modern Hopfield model.
This problem is  of practical importance in the {era} of Large Foundation Models \cite{bommasani2021opportunities}, i.e.,  huge transformer-based models, pretrained on massive datasets.
They play a central role not only in
machine learning but also in a wide range of scientific domains, such as ChatGPT \cite{brown2020language,floridi2020gpt} for natural language, BloombergGPT \cite{wu2023bloomberggpt} for finance, DNABERT \cite{zhou2024dnabert,zhou2023dnabert,ji2021dnabert} for genomics, and many others. 
Specifically, the problem of outlier inefficiency in these large models stems from their tendency to allocate attention to less informative tokens (the ``no-op'' outliers), including delimiters and punctuation marks. 
This tendency arises because these large models assign non-zero attention probabilities to low-information tokens, diluting the overall effectiveness of the attention mechanism \citep[Section~3]{bondarenko2023quantizable}.
As training progresses, the influence of these ``no-op'' outliers magnifies due to the softmax function's inability to assign zero probability.
Consequently, it leads to a scenario where even irrelevant tokens contribute to the model's outputs. Besides, it makes the model need unnecessarily large GPU memory space to host due to the extra bits that outliers take.
This hampers the model's processing efficiency and potential accuracy.

\begin{figure}[H]
\vspace{-1em}\includegraphics[width=\linewidth]{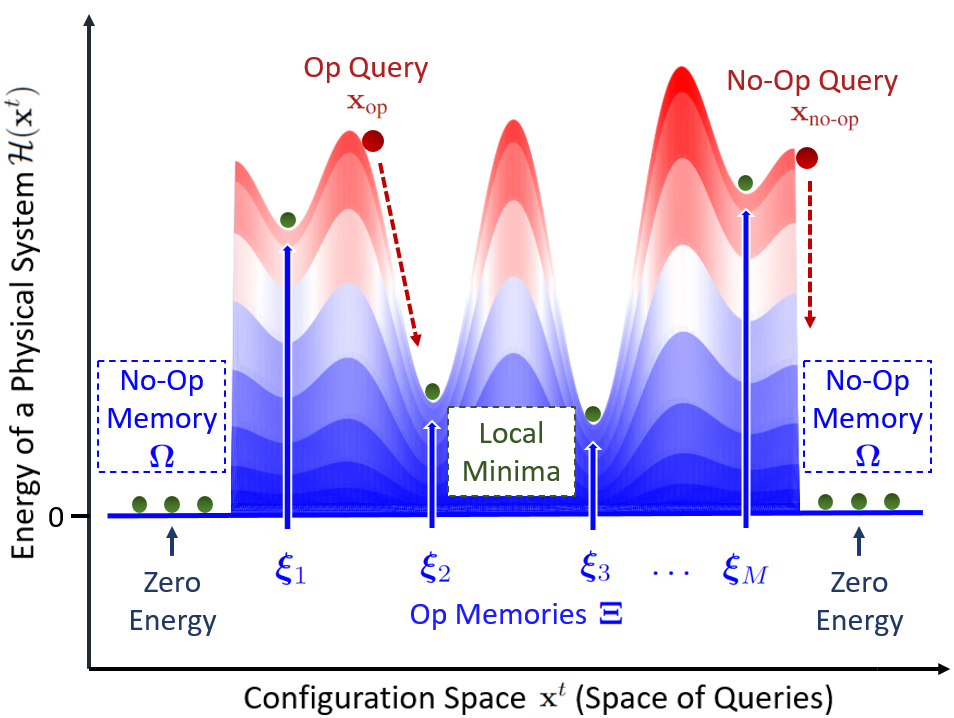}
    \vspace{-2em}
    \caption{Visualization of Outlier-Efficient Hopfield Model.
    }
    \label{fig:energy-land}
    \vspace{-1em}
\end{figure}

To combat this, 
we take a route from the deep learning compatible modern Hopfield models \cite{wu2024uniform,wu2023stanhop,hu2024nonparametric,hu2024computational,hu2023SparseHopfield,ramsauer2020hopfield}.
Through the associative memory model interpretation of transformer attention,
we introduce a novel outlier-efficient modern Hopfield model. 
This model's memory retrieval dynamics approximate an outlier-efficient attention mechanism ($\Softmax_1$) \cite{miller2021}. 
This allows us to debut novel outlier-efficient Hopfield layers as outlier-efficient alternatives for vanilla attention \cite{vaswani2017attention}.
The fundamental idea of our model is to add one extra ``no-op classification'' dimension into state/configuration space of the Hopfield energy function.
This dimension classifies whether a stored memory pattern is a ``no-op'' outlier, see \cref{fig:energy-land} for a visualization.
We regard the “no-op” outliers as distinct or rare patterns with no similarity to other memory patterns.
Then,
we present
an outlier-efficient Hopfield energy function with a refined log-sum-exponential function.
Consequently, 
this energy-based associative memory model allocates this ``no-op'' pattern to the zero-energy point of the energy function, remaining unaffected by state updates (retrievals).
Remarkably, 
by the standard CCCP derivation for modern Hopfield models, 
this new energy function leads to
a memory-retrieval dynamics that not only retrieves stored memories in an outlier-efficient fashion but also subsumes the $\Softmax_1$ attention \cite{miller2021} as its special case (when limited to a single update).    

\paragraph{Contributions.}
We propose the Outlier-Efficient Modern Hopfield Model. 
Our contributions are as follows:
\begin{itemize}
    \item 
    We propose an associative memory model capable of outlier-efficient memory retrievals with strong physics intuition.
    Theoretically, 
    we analyze the proposed model
    equips the standard properties of modern Hopfield models:
    fixed point convergence (\cref{lemma:convergence_sparse}) and exponential memory capacity (\cref{lemma:capacity}).
    Importantly, we derive an outlier-efficient Hopfield layer $\mathtt{OutEffHop}$ as a promising attention alternative (\cref{sec:DL}).
    Moreover, 
    we provide a model-based interpretation for the $\Softmax_1$ attention \cite{miller2021}:
    it is an approximation of the memory retrieval dynamics of the outlier-efficient modern Hopfield model (\cref{lem:retrieval_dyn}).
    
    \item 
    Methodologically, we introduce outlier-efficient Hopfield layers as new components in deep learning. 
    These layers tackle the outlier problem of large models by reducing the probability assigned to low-information vectors. 
    In addition to outlier reduction, we explore the generalization of $\mathtt{OutEffHop}$. 
    We establish a generalization bound (\cref{lemma:gen_OutEffHop}) that scales with $N^{-1/2}\log N$ in sample size and $\log(dM)$ in the pattern dimension $d$ and the size of the stored memory set $M$. 
    This positions $\mathtt{OutEffHop}$ as a promising alternative to transformer attention.

    \item 
    Empirically, we validate the proposed method on 3 common large transformer-based and 1 Hopfield-based models (BERT \cite{devlin-etal-2019-bert}, Open Pre-trained Transformer (OPT) \cite{zhang2022opt}, Vision Transformer (ViT) \cite{dosovitskiy2020image} and STanHop-Net \cite{wu2023stanhop}). 
    {Specifically, $\mathtt{OutEffHop}$ reduces average kurtosis and maximum infinity norm by $\sim$22+\% and $\sim$26+\%, respectively\footnote{See \cref{tab:result1} for details and \href{https://huggingface.co/collections/magicslabnu/outeffhop-6610fcede8d2cda23009a98f}{Hugging Face Hub} for models.}} and improves the same metrics by an average of 3\% and 4\% compared to 3 variants of STanHop-Net and ranks among the top two in outlier efficiency in 25 out of 30 settings.

\end{itemize}

\section{Outlier-Efficient Hopfield Model}
\label{sec:method}
This section introduces the Outlier-Efficient Modern Hopfield Model.
\cref{sec:classification_mech} presents an internal “no-op classification” mechanism for all memory patterns.
Then, \cref{sec:outlier_energy} utilizes this mechanism to 
construct a model  facilitating outlier-efficient associative memory retrievals.
Importantly, the retrieval dynamics of this model subsumes an outlier-efficient attention as its special case, and  \cref{sec:DL} debuts outlier-efficient Hopfield layers for deep learning.

\subsection{Background}\label{background}
This section presents the ideas we build on.

\paragraph{``No-Op" Outliers in Attention Heads.}
\citet{clark-etal-2019-bert, kovaleva2019revealing} identify specific tokens in BERT, such as delimiters and punctuation mark, receive larger attention weights.
Furthermore, \citet{kobayashi2020attention} reveal that tokens with small value vectors tend to receive significantly large attention weights.
As stated in \cite{bondarenko2023quantizable}, low-information tokens within BERT and background patches in the Vision Transformer (ViT) attract large attention probability to achieve no-update.

To see this, 
we consider an input sequence $\bX=[\bx_1,\ldots,\bx_L]\in\R^{d\times L}$ and 
 the attention mechanism 
\begin{align*}
    {\rm{Attention}}(\bX) = \Softmax{\(\bQ\bK^{\sT}\)}\bV = \bA. 
\end{align*}
We focus on the part of transformer right after attention
\begin{align}
\label{eqn:output}
    {\rm{Output}} = {\rm{Residual}}(\bX + \bA ). 
\end{align}
If the input $\bX$ already has enough information and does not require further feature extraction,
the attention mechanism tends to behave like an identity map, 
and output a zero $\bA$.
This is known as the \textit{no-update situation}: 
the output of \eqref{eqn:output} is the same as input $\bX$.
A direct consequence of this is that --- 
the attention mechanism forces
tokens with large values (as in $\bV$) receive \textit{close-to-zero} attention probability (as in $\Softmax{\(\bQ \bK^{\sT}\)}$), resulting small-value tokens to have large attention probability.
By the normalization nature of softmax function, 
this operation forces its input $\bQ  \bK^{\sT}$ to have a wide range.
This is the fundamental source of outliers: 
there must be some tokens causing the ``wide range" of $\bQ  \bK^{\sT}$, namely \textit{outliers}. 
Since attention to these tokens behaves as a ``no-op", as mentioned in \cite{clark-etal-2019-bert}, we term these outliers as ``no-op" outliers.
Furthermore, 
since the softmax function never reaches exact zero, 
it always sends back a gradient signal, leading to the magnification of outliers during training \cite{bondarenko2023quantizable}.

\paragraph{Modern Hopfield Models.}
Let $\bx \in \mathbb{R}^{d}$ represent the query patterns and $\Xi = [\xi_1, \cdots, \xi_M ] \in \mathbb{R}^{d\times M}$ the memory patterns.
  \citet{krotov2016dense} introduce the dense associative memory model  encoding memory patterns $\Xi$ into energy function $\calH(\bx)$ using \textit{overlap-construction}: $\calH(\bx) = F(\Xi^{\sT} \bx)$, where $F: \mathbb{R}^M \rightarrow \mathbb{R}$ is a smooth function.
 The choice of energy function and the corresponding retrieval dynamics results in different Hopfield models types \cite{krotov2016dense,krotov2020large,demircigil2017model,ramsauer2020hopfield,hu2023SparseHopfield,hu2024nonparametric,wu2024uniform,wu2023stanhop}. 
Inspired by the dense associative memory models,  
\citet{ramsauer2020hopfield} introduce the modern Hopfield models  with the energy function of the form 
\begin{align*}
\calH(\bx) = -\text{lse}\(\beta, \Xi^{\sT} \bx\) + \half\Braket{\bx,\bx} + \text{Const.},
\end{align*}
where $\text{lse}(\beta, \bz) \coloneqq \beta^{-1} \log\sum^M_{\mu=1} \exp{\beta z_{\mu}}$.
In addition, they introduce the corresponding retrieval dynamics as 
\begin{align}
\label{eqn:origianl_retrieval_dyn}
\bx_{\text{new}}\gets \calT(\bx) = \Xi \Softmax(\beta \Xi^{\sT}\bx),
\end{align}
for any input query $\bx\in\R^d$.
The modern Hopfield model possesses several desirable properties, including:
\begin{enumerate}[before=\vspace{-.5em}, after=\vspace{-0.5em}, itemsep=-0.2em]
    \item \textbf{Exponential Memory Capacity:}  
    Achieved by highly non-linear energy functions.
    \item \textbf{One-step Retrieval Dynamics:}
    Achieved by guaranteeing monotonic energy function minimization.
    \item \textbf{Compatibility with Deep Learning Architectures:}
    Achieved by the link between their retrieval dynamics and attention mechanisms.
\end{enumerate}

\subsection{One Dimension More}
\label{sec:classification_mech}
As models of associative memory, modern Hopfield models aim to retrieve a memory pattern $\bx_{\text{new}}$ from the stored memories $\Xi$, closest to the input query $\bx$.
By \eqref{eqn:origianl_retrieval_dyn}, 
they do this by computing the output $\bx_{\text{new}}$ as the \textit{expectation value} of $\Xi$ over the distribution $\Softmax{(\Xi^{\sT} \bx)}$. 
Crucially, the weight of $\Softmax{(\Xi^{\sT} \bx)}$, i.e., $\Xi^{\sT} \bx$, represents the inner-product similarity measure between the input query $\bx$ and each stored memory $\bxi_{\mu}$. 
Namely, the greater $\Braket{\xi_\mu, \bx}$ is, the stronger their correlation.

Under this interpretation, 
for a given query $\bx$,
the memory patterns with low similarity inevitably deviate the expectation value from the ground truth.
This occurs because the softmax function always assigns non-zero probability weights, even for near zero similarity $\Braket{\xi_\mu, \bx}\simeq 0$.
Consequently, this results to more iterative retrievals for the retrieval dynamics to converge to the ground truth memory (w.r.t $\bx$).
We refer to these low-similarity memory patterns as ``\textbf{no}-\textbf{op} patterns," as they are unrelated to the presented query and should \textbf{no}t \textbf{op}erate during the retrieval process.

Motivated by above, 
we introduce a new dimension into the pattern vectors to distinguish ``no-op patterns" from the relevant ones, via the following ``no-op classification.''

\paragraph{No-Op Classification Mechanism.}
Given an input query pattern $\bx=(x_1,\ldots,x_d)$ and memory patterns $\xi^\mu = (\xi_1^\mu, \cdots , \xi_d^\mu)$ with $\mu\in[M]$. 
We extend their dimension such that 
\begin{align*}
\bar{\bx}=(x_1,\ldots,x_d,0),\quad
\bar{\xi}^\mu = (\xi_1^\mu, \cdots , \xi_d^\mu,\omega),
\end{align*}
with an extra $\omega\in \R$.
In addition, for memory patterns, we set this extra dimension $\omega$ to be 
\begin{itemize}[before=\vspace{-1.2em}, after=\vspace{-1.2em}, itemsep=-0.5em]
    \item $\omega \neq 0$: non-zero for no-op outliers, and
    \item $\omega = 0$: zero for the rest memory patterns,
\end{itemize}
assuming we are aware of which patterns are outliers\footnote{We can do this by  either ad-hoc assignment or similarity measure thresholding (See \ref{sec:softmax1_nb} for details).}.
Then we introduce the following function: 
\begin{align}
\label{eqn:out_class}
\Lambda(\bar{\xi}_\mu) = \left\{
    \begin{aligned}
         &(\xi_1^\mu, \cdots, \xi_d^\mu, 0) = \bar{\xi}^{\mu}_{\text{op}}\in\R^{d+1},&\ \text{if} \  \omega=0, \\
        &(\underbrace{0, \cdots, 0}_{d}, C) = \Omega \in\R^{d+1}, &\ \text{if} \ \omega\neq 0
    \end{aligned}
    \right.
,\end{align}
with some $C\in\R $ and for all $\mu\in[M]$, to map all ``no-op patterns" into an unique ``\textit{no-op memory class vector} $\Omega$."
By design, the inner product of the vector $\Omega$ with the query $\bar{\bx}$ is zero: $\Braket{\Omega, \bar{\bx}} = 0$.
We term the $\Lambda$ function \eqref{eqn:out_class} the ``no-op classification mechanism.''
It enforces all outlier memory patterns to have zero inner product with the input query.
\begin{remark}
\label{remark:moreC}
It is also feasible to design $\Lambda$ so that each outlier pattern maps to a distinct, non-repeated $C$.
However, the form presented here offers better elegance and simplicity.
\end{remark}

In sum, for any set of ($d$-dimensional patterns) $\bx$ and $\Xi=[\bxi_1,\ldots,\bxi_M]$, 
we obtain a set of  ($(d+1)$-dimensional patterns) $\bar{\bx}$ and $\bar{\Xi}=[\bar{\bxi}_1,\ldots,\bar{\bxi}_M]$.
Suppose there are $K$ outliers in $ \bar{\Xi}$. 
Then, with $\Lambda$, we further categorize $\bar{\Xi}$ into $(M-K) $ $\{\bar{\xi}^{\mu}_{\text{op}}\}_{\mu\in[M-K]}$ and a single $\Omega$.

\subsection{Hopfield Energy and Retrieval Dynamics}
\label{sec:outlier_energy}
Now, we utilize above to construct the Outlier-Efficient Modern Hopfield Model.
For the ease of presentation, in the following,
we set 
\begin{align*}
    \bx\gets \bar{\bx},\quad
    \bxi_\mu\gets \bar{\bxi}_\mu\quad(\text{i.e., }\Xi\gets \bar{\Xi}),
\end{align*}
for query and memory patterns.
Move rover, since we only need a single $\Omega$ for outliers, we set
\begin{align*}
    d\gets (d+1),\quad M\gets (M-K),
\end{align*}
for pattern dimension and the number of ``op'' memory patterns.
We introduce the outlier-efficient Modern Hopfield energy as:
\begin{align}
\label{eqn:H_energy}
\calH(\bx)= -\lse_1\(\beta, \Xi^\sT \bx\)+\half\Braket{\bx,\bx} + \text{Const.},
\end{align}
where $\lse_1$ is a refined log-sum-exponential fucntion:
\begin{align}
\label{eqn:lse1}
    &~
    \lse_1\(\beta, \Xi^\sT \bx\) \nonumber\\ 
    \coloneqq   &~
    \beta^{-1}\log \(\sum^{M}_{\mu=1} \exp{\beta \Braket{\xi_\mu, \bx}} + \exp{\beta \Braket{\Omega, \bx}} \)\nonumber\\
    = &~ \beta^{-1}\log \(\sum^{M}_{\mu=1}\exp{\beta \Braket{\xi_\mu, \bx}} + 1 \).
\end{align}

\remark{
Since the $\log$ function is monotonic,
\eqref{eqn:lse1} has a physical interpretation of an energy function with a 
\textit{zero-energy point}\footnote{This zero-energy point does not mean $\calH = 0$.} associated with $\exp{\beta \Braket{\Omega, \bx}}=\exp{0}$.
Naturally, this zero-energy point serves as one of the local minima of $\calH$, and thus corresponds to a memory pattern.
More precisely,
we retrieve ``no-op'' memory from there 
with proper retrieval dynamics $\calT$.

\begin{remark}\label{remark:softmax_K}
Echoing with \cref{remark:moreC},
if we twist \eqref{eqn:out_class} to have one $\Omega_C$ for each no-op patterns with non-repeated $C$'s,
then \eqref{eqn:lse1} simply becomes 
$\text{lse}_K\(\beta, \Xi^\sT \bx\)\coloneqq \beta^{-1}\log \(\sum^{M}_{\mu=1}\exp{\beta \Braket{\xi_\mu, \bx}} + K \).$

\end{remark}

\paragraph{Retrieval Dynamics.}
With \eqref{eqn:H_energy}, we derive the following memory retrieval dynamics:
\begin{lemma}[Retrieval Dynamics]
\label{lem:retrieval_dyn}
Let $\Softmax_1(\bz) \coloneqq \exp{\bz}/\({\sum_{\mu=1}^M \exp{z_\mu}+1}\)$ for any $\bz\in\R^M$ and $t$ be the iteration number.
The memory retrieval dynamics:
\begin{align}
\label{eqn:retrieval_dyn}
\calT_{\text{OutEff}}\(\bx_t\)\coloneqq\Xi \Softmax_1\(\beta\Xi^\sT \bx_t\)=\bx_{t+1},
\end{align}
monotonically minimizes the energy \eqref{eqn:H_energy}  over $t$.
\end{lemma}
\vspace{-.5em}
\begin{proof}[Proof Sketch]
Since \eqref{eqn:lse1} is concave by design,
we prove this by standard CCCP derivation following \cite{hu2023SparseHopfield}. 
See \cref{apdix: proof_retrieval_dyn} for a detailed proof.
\end{proof}\vspace{-1em}
Due to the monotonic decreasing property of \cref{lem:retrieval_dyn},
for any given input query $\bx$,
\eqref{eqn:retrieval_dyn} retrieves a memory closest to it by approaching to the nearest local minimum of $\calH$.
Interesting, when $\calT_{\text{OutEff}}$ is applied only once, \eqref{eqn:retrieval_dyn} is equivalent to an outlier-efficient attention \cite{miller2021}.
\begin{remark}
Importantly, this set of $\calH$ and $\calT$ enables outlier-efficient associative memory retrievals. 
When we identify specific memory patterns as outliers relative to the input query and classify them into $\Omega$, they no longer contribute to the retrieval output defined by \eqref{eqn:retrieval_dyn}.
\end{remark}

\subsection{Connection to Deep Learning}
\label{sec:DL}
Outlier-efficient Hopfield model is applicable to nowadays deep learning architectures, due to its connection to transformer attention mechanism when the retrieval dynamics $\calT_{\text{OutEff}}$ undergoes a single iteration. 
Consider the raw query ${\bR}$ and memory pattern $\bY$.
We define the \textit{query} and \textit{memory} associative (or embedded) spaces through transformations: 
$\bX^\sT={\bR}\bW_Q\coloneqq \bQ \quad \text{and} \quad \Xi^\sT=\bY \bW_K \coloneqq \bK$,
with matrices $\bW_Q$ and $\bW_K$.
By transposing the retrieval dynamics \eqref{eqn:retrieval_dyn} and multiplying with $\bW_V$ (letting $\bV\coloneqq \bK\bW_V$), we get:
    $\bQ^{\text{new}} \bW_V = \Softmax_1(\beta  \bQ\bK^\sT)\bV$.

This equation resembles the attention mechanism but with a $\Softmax_1$ activation. 
When substituting the original patterns $\bR$ and $\bY$, we present the Outlier-Efficient Hopfield ($\mathtt{OutEffHop}$) layer:
\begin{align}
\label{eqn:OutEffHop}
\bZ&=\mathtt{OutEffHop}\(\bR,\bY\) \nonumber\\
&= \Softmax_1\(\beta  {\bR}\bW_Q \bW_K^\sT\bY^\sT\)\bY\bW_K \bW_V.
\end{align}
This layer is readily incorporated into deep learning models.
To elaborate, the $\mathtt{OutEffHop}$ layer takes $\bR$ and $\bY$ as input, paired with weight matrices $\bW_Q$, $\bW_K$, and $\bW_V$. 
Similar to \cite{hu2024nonparametric,wu2023stanhop,hu2023SparseHopfield,ramsauer2020hopfield}, its configuration determines its behavior:
\vspace{-1.8em}
\begin{itemize}
    \item \textbf{Memory Retrieval:} This mode does not require learning.
        The matrices $\bW_K$, $\bW_Q$, and $\bW_V$ are identity matrices. $\bR$ acts as the query to retrieve memory patterns $\bY$.    

    \item \textbf{\texttt{OutEffHop}:}
    In this design, $\bR$ and $\bY$ are inputs. 
    The matrices $\bW_K$, $\bW_Q$, and $\bW_V$ are adjustable, offering an alternative to the usual attention mechanism with outlier efficiency.
    $\bR$, $\bY$, and $\bY$ function as the sources of query, key, and value, respectively. 
    To mimic a self-attention mechanism, we set $\bR$ equal to $\bY$.

    \item \textbf{\texttt{OutEffHopPooling}:} 
    Here, $\bY$ is the only input of the layer. $\bQ$ acts as a learnable query that can search static prototype patterns in $\bY$.
    We consider this layer as a pooling layer if only one static state pattern (query) exists.

    \item \textbf{\texttt{OutEffHopLayer}:} With just $\bR$ as input (which denotes the query pattern), the adaptive matrices $\bW_K$ and $\bW_V$ act as repositories for stored patterns and pattern projections. This implies that keys and values are independent of input, suggesting an interpretation of $\bY$ as an identity matrix.
\end{itemize}
\begin{remark}
    For outlier efficient Hopfield model with $\lse_K$ energy (\cref{remark:softmax_K}), 
    the corresponding deep learning layer becomes $\Softmax_K(x_i)=\exp(x_i)/(K+\sum_j\exp(x_j))$.\footnote{\url{https://github.com/softmax1/Flash-Attention-Softmax-N}} 
\end{remark}

\section{Theoretical Analysis}
\label{sec:theory}
In this section, we validate our model as a theoretically robust Hopfield model.
Furthermore, by establishing a lower upper bound on retrieval error, we prove the proposed model's enhancements over its original counterpart, including expanded memory capacity.

\subsection{Convergence Guarantee}

We start our analysis with the notion of memory storage and retrieval\footnote{
A fixed point of $\calT$ with respect to $\calH$ is a point where $\bx = \calT(\bx)$, and a generalized fixed point is a point where $\bx\in\calT(\bx)$. For more details, refer to \cite{sriperumbudur2009convergence}.} of modern Hopfield models following \cite{wu2023stanhop,hu2023SparseHopfield,ramsauer2020hopfield}.
\begin{definition} [Storage and Retrieval]
\label{def:stored_and_retrieved}
For all $\mu\in[M]$, let $R\coloneqq \half \Min_{\mu,\nu\in[M];\mu\neq\nu}\norm{\bxi_\mu-\bxi_\nu}$ be the finite radius 
 of each sphere $\calS_\mu$ centered at memory pattern $\bxi_\mu$.
We say $\bxi_\mu$ is \textit{stored} if all $\bx\in\calS_\mu$ are generalized fixed points of $\calT$, $\bx^\star_\mu \in \calS_\mu$, and $\calS_\mu \cap \calS_\nu=\emptyset$ for $\mu \neq \nu$.
We say $\bxi_\mu$ is $\epsilon$-\textit{retrieved} by $\calT$ with $\bx$ for an error $\epsilon$, if $\norm{\calT(\bx)-\bxi_\mu}\le \epsilon$.
\end{definition}

\cref{def:stored_and_retrieved} does not guarantee alignment between $\calT_{\text{OutEff}}$'s fixed points and $\calH$'s stationary points.
Additionally, the monotonicity of equation \eqref{eqn:retrieval_dyn} does not ensure the existence of stationary points concerning energy $\calH$ \cite{sriperumbudur2009convergence}.
In the following lemma, we establish our proposed model as a well-defined Hopfield model by demonstrating two types of convergence.

\begin{theorem}[Convergence of $\calT_{\text{OutEff}}$]
\label{lemma:convergence_sparse}
Suppose $\calH$ is given by \eqref{eqn:H_energy} and $\calT_{\text{OutEff}}(\bx)$ is given by \eqref{eqn:retrieval_dyn}.
For any sequence $\{\mathbf{x}_t\}_{t=0}^{\infty}$ defined by $\mathbf{x}_{t'+1} = \calT_{\text{OutEff}}(\mathbf{x}_{t'})$, all limit points of this sequence are stationary points if they are obtained by iteratively applying $\calT_{\text{OutEff}}$ to $\mathcal{H}$.
\end{theorem}
\vspace{-.5em}\begin{proof}[Proof Sketch]
Following  \cite{hu2023SparseHopfield}, we first show that $\calH$ converges to its generalized fixed point $\bx^\star_\mu$ through $\calT_{\text{OutEff}}$ (1st convergence guarantee).
Then, we show that ${\bx^\star_\mu}$ corresponds to the stationary points of the energy minimization, and hence $\calH$ converges to local optimum (2nd convergence guarantee).
See \cref{appdix:convergence_sparse} for a proof.
\end{proof}\vspace{-1em}

\subsection{Retrieval Error Analysis}
Calibrating against the standard results \cite{ramsauer2020hopfield}, we prove the superiorities of the proposed model.

\begin{theorem}[Retrieval Error]\label{thm:retrieval_error}
Let $\calT_{\text{original}}$ be the retrieval dynamics of the original modern Hopfield model \cite{ramsauer2020hopfield}. 
$\left\| \calT_{\text{OutEff}}(\bx) - \bxi_{\mu}\right\|$ has lower upper bound than $\left\| \calT_{\text{original}}(\bx) - \bxi_{\mu}\right \|$ for all $\bx \in S_{\mu}$
\end{theorem}
\begin{corollary}[Tighter Retrieval Error]\label{coro:small_retrie_error}
Assume all patterns $\bx$ and $\{\bxi_\mu\}_{\mu\in[M]}$ are normalized.
Let $\gamma \coloneqq \sum\limits_{\mu=1}^{M} \[\Softmax_1\(\beta\bm{\bm{\Xi}}^\sT \bx\)\]_\mu $ and $\alpha$ be the angle between $\calT_{\text{original}}(\bx)$ and $\bxi_{\mu}$.
    It holds $\left\| \calT_{\text{OutEff}}(\bx) - \bxi_{\mu}\right\| \leq \left\| \calT_{\text{original}}(\bx) - \bxi_{\mu}\right\|$ when $ (\gamma+1)/2 \geq \cos{\alpha}$.
\end{corollary}
\vspace{-.5em}\begin{proof}
See \cref{appdix:retrieval_error} and \cref{appdix:coro_retrieval} for detailed proofs of \cref{thm:retrieval_error} and \cref{coro:small_retrie_error}.
\end{proof}\vspace{-.5em}
\begin{remark}
\cref{coro:small_retrie_error} is typically observed at the beginning of retrieval.
\end{remark}

\begin{theorem}[Memory Capacity Lower Bound, Informal]
\label{lemma:capacity}
    Assume all memory patterns are randomly sampled from a sphere of radius $m$.
    For any $\beta>0$, our proposed model's capacity to store and retrieve patterns scales exponentially with the pattern size $d$, and has a larger capacity lower bound than that of original modern Hopfield model \cite{ramsauer2020hopfield}:
    $M\ge M_{\text{original}}$.
\end{theorem}
\vspace{-.5em}\begin{proof}
    See \cref{appdix:capacity} for a detailed proof.
\end{proof}\vspace{-.5em}
\begin{remark}
    Comparing previous asymptotic larger capacity results of sparse models \cite{hu2023SparseHopfield,wu2023stanhop} with large $\beta$,
    \cref{lemma:capacity} is exact for all $\beta>0$.

\end{remark}

\subsection{Generalization Bound}

\begin{table*}[!t]
    \centering
    \caption{\textbf{Comparing   OutEffHop with Vanilla Attention in BERT, OPT, {ViT} and STanHop-Net.}
    We showcase the outlier efficiency of $\mathtt{OutEffHop}$ in 3 large transformer-based and 1 Hopfield-based models, using  Average Kurtosis and Maximum Infinity Norm $\norm{\bx}_{\infty}$. Additionally, 
    we showcase the quantization performance of $\mathtt{OutEffHop}$, 
    by comparing FP16 and W8A8 (Weight-8bit-Activation-8bit) performance. 
    The best results are highlighted in bold, and the second-best results are underlined. 
    In all settings, $\mathtt{OutEffHop}$ delivers significant outlier reduction, and further enhances its combinations with  $\mathtt{Clipped\_Softmax}$ and $\mathtt{Gated\_Attention}$.
    {$^*$For FP16 and W8A8, we report  \textit{Perplexity Score} for BERT and OPT,  \textit{Top-1 Accuracy} for ViT, and  \textit{Mean Square Error} (MSE) for STanHop-Net.}
}
\vspace{1em}
    \resizebox{1\textwidth}{!}{%
    \begin{tabular}{ccccccc}
        \toprule
    Model & Method & Avg. kurtosis & Max inf. norm & FP16$^*$ & W8A8$^*$ & Parameters \\
    \midrule
    \multirow{6}{1em}{\rot{BERT}} & Vanilla & 418.724 $\pm$ 0.814 & 255.859 $\pm$ 0.004 & 6.237 $\pm$ 0.001 & 7.154 $\pm$ 0.009 & \multirow{4}{*}{108.9m} \\
    & OutEffHop & \cellcolor{LightCyan} 26.564 $\pm$ 0.022 & \cellcolor{LightCyan} 33.618 $\pm$ 0.000 & \cellcolor{LightCyan} 6.209 $\pm$ 0.001 & \cellcolor{LightCyan} 6.295 $\pm$ 0.001 \\
    \cline{2-6}
    & Clipped Softmax & \underline{14.210} $\pm$ 0.003 & 33.619 $\pm$ 0.001 & \textbf{6.118} $\pm$ 0.002 & \textbf{6.189} $\pm$ 0.001 \\
    & Clipped OutEffHop & \textbf{11.839} $\pm$ 0.001 & \textbf{30.107} $\pm$ 0.001 & \underline{6.133} $\pm$ 0.000 & \underline{6.199} $\pm$ 0.001 \\
    \cline{2-7}
    & Gated Attention & 17.779 $\pm$ 0.014 & 34.082 $\pm$ 0.000 & 6.230 $\pm$ 0.001 & 6.299 $\pm$ 0.003 & \multirow{2}{*}{109m} \\
    & Gated OutEffHop & 15.625 $\pm$ 0.012 & \underline{32.777} $\pm$ 0.000 & 6.214 $\pm$ 0.001 & 6.279 $\pm$ 0.003 \\
    \midrule
    \multirow{6}{1em}{\rot{OPT}} & Vanilla & 23341.513 $\pm$ 27.363 & 92.786 $\pm$ 0.002 & 15.974 $\pm$ 0.001 & 42.012 $\pm$ 19.514 & \multirow{4}{*}{124.06m} \\
    & 
 OutEffHop & \cellcolor{LightCyan} 
     \underline{21.542} $\pm$ 0.000 & \cellcolor{LightCyan} 
     \underline{13.302} $\pm$ 0.001 & \cellcolor{LightCyan}  15.916 $\pm$ 0.002 & \cellcolor{LightCyan}  16.429 $\pm$ 0.013\\
    \cline{2-6}
    & Clipped Softmax & 9731.110 $\pm$ 0.000 & 43.803 $\pm$ 0.000 & 16.042 $\pm$ 0.000 & 30.825 $\pm$ 0.330 \\
    & Clipped OutEffHop & 24127.332 $\pm$ 0.000 & 67.602 $\pm$ 0.000 & 16.118 $\pm$ 0.000 & 29.269 $\pm$ 0.184 \\
    \cline{2-7}
    & Gated Attention & 90.321 $\pm$ 0.000 & 13.704 $\pm$ 0.000 & \textbf{15.677} $\pm$ 0.000 & \underline{16.236} $\pm$ 0.074 & \multirow{2}{*}{124.07m} \\
    & Gated OutEffHop & \textbf{11.449} $\pm$ 0.000 & \textbf{7.568} $\pm$ 0.000 & \underline{15.751} $\pm$ 0.000 & \textbf{16.148} $\pm$ 0.005\\
    \midrule
    \multirow{6}{1em}{\rot{{ViT}}} & {Vanilla} & {37.104 $\pm$ 0.000} & {272.198 $\pm$ 0.000} & {\underline{76.810} $\pm$ 0.000} & {74.935 $\pm$ 0.046} & {\multirow{4}{*}{22.03m}} \\
    & 
 {OutEffHop} & \cellcolor{LightCyan}  {31.601 $\pm$ 0.001} & {\cellcolor{LightCyan}  249.163 $\pm$ 0.000} & {\cellcolor{LightCyan} 
 76.788 $\pm$ 0.000} & {\cellcolor{LightCyan} \textbf{76.313} $\pm$ 0.012} \\
    \cline{2-6}
    & {Clipped Softmax} & { 33.868 $\pm$ 0.00} & {257.613 $\pm$ 0.00} & {76.612 $\pm$ 0.000} & {75.179 $\pm$ 0.013} \\
    & {Clipped OutEffHop} & {\underline{24.642} $\pm$ 0.000} & {\underline{196.199} $\pm$ 0.001} &  {\textbf{76.871} $\pm$ 0.001} & {\underline{76.083} $\pm$ 0.007} \\
    \cline{2-7}
    & {Gated Attention} & {45.145 $\pm$ 0.864} & {269.279 $\pm$ 1.426} & {69.922 $\pm$ 2.436} & {67.479 $\pm$ 1.447} &  {\multirow
{2}{*}{22.04m}} \\
    & {Gated OutEffHop}  & {\textbf{21.979} $\pm$ 0.254} & {\textbf{60.169} $\pm$ 1.153} & {74.089 $\pm$ 2.585} & {73.958 $\pm$ 3.126}\\
    \midrule
    \multirow{6}{1em}{\rot{STanHop-Net}} & Vanilla & 2.954 $\pm$ 0.063 & 5.048 $\pm$ 0.232 & \underline{0.360} $\pm$ 0.008 & 0.362 $\pm$ 0.000 & \multirow{4}{*}{35.13m} \\
    & 
 OutEffHop & \cellcolor{LightCyan}  2.897 $\pm$ 0.011 & \cellcolor{LightCyan}  4.565 $\pm$ 0.209 & \cellcolor{LightCyan} 
 \textbf{0.360} $\pm$ 0.004 & \cellcolor{LightCyan} 0.355 $\pm$ 0.000 \\
    \cline{2-6}
    & Clipped Softmax & 2.995 $\pm$ 0.05 & 4.890 $\pm$ 0.17 & 0.553 $\pm$ 0.03 & 0.591 $\pm$ 0.000 \\
    & Clipped OutEffHop & 2.864 $\pm$ 0.06 & \textbf{4.145} $\pm$ 0.23 &  0.506 $\pm$ 0.05 & 0.517 $\pm$ 0.000 \\
    \cline{2-7}
    & Gated Attention & \underline{2.487} $\pm$ 0.017 & 4.277 $\pm$ 0.163 & 0.380 $\pm$ 0.006 & 0.375 $\pm$ 0.000 &  \multirow{2}{*}{35.15m} \\
    & Gated OutEffHop  & \textbf{2.459} $\pm$ 0.041 & \underline{4.240} $\pm$ 0.155 & 0.376 $\pm$ 0.007 & 0.367 $\pm$ 0.000\\
        \bottomrule
    \end{tabular}
    }
    \label{tab:result1}
    \vspace{-1em}
\end{table*}

Following notations from \cref{sec:DL}, we analyze the generalization of the proposed layers. 
    Consider the input query $\bQ=[\bq_1, ... , \bq_T]^\top \in \mathbb{R}^{T \times d}$ and memory pattern $\bY=[\by_1, ... , \by_M]^\top \in \mathbb{R}^{M \times a}$, where $\by\in \mathbb{R}^a$ and $\bq\in \mathbb{R}^d$.

As standard supervised learning setting, we set the sample size (number of sequences) to be $N$, i.e. \textit{input} queries $\bQ^{(1)}, \bQ^{(2)}, ... , \bQ^{(N)}$, and the corresponding \textit{target} memory sets to be $Y^{(1)}, Y^{(2)}, ... , Y^{(N)}$. 
For vectors, $\norm{\cdot}_{p}$ and $\norm{\cdot}$ denote the $\ell_p$-norm and $\ell_2$-norm of vectors, respectively. 
For matrices, $\|\cdot\|_{p}$ denotes the $\ell_p$-norm, and $\|\cdot\|_{p, q}$ the $(p, q)$ matrix norm, which is $q$-norm of the $p$-norm of the columns of a matrix.
Namely, $\norm{\bA}_{p, q}=\|(\norm{\ba_{1
}}_{p}, \ldots,\norm{\ba_{i}}_{p},\ldots )\|_{q}$, where $\ba_{i}$ is  the $i$-th column vector of $\bA$.

Here, we consider the transpose of \eqref{eqn:OutEffHop} and taking $\tilde{\bW}_V:=\bW_K \bW_V$, while taking query $\bQ$ and raw memory pattern $\bY$ as inputs.
We write the Outlier-Efficient Modern Hopfield mechanism as a function $f_{\text{hop}}: \mathbb{R}^{M \times a} \times \mathbb{R}^{T \times d} \rightarrow \mathbb{R}^{d \times T}$ (i.e. $f_{\text{hop}}$ is the transpose of $\mathtt{OutEffHop}$ in \eqref{eqn:OutEffHop}):
\begin{align*} 
    f_{\text{hop}}(\bY,\bQ;\bW_{K}, \tilde{\bW}_{V})
    =\tilde{\bW}_V^\sT \bY^\sT  \Softmax_1 \( \beta \bY \bW_K \bQ^\sT \),
\end{align*}
where $\bW_K \in \mathbb{R}^{a\times d}$ and $\tilde{\bW}_V \in \mathbb{R}^{a \times d}$.
 Also, we write the corresponding function class
\begin{align*}
     & \calF_{\text {hop }} 
      \coloneqq
    \{(\bY,\bQ) \mapsto f_{\text{hop}}(\bY,\bQ;\bW_{K}, \tilde{\bW}_{V}) \mid
    \\
    &\quad\quad\quad\quad\quad\quad\quad\quad\quad\quad
    \bW_{K}\in \mathcal{W}_{K},  \tilde{\bW}_{V}\in \tilde{\mathcal{W}}_{V} \}.
    \nonumber
\end{align*}
and make the following mild assumptions.
\begin{assumption}[Norm Bounds]
\label{assumption:norm_bounds}
We assume that

    \textbf{(A1).} \label{item:A1}
    Query vectors $\br_\tau$ are bounded by 1 in $\ell_{2}$-norm
        \bea 
        \norm{\bq_\tau}\le 1\quad\forall \tau\in[T].\nonumber
        \eea 
    \textbf{(A2).} \label{item:A2}
    Memory vectors $\by_t$ are bounded in $\ell_{2}$-norm
        \bea 
        \norm{\by_t}\le B_Y \quad\forall t\in[M].\nonumber
        \eea 
    \textbf{(A3).} \label{item:A3}
    $\bW_K$ is bounded in $\ell_{2,1}$-norm
        \begin{align}
        \mathcal{W}_K:
        \{  \bW_K \in \mathbb{R}^{a \times d}\mid \norm{ \bW_{K}^\top}_{2} \leq B_{K},
        \left\| \bW_K\right\|_{2,1} \leq B_{K}^{2,1}  \}.
        \nonumber
        \end{align}
    \textbf{(A4).} \label{item:A4}
    $\tilde{\bW}_{V}$ is bounded in $\ell_{2}$-norm and $\ell_{2,1}$-norm
        \begin{align}
        \tilde{\mathcal{W}}_V:
        \{ \tilde{\bW}_V \in \mathbb{R}^{a \times d} \mid
        \big\| \tilde{\bW}_V^\sT \big\|_{2} \leq B_{V}, 
        \big\| \tilde{\bW}_V \big\|_{2,1} \leq B_{V}^{2,1}
        \}.
        \nonumber
        \end{align}
\end{assumption}
Then, we state our generalization results.
\begin{theorem}[Outlier Efficient Hopfield Layer Generalization Bound]
\label{lemma:gen_OutEffHop}
For any $\delta>0$, with probability at least $1-\delta$,
\begin{align}
\varepsilon_{\mathrm{gen}}(f_{\text{hop}}) 
\le
\tilde{\calO}
\(
\sqrt{N^{-1}}
\[
\sqrt{
\(E_1+E_2\)^3}+
\sqrt{\log\(1/\delta\)}
\]
\)
\nonumber, 
\end{align}
where $E_1=\big[  4 B_V^2 B_Y^2 ( \beta 
 B_{K}^{2,1} )^{2} \log \left(d N M\right)
    \big]^{1/3} $, $E_2=\big[  ( B_V^{2,1} )^2 \log(d N M) \big]^{1/3}$.
\end{theorem}
\vspace{-.5em}\begin{proof}[Proof Sketch]
    We first derive the covering number bound of the Outlier-Efficient Hopfield Layer by showing the Lipschitzness of $f_\text{hop}$ (\cref{lemma:lip_fhop_parameter}).
    By Dudley's Theorem, we obtain the generalization bound via covering number (\cref{lemma:gen_thru_cn}).
    See \cref{appendix:theorem_gen_outeffhop} for a detailed  proof.
\end{proof}\vspace{-1em}
Our results indicate that the generalization error remains controllable as long as the size of the data $N$ \textit{at least} scales logarithmically with the pattern dimension $d$ and the size of stored memory set $M$.
In addition, the length of the sequence, $T$, does not impact generalization, making Hopfield layers as a promising alternative to transformer attention.

\section{Experimental Studies}
\label{sec:exp}

\begin{figure*}[!h]
\begin{tabular}{cccc}
\raisebox{+\height}{\rot{\quad\tiny (a) BERT}} & \includegraphics[width=0.3\linewidth]{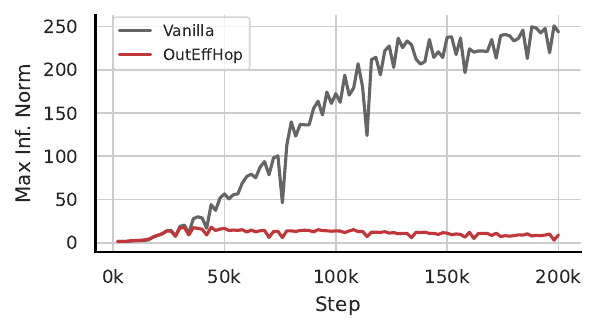} &
\includegraphics[width=0.3\linewidth]{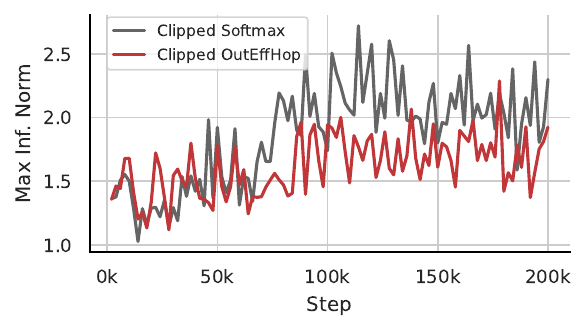} &
\includegraphics[width=0.3\linewidth]{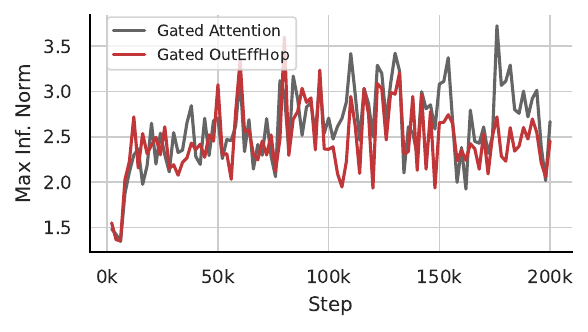} 
\vspace{-1.4em}\\
\raisebox{+\height}{\rot{\quad\tiny (b) OPT}} & \includegraphics[width=0.3\linewidth]{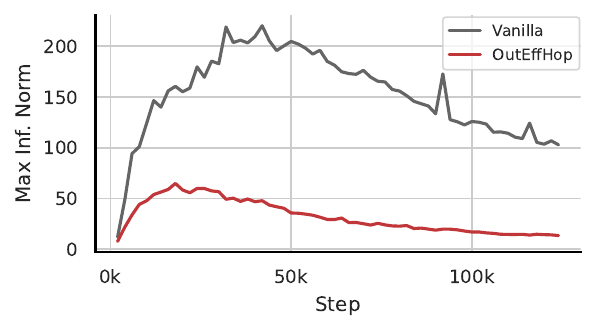} &
\includegraphics[width=0.3\linewidth]{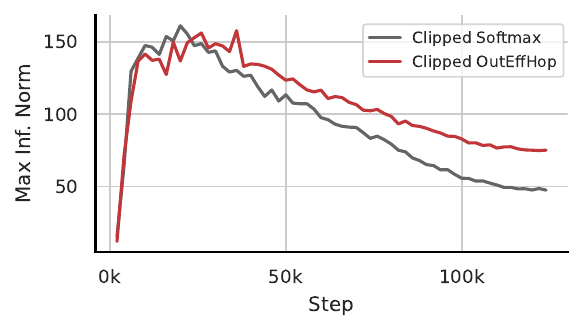} &
\includegraphics[width=0.3\linewidth]{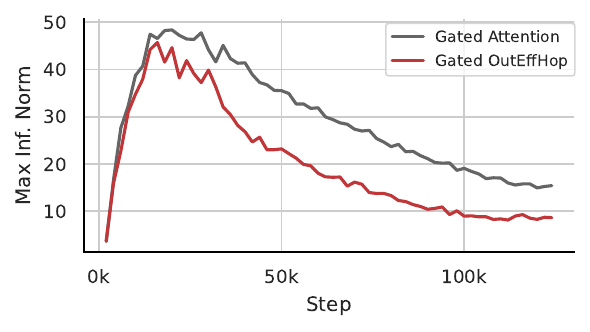} 
\vspace{-1.5em}\\
\raisebox{+\height}{\rot{\quad\tiny (c) ViT}} & 
\includegraphics[width=0.3\linewidth]{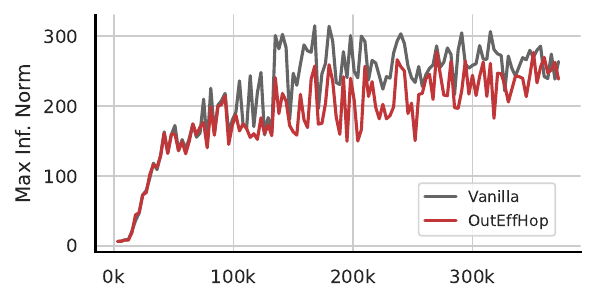}&
\includegraphics[width=0.3\linewidth]{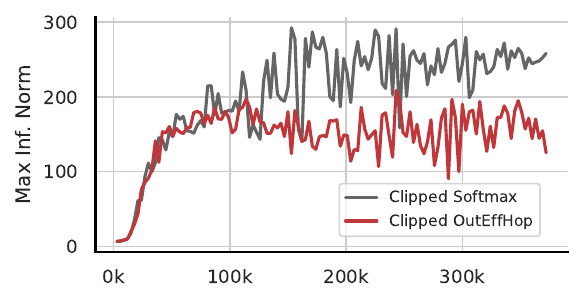} &
\includegraphics[width=0.3\linewidth]{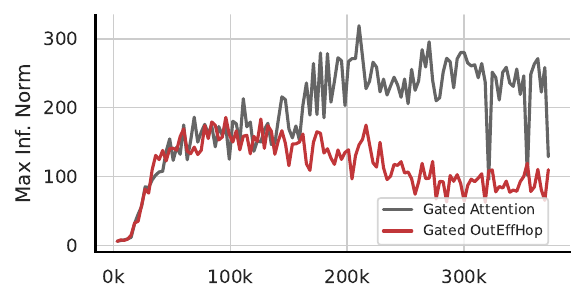} 
 \vspace{-0.8em}\\
\raisebox{+\height}{\rot{\hspace{-.5em}\tiny (d) STanHop-Net}} & 
\includegraphics[width=0.3\linewidth]{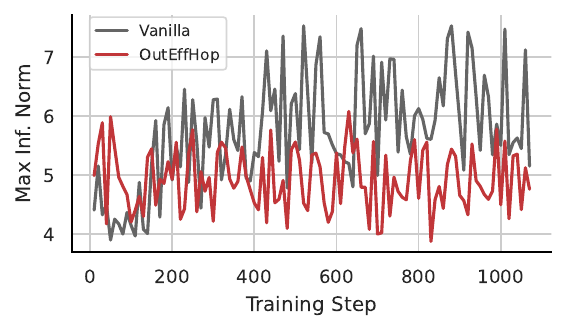}&
\includegraphics[width=0.3\linewidth]{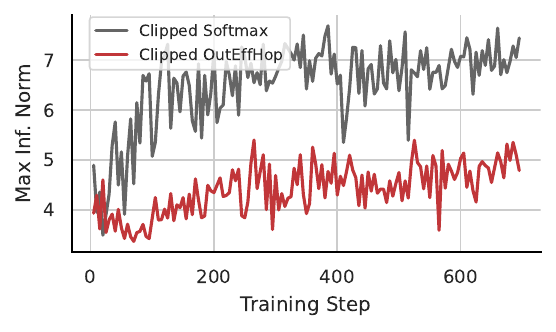} &
\includegraphics[width=0.3\linewidth]{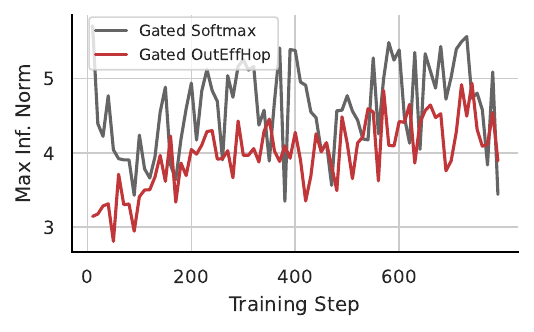} 
\\
\end{tabular}
\caption{
\textbf{The Impact of $\mathtt{OutEffHop}$ on Maximum Infinity Norm $\norm{\mathbf{x}}_{\infty}$ Changes During Pretraining of (a) BERT, (b) OPT, {(c) ViT}, and (d) STanHop-Net.} 
The plots, from left to right, compare $\mathtt{OutEffHop}$  with the vanilla attention baseline and their combination with $\mathtt{Clipped\_Softmax}$ and $\mathtt{Gated\_Attention}$ as per \cite{bondarenko2023quantizable}. 
Each figure's y-axis scale varies. For better visualization, 
we focus on the outlier reduction in layer 10 of the BERT, {ViT} and
OPT model, and in layer 9 of the STanHop-Net.
In all settings, $\mathtt{OutEffHop}$ delivers  significant reduction of the 
$\norm{\mathbf{x}}_{\infty}$ compared to the vanilla attention and improves $\mathtt{Clipped\_Softmax}$ and $\mathtt{Gated\_Attention}$. 
}
\label{fig:combined}
\end{figure*}

\begin{figure}[!h]
\minipage{0.5\textwidth}
\minipage{0.5\textwidth}
\includegraphics[width=\linewidth]{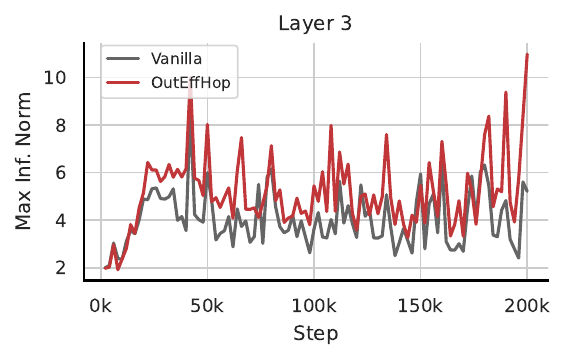}
\endminipage\hfill
\minipage{0.5\textwidth}
\includegraphics[width=\textwidth]{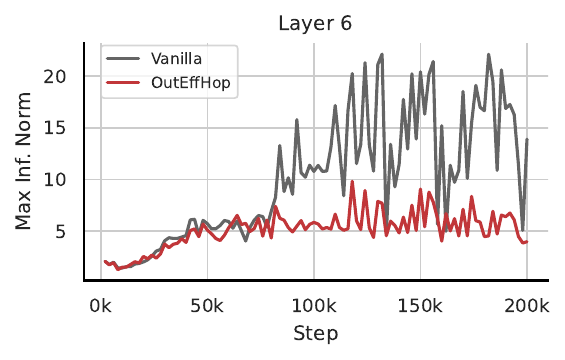}
\endminipage
\endminipage\hfill
\minipage{0.5\textwidth}
\minipage{0.5\textwidth}
\includegraphics[width=\textwidth]{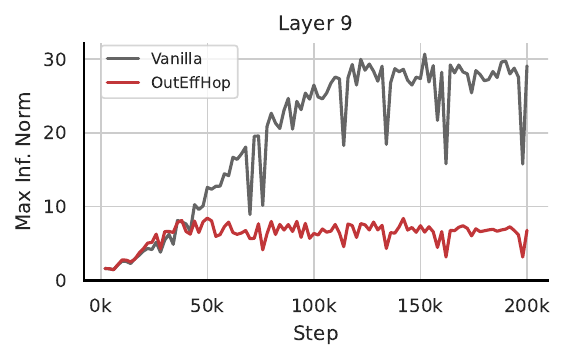}
\endminipage\hfill
\minipage{0.5\textwidth}
\includegraphics[width=\textwidth]{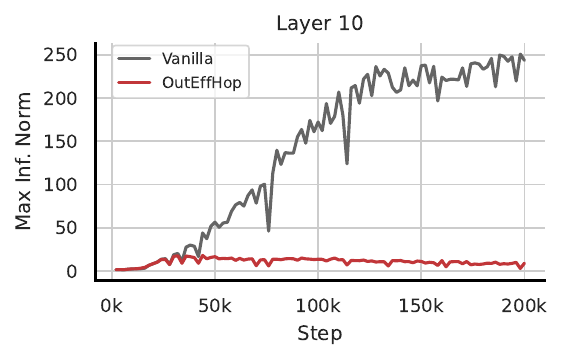}
\endminipage\hfill
\endminipage
\vspace{-1em}
\caption{The trend of Feed-Forward Network (FFN) output maximum infinity norm values in layers 3, 6, 9, and 10 of a BERT encoder is analyzed using two softmax variations: $\mathtt{OutEffHop}$ (represented in red) and vanilla $\Softmax$ (in grey). 
The findings indicate that $\mathtt{OutEffHop}$ significantly reduces outliers in the model compared to the vanilla $\Softmax$.}
\label{fig:differnet_layer}
\end{figure}

\begin{figure}[!h]
\minipage{0.5\textwidth}
\minipage{0.5\textwidth}
\includegraphics[width=\linewidth]{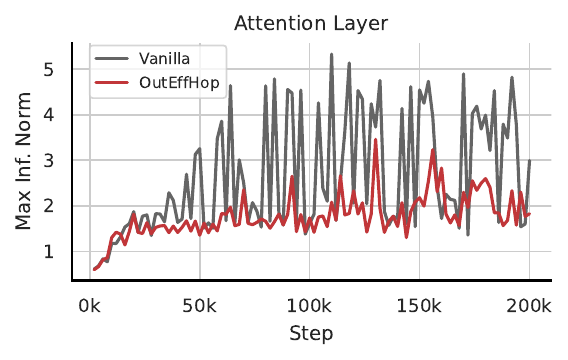}
\endminipage\hfill
\minipage{0.5\textwidth}
\includegraphics[width=\linewidth]{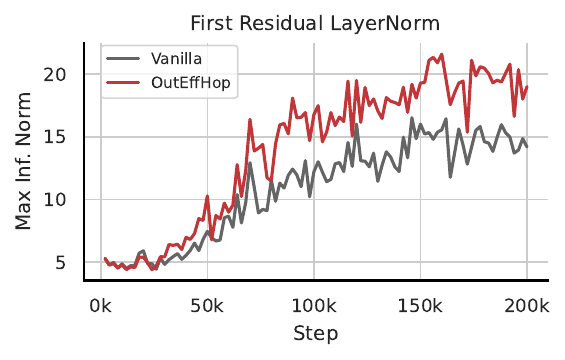}
\endminipage
\endminipage\hfill
\minipage{0.5\textwidth}
\minipage{0.5\textwidth}
\includegraphics[width=\linewidth]{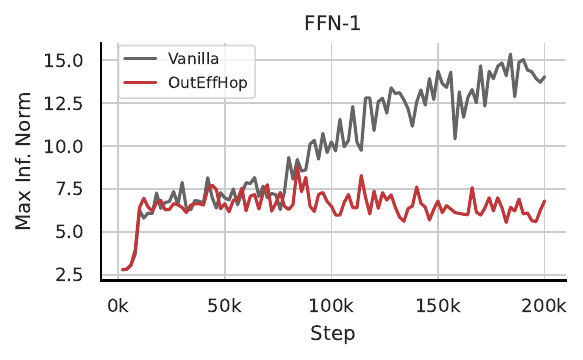}
\endminipage\hfill
\minipage{0.5\textwidth}
\includegraphics[width=\linewidth]{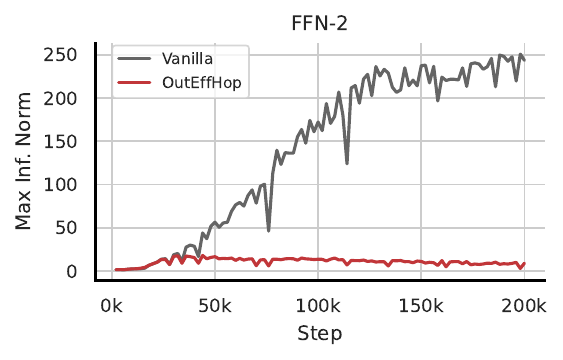}
\endminipage\hfill
\endminipage
\vspace{-1em}
\caption{
Maximum infinity norm $\| \mathbf{x} \|_{\infty}$ for different tensor components within layer 10 of BERT. Our work is analysed using two softmax variations: $\mathtt{OutEffHop}$ (represented in red) and vanilla $\Softmax$ (in grey). We find $\mathtt{OutEffHop}$ suppresses the outliers growing in both FFN layers.
}
\label{fig:same_layer}
\end{figure}

We conduct a series of experiments to validate the Outlier-Efficient Modern Hopfield Model and layers.
Specifically, 
we test our model in accordance with SOTA methods outlined in \cite{bondarenko2023quantizable}, {with 3 common large transformer-based models and 1 Hopfield-based model.}

\subsection{Outlier Efficiency of $\mathtt{OutEffHop}$}
\label{sec:exp_default}

To test the model's robustness against outliers,
we use $\mathtt{OutEffHop}$ in BERT \cite{devlin-etal-2019-bert}, Open Pretrained Transformers (OPT) \cite{zhang2022opt}, {Vision Transformers (ViT) \cite{dosovitskiy2020image}} and STanHop-Net \cite{wu2023stanhop} to replace the vanilla attention layer \cite{vaswani2017attention} and Hopfield layer \cite{hu2023SparseHopfield,ramsauer2020hopfield}. 
We then train these models from scratch and evaluate them on the validation set.
We conduct each evaluation three times with different random seeds and present the average and standard deviation for each metric.

\paragraph{Metrics.}
We report \textit{maximum infinity norm} $\norm{\mathbf{x}}_{\infty}$ of the activation tensors $\mathbf{x}$ across all the Transformer layers as the metrics for outliers' existence.
Also, we report the \textit{average kurtosis} of $\mathbf{x}$.
For BERT, we only average on the outputs tensors from the Feed-Forward Network (FFN) layer and Layer Normalization.
These two parts are known for outlier presence, as confirmed by our experiments and previous studies \cite{bondarenko2021understanding, wei2022outlier, bondarenko2023quantizable}.
In the case of OPT, {ViT} and STanHop, we average over every output component in the transformer layers.
These two metrics have been shown to correlate well with the model quantizability (i.e., robustness against outliers) \cite{bondarenko2021understanding, shkolnik2020robust}. 
Specifically, previous studies \cite{llm.int.8, wei2022outlier, bondarenko2021understanding} highlight a substantial decline in model performance attributed to quantization in the presence of outliers. 
As a result, 
we report the models' performance before and after quantization.
For before quantization performance, we report (i) \textbf{FP16} (in 16-bit floating-point) \textit{Perplexity Score} for BERT and OPT, {(ii) \textbf{FP32} \textit{Top-1 Accuracy} for ViT, } and (iii) \textit{Mean Square Error} (MSE) for STanHop-Net.
For after quantization performance \textbf{W8A8} (in 8-bit floating-point), 
we report the same metrics.

\paragraph{Datasets.}
We use 4 real-world datasets: Bookcorpus \cite{Zhu_2015_ICCV}, wiki40b/en \cite{guo2020wiki}, {ImageNet-1k \cite{imagenet15russakovsky} }and ETTh1 \cite{zhou2021informer}. 
The first two are for language models, i.e. OPT and BERT, {the third is for vision model, i.e. ViT,} and the last is for time series model, i.e. STanHop-Net.

\paragraph{Models.}
Following \citet{bondarenko2023quantizable}, we validate our method ($\mathtt{OutEffHop}$ layers) with 4 popular models:
2 language models (BERT, OPT), {1 vision model (ViT)} and 1 time series model (STanHop).
For BERT, we adopt the BERT-base-uncased model of size 109 million parameters\footnote{\url{https://huggingface.co/bert-base-uncased}}. 
We pretrain this model with the masked language modeling (MLM) technique, following the original BERT paper \cite{devlin-etal-2019-bert}. 
As for OPT, we adopt a OPT model of size 125 million parameters\footnote{\url{https://huggingface.co/facebook/opt-125m}}. 
For this model, we employ causal language modeling (CLM) as the pre-training objective. 
To optimize training efficiency, 
we set specific constraints on sequence length: 
sequence length of 128 for BERT and of 512 for OPT.
{As for ViT, 
we adopt the ViT-S\_16 variant of size  {22.03} million parameters\footnote{\url{https://huggingface.co/WinKawaks/vit-small-patch16-224}}. 
We pretrain this model with standard image classification objective.}
As for STanHop-Net, 
we adopt a STanHop-Net of size 35.13 million parameters\footnote{\url{https://github.com/MAGICS-LAB/STanHop}}. 
We pretrain this model on a multivariate time series prediction objective.

\textbf{Results.}
In \cref{tab:result1} and \cref
{fig:combined}, our results show that $\mathtt{OutEffHop}$ achieves performance in outlier reduction comparable to $\mathtt{Clipped\_Softmax}$ and $\mathtt{Gated\_Attention}$. 
Moreover, combining $\mathtt{OutEffHop}$ with these two methods further improves the effect, {achieving an average reduction of $\sim$22+\% in average kurtosis and $\sim$26+\% in maximum infinity norm across four test models.}
The only exception is Clipped $\mathtt{OutEffHop}$ in the OPT model.
This anomaly aligns with the findings of \citet{bondarenko2023quantizable}, which suggest that the $\mathtt{Clipped\_Softmax}$ approach does not perform well with OPT. 
In sum,
the efficacy of $\mathtt{OutEffHop}$ is also apparent in the reduction of the maximum infinity norm value during the pre-training process, 
particularly noticeable in layer 10 of the BERT, {ViT} and OPT model, and in layer 9 of the STanHop models, as depicted in \cref{fig:combined}.  $\mathtt{OutEffHop}$ is more efficient at reducing outliers during the pretraining process compared to its baseline methods, with particularly notable improvements in the OPT model.

\begin{table*}[t]
\centering
\caption{\small
\textbf{STanHop-Net \cite{wu2023stanhop}: {Outlier Reduction of} Multivariate Time Series Predictions.}
We implement 4 STanHop variants, \textbf{Hopfiled} with \textbf{D}ense $\mathtt{Hopfield}$ layer \cite{ramsauer2020hopfield}, \textbf{SparseHopfiled} with \textbf{S}parse $\mathtt{SparseHopfield}$ layer \cite{hu2023SparseHopfield}, \textbf{STanHop-Net} with GSH layer \cite{wu2023stanhop} and \textbf{OutEffHop} with our $\Softmax_1$ layer respectively. 
{
To evaluate outlier reduction performance, we report the maximum infinity norm and average kurtosis metrics. 
}
We also report the average Mean Square Error (MSE) and Mean Absolute Error (MAE) metrics with variance omitted as they are all $\le 2$\%.
We evaluate each dataset with different prediction horizons (shown in the second column).
We have the best results \textbf{bolded} and the second best results \underline{underlined}.
In 25 out of 30 settings, $\mathtt{OutEffHop}$ ranks either first or second.
{Our results indicate that our proposed $\mathtt{OutEffHop}$ delivers consistent top-tier outlier-reduction performance compared to all the baselines.}
}
\vspace{.5em}
\resizebox{\textwidth}{!}{    
\begin{tabular}{ccccccccccccccccgg}
\toprule
 \multicolumn{2}{c}{Models} &  
 \multicolumn{4}{c}{\footnotesize\textbf{\texttt{Hopfield}}} &
 \multicolumn{4}{c}{\footnotesize\textbf{\texttt{SparseHopfield}}} &
 \multicolumn{4}{c}{\textbf{STanHop-Net} (\texttt{\textbf{GSH}})} &
  \multicolumn{4}{c}{\textbf{\texttt{OutEffHop}}} \\
\midrule
 \multicolumn{2}{c}{Metric} &   MSE & MAE & \makecell{Avg. \\ kurtosis} & \makecell{Max inf.\\ norm}  & MSE & MAE &\makecell{Avg. \\ kurtosis} & \makecell{Max inf.\\ norm}   & MSE & MAE & \makecell{Avg. \\ kurtosis} & \makecell{Max inf.\\ norm}  & MSE & MAE & \cellcolor{white} \makecell{Avg. \\ kurtosis} & \cellcolor{white}\makecell{Max inf.\\ norm}  \\
\midrule
 \multirow{5}{1em}{\rot{ETTh1}} 
 & 24 
 & 0.360 & 0.401 & \underline{2.954} $\pm$ 0.063 &  5.048 $\pm$ 0.232 & 0.388 & 0.411 & 3.311 $\pm$ 0.082 & 4.954 $\pm$ 1.064 & 0.395 & 0.415 & 3.269 $\pm$ 0.117 & \underline{4.947} $\pm$ 0.173 & 0.361 & 0.397 &  \textbf{2.897} $\pm$ 0.011 & \textbf{4.565} $\pm$ 0.209 \\ 
 & 48 
 & 0.405 & 0.424 & \underline{2.968} $\pm$ 0.039 &  4.969 $\pm$ 0.033 & 0.466 & 0.452 & 3.295 $\pm$ 0.136 & 4.749 $\pm$ 0.517 & 0.458 & 0.448 & 3.271 $\pm$ 0.200 & \underline{4.644} $\pm$ 0.341 & 0.409 & 0.426 &  \textbf{2.965} $\pm$ 0.004 & \textbf{4.570} $\pm$ 0.424 \\
 & 168 
 & 0.881 & 0.710 & \underline{2.545} $\pm$ 0.004 &  \underline{3.923} $\pm$ 0.115 & 1.422 & 0.921 & 3.149 $\pm$ 0.015 & 4.348 $\pm$ 0.085 & 1.422 & 0.926 & 3.093 $\pm$ 0.065 & 4.160 $\pm$ 0.285 & 0.872 & 0.704 &  \textbf{2.526} $\pm$ 0.011 & \textbf{3.865} $\pm$ 0.035 \\
 & 336
 & 0.755 & 0.648 & \underline{2.436} $\pm$ 0.003 &  \underline{3.536} $\pm$ 0.230 & 1.223 & 0.851 & 3.071 $\pm$ 0.009 & 4.156 $\pm$ 0.199 & 1.381 & 0.909 & 3.043 $\pm$ 0.021 & 4.248 $\pm$ 0.159 & 0.780 & 0.658 & \textbf{2.433} $\pm$ 0.009 & \textbf{3.416} $\pm$ 0.042 \\
& 720 
& 0.852 & 0.709 & \textbf{2.443} $\pm$ 0.006 & \underline{3.266} $\pm$ 0.132 & 1.134 & 0.824 & 3.030 $\pm$ 0.015 & 4.179 $\pm$ 0.054 & 1.360 & 0.904 & 3.062 $\pm$ 0.089 & 4.238 $\pm$ 0.197 & 0.894 & 0.788 & \underline{2.450} $\pm$ 0.035 & \textbf{3.218} $\pm$ 0.142 \\ 
\midrule
 \multirow{5}{1em}{\rot{ETTm1}} 
 & 24 
 & 0.272 & 0.339 & 3.617 $\pm$  0.003 & 4.717 $\pm$ 0.353  
 & \underline{0.265} & \underline{0.331} & \textbf{3.357} $\pm$ 0.045  & \underline{4.334} $\pm$ 0.087 
 & \textbf{0.261} & \textbf{0.328} & \underline{3.547} $\pm$ 0.096 & 4.696 $\pm$ 0.279 
 & 0.347 & 0.429 & 3.584 $\pm$ 0.136 & \textbf{4.212} $\pm$ 0.262 \\ 
 & 48 
 & 0.352 & 0.387 & \underline{4.211} $\pm$ 0.113  & \underline{5.603} $\pm$ 0.854 
 & \textbf{0.304}  & \textbf{0.355} & 4.280 $\pm$ 0.102 & 6.296 $\pm$ 0.479 
 & \underline{0.328} & \underline{0.367}  & 4.384 $\pm$ 0.415 & \textbf{5.557} $\pm$ 4.188  
 & 0.375 & 0.409 & \textbf{3.967} $\pm$ 0.253 & 5.816 $\pm$ 0.209 \\ 
 & 96 
 & 0.396 & 0.412 & \underline{3.102} $\pm$ 0.026 & 4.534 $\pm$ 0.328 
 & \underline{0.345} & 0.383 & 3.568 $\pm$ 0.127 & \underline{4.441} $\pm$ 0.650 
 & \textbf{0.344} & 0.375 & 3.609 $\pm$ 0.364 & 4.618 $\pm$ 0.319 
 & 0.529 & 0.487 & \textbf{3.014} $\pm$ 0.042 & \textbf{4.333} $\pm$ 0.394  \\ 
& 288 
&  0.600 & 0.540 & \underline{2.643} $\pm$ 0.005  & 3.179 $\pm$ 1.798 
& \textbf{0.500} & \textbf{0.471} & 2.783 $\pm$ 0.075  & \underline{3.172} $\pm$ 0.048 
& \underline{0.515}  & \underline{0.483} & 2.803 $\pm$ 0.101  &  3.228 $\pm$ 0.056 
& 0.572 & 0.513 &  \textbf{2.498} $\pm$ 0.031 & \textbf{3.151} $\pm$ 0.072 \\ 
& 672 
& 0.784 & 0.627 & \underline{2.674} $\pm$ 0.079 & 3.740 $\pm$ 0.318 
& \textbf{0.537} & \textbf{0.495} & 3.429 $\pm$ 0.206 &  3.875 $\pm$ 0.380  
& \underline{0.571} & \underline{0.519} & 3.427 $\pm$ 0.138 &  \textbf{3.439} $\pm$ 0.093 
& 0.752 & 0.607 & \textbf{2.553} $\pm$ 0.081 & \underline{3.641} $\pm$ 0.091 \\ 
\midrule
\multirow{5}{1em}{\rot{WTH}}
& 24 
& 0.357 & 0.404 & \textbf{3.616} $\pm 0.117$ & $6.668 \pm 1.102$
& 0.378 & 0.429 & \underline{3.656} $\pm 0.082$ & \underline{5.609} $\pm 0.154$
& 0.370 & 0.394 & $3.726 \pm 0.231$ & $9.126 \pm 0.322$
& 0.378 & 0.423 & $3.711 \pm 0.017$ & \textbf{5.428} $\pm 0.093$  \\
& 48 
& \underline{0.441} & \textbf{0.464} & \underline{3.904} $\pm$ 0.090 & \textbf{6.481} $\pm 0.417$   
& \textbf{0.441} & \underline{0.474} & $3.957 \pm 0.184$ & $7.409 \pm 1.445$  
& 0.472 & 0.500 & $3.911 \pm 0.282$ & $6.730 \pm 0.150$ 
& 0.464 & 0.480 & \textbf{3.663} $\pm 0.144$ & \underline{6.649} $\pm 0.586$ \\ 
& 168 
& \textbf{0.549} & \underline{0.562} & \underline{2.617} $\pm 0.046$ & $\underline{3.028} \pm 0.097$ 
& 0.575 & 0.575 & $2.835 \pm 0.012$  & $3.364 \pm 0.045$ 
& \underline{0.561} & 0.565 & $2.712 \pm 0.040$  & $3.087 \pm 0.089$  
& 0.562 & \textbf{0.561} & $\textbf{2.552} \pm 0.031$ & $\textbf{2.931} \pm 0.068$ \\ 
& 336 
& \underline{0.572} & \underline{0.579} & \textbf{2.565} $\pm$ 0.082 & \underline{3.185} $\pm$ 0.055 
& 0.598 & 0.593 & 2.849 $\pm$ 0.031 & 3.640 $\pm$ 0.078 
& \textbf{0.552} & \textbf{0.557} & 2.710 $\pm$ 0.072 & \textbf{3.087} $\pm$ 0.043 
& 0.613 & 0.604 & \textbf{2.516} $\pm$ 0.057 & 3.383 $\pm$ 0.063 \\ 
& 720 
& 0.727 & 0.670 & \underline{2.578} $\pm$ 0.027 & 3.617 $\pm$ 0.443 
& \textbf{0.591}  & \underline{0.587} & 2.737 $\pm$ 0.009 & \underline{3.228} $\pm$ 0.078 
& \textbf{0.571} & \textbf{0.573} & 2.737 $\pm$ 0.009 & \textbf{3.219} $\pm$ 0.073  
& 0.794 & 0.710 & \textbf{2.543} $\pm$ 0.006  & 3.524 $\pm$ 0.261  \\ 
\bottomrule
\end{tabular}
}
\label{table:result}
\vspace{-1.5em}
\end{table*}

\subsection{$\mathtt{OutEffHop}$ Improves Hopfield-Centric Deep Learning Model: A Case Study on STanHop-Net}
We also test our method on STanHop-Net \cite{wu2023stanhop}, 
a Hopfield-based time series prediction model. 
We conduct a comparison between our method and common modern Hopfield layers \cite{hu2023SparseHopfield,ramsauer2020hopfield}.

\paragraph{Data.}
Following \citet{wu2023stanhop}, 
we use 3 realistic datasets for multivariate time series prediction tasks:
ETTh1 (Electricity Transformer Temperature-hourly), ETTm1 (Electricity Transformer Temperature-minutely), WTH (Weather). 
We divide these datasets into training, validation, and test sets with a ratio of 14/5/5.
For each dataset, we conduct evaluations across various prediction horizons.

\paragraph{Metrics.}
To evaluate the outlier efficiency, 
we use the same metrics as the above experiments:
the maximum infinity norm $\norm{\bx}_{\infty}$ and \textit{average kurtosis} over 12 decoder layers.
To evaluate the prediction accuracy, we use Mean Squared Error (MSE) and Mean Absolute Error (MAE).
We repeat each experiment 10 times and report the average results. 

\paragraph{Results.} 
In \cref{table:result},
our results demonstrate the effectiveness of $\mathtt{OutEffHop}$ in enhancing  outlier efficiency of modern Hopfield network architectures.
$\mathtt{OutEffHop}$ delivers significant improvements on outlier efficiency with marginal sacrifice of model performance. $\mathtt{OutEffHop}$ achieves top-tier outlier-efficiency in 25 out of 30 evaluated scenarios, ranking either first or second in these settings. 
In STanHop-Net,  $\mathtt{OutEffHop}$ model demonstrates a notable enhancement in outlier efficiency compared to Vanilla and Sparse, Generalized Sparse Modern Hopfield Models. 
Specifically, there are 3\% and 4\% reductions in $\norm{\bx}_{\infty}$ and average kurtosis, respectively.

\subsection{Additional Experimental Results (\cref{sec:add_exps})}

\paragraph{\cref{fig:differnet_layer} \& \cref{fig:same_layer}.}
To supplement \cref{sec:exp_default},
We conduct in-depth case studies on the BERT model. 
In \cref{fig:differnet_layer}, we focus on the outlier performance in selected layers, and in  \cref{fig:same_layer}, we delve into the maximum infinity norm $\norm{\mathbf{x}}_{\infty}$ within the 10th layer's various tensor components.
$\mathtt{OutEffHop}$ offers evidence of its effectiveness in mitigating outliers within our approach. 
Additionally, we observe that this mitigation effect becomes particularly pronounced in the final several layers. 
See \cref{sec:sup_exp_details} for more details.

\paragraph{Verifying Theoretical Results.}
Following \cite{hu2023SparseHopfield,wu2023stanhop,ramsauer2020hopfield}, 
we validate the superiority of $\mathtt{OutEffHop}$'s theoretical results on memory retrieval and MIL learning tasks on 3 datasets, benchmarking against \cite{krotov2016dense,ramsauer2020hopfield,hu2023SparseHopfield,wu2023stanhop}.
See \cref{sec:thm_exps} for more details.

\vspace{-.5em}
\section{Conclusion and Discussion}
\label{sec:conclusion}We present the Outlier-Efficient Modern Hopfield Model to manage the computational challenges posed by outliers in large transformer-based models.
Our model not only inherits the appealing features of modern Hopfield models,
but also introducing the $\mathtt{OutEffHop}$ layers as new deep learning components for large transformer-based models with strong outlier-reducing capabilities.
{Empirically, $\mathtt{OutEffHop}$ achieves an average reduction of $\sim$22+\% in average kurtosis and $\sim$26+\% in maximum infinity norm across four test models.}
Additionally, 
it improves the same metrics by an average of 3\% and 4\% compared to 3 variants of STanHop-Net and ranks among the top two in outlier efficiency in 25 out of 30 settings.

\textbf{Limitation and Future Work.}
One limitation is that $\mathtt{OutEffHop}$ does not address outliers induced by LayerNorm (see First Residual LayerNorm in \cref{fig:same_layer}). 
In fact, \citet{wei2022outlier} observe that LayerNorm outliers arise from mechanisms different from those of attention, as studied here. 
We plan to integrate these different types of outliers with $\mathtt{OutEffHop}$ in future research.

\vspace{-1em}
\section*{Impact Statement}
We believe this methodology offers an opportunity to enhance the foundations of foundation models, including large language models, through insights from associative memory models. However, this approach could intensify biases in the training data, potentially resulting in unfair or discriminatory outcomes for underrepresented groups.

\clearpage
\newpage
\normalsize
\titlespacing*{\section}{0pt}{*1}{*1}
\titlespacing*{\subsection}{0pt}{*1.25}{*1.25}
\titlespacing*{\subsubsection}{0pt}{*1.5}{*1.5}

\setlength{\abovedisplayskip}{10pt}
\setlength{\abovedisplayshortskip}{10pt}
\setlength{\belowdisplayskip}{10pt}
\setlength{\belowdisplayshortskip}{10pt}
\onecolumn
\appendix
\part*{Supplementary Material}
\label{sec:append}

{
\setlength{\parskip}{-0em}
\startcontents[sections]
\printcontents[sections]{ }{1}{}
}
\begin{itemize}
    
    \item \hyperref[sec:related_works]{\textbf{Appendix~A.}} 
    \textbf{Related Works}

    \item  \hyperref[sec:sup_background]{\textbf{Appendix~B.}} \textbf{Supplementary Backgrounds}

    \item \textbf{\hyperref[sec:proofs]{Appendix~C.} Proofs of Main Text}

    \item 
    \textbf{\hyperref[sec:add_exps]{Appendix~D.} Additional Numerical Experiments}
\end{itemize}

\section{Related Works}
\label{sec:related_works}
\paragraph{Associative Memory Models for Deep Learning.}
The classical Hopfield models \cite{hopfield1984neurons,hopfield1982neural,krotov2016dense} mirror the associative memory of the human brain, focusing on the storage and retrieval of specific memory patterns.
Recently, a resurgence in associative memory model research is attributable to (i) advancements in memory storage capacities \cite{wu2024uniform,chaudhry2024long,krotov2016dense, demircigil2017model}, (ii) the innovative architectural designs \cite{wu2023stanhop,hoover2023energy,seidl2022improving,furst2022cloob,ramsauer2020hopfield}, and (iii) their biological plausibility \cite{burns2024semantically,kozachkov2022building,krotov2020large}.
Notably, the associative memory networks a.k.a. the modern Hopfield models \cite{hu2024nonparametric,hu2024computational,wu2023stanhop,burns2023simplicial,hu2023SparseHopfield,hopfeildblog2021,ramsauer2020hopfield} exhibit favorable properties, including fast convergence speed and exponential memory capacity.
They form a bridge to Transformer architecture \cite{hu2024nonparametric,hu2023SparseHopfield,wu2023stanhop,cabannes2023scaling,bietti2024birth,ramsauer2020hopfield}, positioning themselves as advanced extensions of attention mechanisms.
Consequently, their applicability extends across various fields, including drug discovery \cite{schimunek2023contextenriched}, immunology \cite{widrich2020modern}, time series forecasting \cite{wu2023stanhop,auer2023conformal}, tabular learning \cite{xu2024bishop}, out-of-distribution detection \cite{hofmann2024energy}, reinforcement learning \cite{paischer2022history}, and vision \cite{furst2022cloob}.
Our study refines this research direction towards efficient models.
We believe that this study is critical in guiding future research towards a Hopfield-driven design paradigm, especially for large-scale models.

\paragraph{Outlier-Efficient Methods.}
Quantization is a method to reduce the computational burden of large models via low-bit precision computing \cite{horowitz20141,229903, 182695}.
For instance, 
common quantization schemes, INT8 and INT4, compress the models' weights and activations by using 8-bit or 4-bit integers encoding \cite{zafrir2019q8bert,bhandare2019efficient,junczys2018marian}.
However, the presence of outliers challenges the quantization performance of transformer-based models due to outlier-induced exploding attention weights \cite{bondarenko2023quantizable,bondarenko2021understanding}.
To combat this,
\citet{wei2022outlier} modify LayerNorm to enable quantization on outlier-free activation tensors, and introduce Token-Wise Clipping to optimize clipping ranges for each token.
Further, \citet{llm.int.8}
quantize outlier features and other features with different degrees of precision.
Yet, 
since outliers stem from the softmax function (see \cref{background} for details), neither of above methods address the outlier issue from its source.
To this end, 
\citet{bondarenko2023quantizable} introduce $\mathtt{Clipped\_Softmax}$ and $\mathtt{Gated\_Attention}$ to force attention mechanism to output exact zeros, thereby tackling the source of outliers.
Specifically, $\mathtt{Clipped\_Softmax}$ extends the output range of the softmax function from (0, 1) to larger span, and $\mathtt{Gated\_Attention}$ determines to keep or nullify the update.
However, these two methods need hyperparameters for optimal performance.
Moreover, $\mathtt{Clipped\_Softmax}$ underperforms with the OPT model and $\mathtt{Gated\_Attention}$ introduces additional training parameters.
In this paper, 
we present a novel modern Hopfield model such that it endows outlier-efficient computation.
Surprisingly, its retrieval dynamics subsumes the $\Softmax_1$ outlier-efficient attention \cite{miller2021} as a special case\footnote{For any $\bx\in\R^d$,
$\Softmax_1(\bx)_i = \frac{\exp{x_i}}{1 + \sum_j \exp{x_j}}$.
Preliminary experimental results \cite{john2023} confirm its outlier efficiency.}.
We expect this work to shed light on research into (Hopfield-based) large foundation models, both theoretically and methodologically.

\paragraph{Outlier Related Transformer Theories.}
Recent works highlight the theoretical advantages of removing outliers in the attention heads of transformer-based large foundation models. \citet{alman2023fast} show that efficient transformers (both vanilla and tensor \cite{as23_tensor}) require bounded attention weights using fine-grained reduction. \citet{hu2024computational} show that efficient modern Hopfield models and corresponding networks also need bounded query and key patterns for sub-quadratic time complexity via fine-grained reduction. \citet{gu2024tensor,gu2024conv,alman2024fine,gao2023fast} show that efficient training of transformer-based models necessitates bounded weight matrices.

\clearpage
\section{Supplementary Backgrounds}
\label{sec:sup_background}

\subsection{How Does Softmax$_1$ Solve the Outlier Problem?}
\label{sec:softmax1_nb}

The ``outlier'' challenge in (multi-head) attention arises from the inherent design of $\Softmax$.
$\Softmax$ forces each attention head to attend to at least one position in the input sequence, even if there is no useful information throughout the entire input sequence. In the case where a \textbf{no-update} behavior is needed, since for $\Softmax$, producing close-to-zero probabilities for all positions is not an option, it has to produce a high probability to a spurious position (such as a comma sign in a sentence) and produces close-to-zero probabilities for all other positions. However, this is a workaround and it still introduces noises.

{
To solve this,
\citet{miller2021} proposes $\Softmax_1$, which adds $1$ to the denominator of $\Softmax$.
This adjustment reduces the relative importance of each head.
As a result, if a head provides less relevant or even misleading information, the model does not depend on it.
This ensures the influence of ``less-relevant heads" remains moderate.
Therefore, $\Softmax_1$ allows a head to ``abstain" or contribute minimally when its information is not beneficial for the current context.

In this paper, we introduce the Outlier-Efficient modern Hopfield model for two purposes:
\begin{enumerate}
    \item[I.] Outlier Efficient Associative Memory Model
    
    \item[II.] Outlier Efficient Attention-like Layer for deep learning
\end{enumerate}

We only have to identify outlier when our model serves as Associtative Memory Models .  
For similarity measure thersholding, it has following process:
\begin{enumerate}
    \item Calculating similarity scores among patterns.
    \item Setting a threshold, patterns with scores below this threshold are considered dissimilar.
    \item Patterns with consistently low similarity scores across the board are identified as ``no-op" outliers.
\end{enumerate} 

As for ad-hoc assignment, we create a provisional classification system to identify ``no-op" outliers.
This temporary framework allows us to segregate data that does not fit into predefined categories.

For Outlier Efficient model implement as attention-like layer for deep learning, the similarity measurement is automatically done by learning.
Thus, it identifies outliers without extra effort. Patterns with small inner products with queries get almost zero attention probability, because of our retrieval dynamic design \eqref{eqn:retrieval_dyn}.

Explicitly, let $z:=(z_1,...,z_M)\in \mathbb{R}^M$.
By \eqref{eqn:lse1} and  \eqref{eqn:retrieval_dyn}, $\Softmax_1(z)$ automatically assigns $\sim 0$ output to $z_i\sim 0$ for all $ i\in[M]$ without requiring other $z_{j\neq i}$ to be super huge, by associating them to zero-point energy state (no-op memories).
Here $z$ is learned according to \eqref{eqn:OutEffHop} when OutEffHop is used as a learning layer.
Hence, it's clear the outlier identification is done automatically through learning.
}

Consider an example involving a negligible input vector in the attention mechanism:
 $n$ = [-10, -10, -10].
Upon passing $n$ through the $\Softmax$ function, it yields relatively large weights:
\begin{align*}
\Softmax(n) \approx [0.33,\ 0.33,\ 0.33].
\end{align*}
To achieve a \textbf{no-update}, the attention mechanism allocates increasing attention to low-information tokens, causing the probability of other tokens to approach zero (See \cref{background} for details).
For instance, if the first element in $n$ represents a low-information token (e.g., [SEP]), the input vector might transform into
\begin{align*}
n' = [100, -10, -10].
\end{align*}
This transformation causes the weights of all but the first token to converge to zero:
\begin{align*}
\Softmax(n') \approx [0.99,\ 2\times10^{-48},\ 2\times10^{-48}].
\end{align*}
This procedure requires the wide range of input vector, leading the emergence of outliers.
However, when $n$ is processed by $\Softmax_1$, the result is as follows:
\begin{align*}
\Softmax_1(n)\approx [5\times10^{-5}, 5\times10^{-5}, 5\times10^{-5} ]
\end{align*}
In this case, all vector values diminish to a level close to zero.
Consequently, the attention head does not need to assign a higher probability mass to specific tokens, resulting in a reduction in the memory space for the vector. 
Therefore, by construction, $\Softmax_1$ is outlier-robust.

\clearpage
\section{Proofs of Main Text}
\label{sec:proofs}

\subsection{\cref{lem:retrieval_dyn}} \label{apdix: proof_retrieval_dyn}
\begin{proof}[Proof of \cref{lem:retrieval_dyn}.] 
    To show monotonic decreasing property of the energy (\ref{eqn:H_energy}), we first derive the outlier-efficient retrieval dynamics by utilizing the convex-concave procedure \cite{cccp} (CCCP). 
    The total energy $\calH(\bx)$ is split into convex term $\calH_1 \coloneqq \half\Braket{\bx, \bx}$ and concave term $\calH_2 \coloneqq -\lse_1(\beta, \Xi^\sT\bx)$.
    In addition, $\calH_1$ and $\calH_2$ are both differentiable
    by definition.
    Every iteration of CCCP applied on $\calH$ gives:
    \begin{align*}
    \underbrace{\nabla_{\bx} \calH_1(\bx_{t+1})}_{=\half \nabla_{\bx} \Braket{\bx_{t+1}, \bx_{t+1}}} 
    = -\nabla_{\bx} \underbrace{\calH_2(\bx_t)}_{=-\lse_1(\beta, \Xi^\sT\bx_t)},
    \end{align*}
    such that
    \begin{align*}
    \bx_{t+1} = \nabla_{\bx}  \lse_1(\beta, \Xi^\sT\bx_t).    
    \end{align*}

    To derive the gradient of $\lse_1(\beta, \Xi^\sT\bx_t)$ , 
    we set $\tau(\beta z_l) \coloneqq \sum_l^N \exp(\beta z_l)$.
    
    Then,
    \begin{align*}
    \nabla_{\bx} \lse_1\(\beta, \Xi^\sT \bx_t\) |_{\bx_t} 
     &= \nabla_{\bx_t} \(\beta^{-1} \log \{ \tau(\beta \Xi^\sT \bx_t) +1\}\)\\
     &= \beta^{-1}  \nabla_{\tau}\log (\tau +1) \cdot \nabla_{\bx_t}\tau(\beta \Xi^\sT \bx_t) \\
     &= \frac{1}{\tau(\beta \Xi^\sT \bx_t) + 1} \cdot \exp \(\beta {\Xi^\sT \bx_t }\) \cdot \Xi^\sT \\
     &= \Xi \cdot \frac{\exp \(\beta \Xi^\sT \bx_t \)}{1 +\sum \exp \(\beta \Xi^\sT \bx_t \)} \\ 
     &= \Xi \cdot \Softmax_1\(\beta \Xi^\sT \bx_t \) .
    \end{align*}
    Hence, we obtain
    \begin{align*}
    \bx_{t+1} = \nabla_{\bx}\lse_1(\beta \Xi^\sT\bx_t) = \Xi \cdot \Softmax_1\( \beta \Xi^\sT \bx_t\)
    \end{align*}
    Due to the concave design of $\calH_2$, we demonstrate that $\calH$ can be monotonically decreased by $\calT_{\text{OutEff}}(\bx)$ given by \eqref{eqn:retrieval_dyn}, following the proof in \citep[Appendix~E.2]{hu2023SparseHopfield}.
\end{proof}

\subsection{\cref{lemma:convergence_sparse}} \label{appdix:convergence_sparse}
With the monotonic decreasing property from \cref{lem:retrieval_dyn}, we prove \cref{lemma:convergence_sparse} following the same strategy as \citep[Lemma~3.3]{wu2023stanhop} and \citep[Lemma~2.2]{hu2023SparseHopfield}.

\subsection{\cref{thm:retrieval_error}} \label{appdix:retrieval_error}
\begin{proof}[Proof of \cref{thm:retrieval_error}]
    Let $\calT_{\text{original}}$ be the retrieval dynamics of the original modern Hopfield model \cite{ramsauer2020hopfield}, and $\left\| \calT_{\text{OutEff}}(\bx) - \bxi_{\mu}\right\|$ and $\left\| \calT_{\text{original}}(\bx) - \bxi_{\mu}\right \|$ be the retrieval error of outlier-efficient and modern Hopfield model, respectively.
    
    To prove $\left\| \calT_{\text{OutEff}}(\bx) - \bxi_{\mu}\right\|$ has tighter upper bound than $\left\| \calT_{\text{original}}(\bx) - \bxi_{\mu}\right \|$, we recall
    the upper bound on $[\Softmax(\beta \Xi^{\sT}\bx)]_\nu$ from \citep[Eqaution~C.37]{wu2023stanhop}:
    \begin{align*}
    [\Softmax(\beta \Xi^{\sT}\bx)]_\nu \leq \exp{-\beta \Tilde{\Delta}_\mu},
    \end{align*}
    where $\Tilde{\Delta}_\mu \coloneqq \Braket{\bxi_\mu, \bx} - \Max_{\mu,\nu \in[M]; \mu\neq \nu} \Braket{\bxi_\mu, \bxi_{\nu}}$.
    
    Since we observe the relation 
    \begin{align*}
    \[\Softmax_1\(\beta\Xi^\sT \bx\)\]_\nu = \( \sum\limits_{\mu=1}^{M} \[\Softmax_1\(\beta\Xi^\sT \bx\)\]_\mu
    \)\[\Softmax\(\beta\Xi^\sT \bx\)\]_\nu
    ,\end{align*}
    it holds
    \begin{align*}
     [\Softmax_1(\beta \Xi^{\sT}\bx)]_\nu \leq \exp{-\beta \Tilde{\Delta}_\mu} \cdot \gamma
    ,\end{align*}
    where $\gamma \coloneqq \sum\limits_{\mu=1}^{M} \[\Softmax_1\(\beta\Xi^\sT \bx\)\]_\mu $.
    Note that $0 <\gamma <1 $.
    
    For any $\beta>0$, there exist a $\delta>0$ such that $\gamma \coloneqq \exp{-\beta \delta}$.
    
    Also, recall the bound of $\left\| \calT_{\text{original}} - \bxi_\mu\right \|$ from \citep[Equation~C.41]{wu2023stanhop} :
    \begin{align}
    \label{eqn:proof_use0}
    \left\| \calT_{\text{original}} - \bxi_\mu\right \| 
    \leq 2 m(M-1)\exp{-\beta\Tilde{\Delta}_\mu}
    \leq 2 m(M-1)\exp{-\beta\(\Delta_{\mu} - 2mR\)}.
    \end{align}
    By \citep[Equation~C.39]{wu2023stanhop}, 
    we know
    $\Tilde{\Delta}_\mu \geq \Delta_\mu - 2mR$ and $R$ is the radius of the sphere $S_\mu$.
    Then we have
    \begin{align*}
     \left\| \calT_{\text{OutEff}} - \bxi_\mu\right\| \leq 
     2 m(M-1)\exp{-\beta\(\Delta_{\mu} - 2mR +\delta\)}
    .
    \end{align*}
    Comparing above with \eqref{eqn:proof_use0}, this complete the proof.
\end{proof}

\subsection{\cref{coro:small_retrie_error}}\label{appdix:coro_retrieval}
\begin{proof}[Proof of \cref{coro:small_retrie_error}.]
    We aim to establish the validity of the following inequality:
    \begin{align*} 
    \left\| \calT_{\text{OutEff}}(\bx) - \bxi_{\mu}\right\| \leq  \left\| \calT_{\text{original}}(\bx) - \bxi_{\mu}\right \|.
    \end{align*}
    It is equivalent to consider 
    \begin{align} \label{eqn:retrieval_error_sqr}
    \left\| \calT_{\text{OutEff}}(\bx) - \bxi_{\mu}\right\|^2 - \left\| \calT_{\text{original}}(\bx) - \bxi_{\mu}\right \|^2 \leq 0.
    \end{align}
    That is, 
    \begin{align}
    \label{eqn:proof_use1}
    \left\| \sum\limits_{\nu=1}^{M} \bxi_{\nu} \[\Softmax_1\(\beta\Xi^\sT \bx\)\]_\nu  - \bxi_{\mu}\right\|^2 - \left\| \sum\limits_{\nu=1}^{M} \bxi_{\nu} \[\Softmax\(\beta\Xi^\sT \bx\)\]_\nu - \bxi_{\mu}\right \|^2 \leq 0 
    .\end{align}
    
    Let $\gamma \coloneqq \sum\limits_{\mu=1}^{M} \[\Softmax_1\(\beta\Xi^\sT \bx\)\]_\mu $ ($0<\gamma<1$).

    For ease of presentation, 
    we set  $\bv_1 \coloneqq \sum\limits_{\nu=1}^{M} \bxi_{\nu} \[\Softmax\(\beta\Xi^\sT \bx\)\]_\nu $, 
    $\bv_2 \coloneqq \sum\limits_{\nu=1}^{M} \bxi_{\nu} \[\Softmax_1\(\beta\Xi^\sT \bx\)\]_\nu =  \gamma \bv_1$ and $w\coloneqq\bxi_{\mu}$.
    
    \eqref{eqn:proof_use1} becomes
    \begin{align*}
    \|\bv_2- w\|^2 - \|\bv_1 - w\|^2 \leq 0,
    \end{align*}
    expanding both terms
    \begin{align*}
    &\bv_2^2 - 2 \bv_2 \cdot w + w^2 - \bv_1^2 + 2\bv_1 \cdot w -w^2 \leq 0 ,
    \end{align*}
    simplifying the expression
    \begin{align*}
    &(\gamma^2-1) \bv_1^2 - 2 (\gamma-1)\bv_1\cdot w  \leq 0 ,
    \end{align*}
    since vectors are normalized
    \begin{align*}
    &(\gamma^2-1) \|\bv_1\| - 2 (\gamma-1) \|w\| \cos{\alpha} 
    \leq 0 ,
    \end{align*}
    and rearranging the terms
    \begin{align*}
    &(\gamma+1)  \|\bv_1\| - 2 \|w\| \cos{\alpha} \geq 0,
    \end{align*}
    where $\cos{\alpha}$ quantifies the overlap between $\bv_1$ and $w$.
    
    If all the memories and queries are normalized, \eqref{eqn:retrieval_error_sqr} holds when
    \begin{align*}
    \frac{\gamma+1}{2} \geq \cos{\alpha}
    .\end{align*}
    As memory patterns and queries exhibit smaller overlap at the beginning of the retrieval process, the proposed model experiences smaller retrieval errors than its original counterpart during the initial phase of memory retrieval.
\end{proof}

\subsection{\cref{lemma:capacity}} \label{appdix:capacity}
\begin{lemma}[Memory Capacity Lower Bound, Formal]
\label{lem:memory_lowbound}
Suppose the probability of successfully storing and retrieving memory pattern is given by $1-p$.
The number of memory patterns sampled from a sphere of radius $m$ that the Outlier-Efficient Hopfield model can store and retrieve has a lower bound: $M \geq \sqrt{p} C^{\frac{d-1}{4}}$, where $C$ is the solution for $C = b/W_0 (\exp{a+\ln b})$ with $W_0(\cdot)$ being the principal branch of Lambert $W$ function \cite{olver2010nist}, 
$ a \coloneqq {\scriptstyle 4} /{ \scriptstyle (d - 1)}\left\{ \ln[{\scriptstyle 2m^2(\sqrt{p}-1)}/{\scriptstyle R}] + 1 - {\scriptstyle \delta} / {\scriptstyle (2\beta mR)} \right\}$
and $b \coloneqq {\scriptstyle 4m^2 \beta}/ {\scriptstyle 5(d-1)}$.
For all $\beta$, we have larger memory capacity lower bound compared to the original modern Hopfield model \cite{ramsauer2020hopfield}: $M\geq M_{\text{original}}$
\end{lemma}
To prove it, we first derive the well-separation condition for the outlier-efficient modern Hopfield model.
\begin{lemma}[Modified from Lemma C.3 of \cite{wu2023stanhop}]
\label{lem:well_sep}
Let $\gamma \coloneqq \sum\limits_{\mu=1}^{M} \[\Softmax_1\(\beta\Xi^\sT \bx\)\]_\mu $ and $1 >\gamma > 0$.
For any $\beta>0$, there exist a $\delta>0$ such that $\gamma \coloneqq \exp{-\beta \delta}$.
Then, the well-seperation condition can be formulated as:
\begin{align*}
\Delta_\mu \geq \frac{1}{\beta}\ln{\(\frac{2(M-1)m}{R}\)}+2mR-\delta.
\end{align*}
\end{lemma}
\begin{proof}

    From \cref{apdix: proof_retrieval_dyn} we obtain the result
    \begin{align*}
    \left\| \calT_{\text{OutEff}} - \bxi_\mu\right\| 
    &\leq 2 m(M-1)\exp{-\beta\(\Delta_{\mu} - 2mR +\delta\)}
    \end{align*}
    Therefore, for $\calT_{\text{OutEff}}$ to be mapping $\calT_{\text{OutEff}}:S_\mu \rightarrow S_\mu$, it is sufficient to obtain
    \begin{align*}
    2(M-1)
    \exp{-\beta\(\Delta_\mu - 2mR+\delta\)} m \leq R.
    \end{align*}
    This leads to the separation condition for the proposed Outlier-Efficient Modern Hopfield Model
    \begin{align} \label{eqn:Delta_mu}
    \Delta_\mu \geq \frac{1}{\beta}\ln{\(\frac{2(M-1)m}{R}\)}+2mR-\delta.
    \end{align}
    Given that \eqref{eqn:Delta_mu} possesses a stricter lower bound compared to its original counterpart \citep[Equation~(300)]{ramsauer2020hopfield},we complete the proof following the similar approach in \citep[Lemma~3.4]{wu2023stanhop}.
\end{proof}

 {
 \begin{lemma}\label{lem:abc}[\cite{ramsauer2020hopfield}]
      If the identity
      \begin{align*}
     ac + c \ln c -b = 0,
     \end{align*}
     holds for all real numbers $a, \ b \in \mathbb{R}$, then $c$ takes a solution:
     \begin{align*}
     c = \frac{b}{W_0(\exp(a + \ln b))}.
     \end{align*}
 \end{lemma}
  \begin{proof}
      By looking at the proof in \cite{wu2023stanhop}.
 \end{proof}
 Then we start our formal proof of \cref{lemma:capacity}.
 \begin{proof}
     Since $\Delta_{\text{min}} = \min\limits_{1 \leq \mu\leq M}\Delta_\mu$, we get
     \begin{align*}
     \Delta_{\text{min}} \geq \frac{1}{\beta}\ln{\(\frac{2(M-1)m}{R}\)}+2mR-\delta.
     \end{align*}
     Following the proof in (\cite{wu2023stanhop}, Appendix Theorem A5), we obtain
     \begin{align*}
     a \coloneqq \frac{4}{(d - 1)}\left\{ \ln\[\frac{ 2m^2(\sqrt{p}-1)}{R}\] + 1 - \frac{\delta}{(2\beta mR)}  \right\},\quad
     b \coloneqq \frac{4m^2 \beta}{5(d-1)}.
     \end{align*}
     By \cref{lem:abc}, $C$ can be expressed as
     \begin{align*}
     C = \frac{b}{W(\exp{a + \ln{b}})}.
     \end{align*}
     We expressed the original counterpart of $a$ and $b$ as
     \begin{align*}
     \Tilde{a} \coloneqq \frac{4}{(d - 1)}\left\{ \ln\[\frac{ 2m^2(\sqrt{p}-1)}{R}\] + 1\right\}
     , \ \ \Tilde{b} = b.
     \end{align*}
     Since 
     \begin{align*}
     \Tilde{a} > a
     \end{align*}
     and 
     \begin{align*}
     \Tilde{C} = \frac{b}{W(\exp{\Tilde{a} + \ln{b}})} < \frac{b}{W(\exp{a + \ln{b}})} = C,
    \end{align*}
     we arrive at 
     \begin{align*}
     M_{\text{original}} = \sqrt{p}\Tilde{C}^{\frac{d-1}{4}} < \sqrt{p}C^{\frac{d-1}{4}} = M.
     \end{align*}
     This completes the proof.
\end{proof}
}

\clearpage
\subsection{\cref{lemma:gen_OutEffHop}}
\label{appendix:theorem_gen_outeffhop}
To bound the generalization error of Outlier-Efficient Modern Hopfield, we utilize the generalization bound via covering number \cite{dudley1978central,edelman2022inductive,liang2016cs229t,zhang2023mathematical}.
\begin{definition}[Covering Number]
    For a given class of vector-valued functions $\mathcal{F}$, the covering number $\mathcal{N}_{\infty}\left(\mathcal{F} ; \varepsilon ;\left\{z^{(i)}\right\}_{i=1}^{m} ;\|\cdot\|\right)$ is the smallest size of a collection (a cover) $\mathcal{C} \subset \mathcal{F}$ such that $\forall f \in \mathcal{F}, \exists \widehat{f} \in \mathcal{C}$ satisfying
\begin{align*}
\max _{i}\left\|f\left(z^{(i)}\right)-\widehat{f}\left(z^{(i)}\right)\right\| \leq \varepsilon .
\end{align*}
Also, define
\begin{align*}
\mathcal{N}_{\infty}(\mathcal{F}, \varepsilon, m,\|\cdot\|)
\quad \coloneqq
  \sup _{z^{(1)} \ldots z^{(m)}} \mathcal{N}_{\infty}\left(\mathcal{F} ; \varepsilon ; z^{(1)}, \ldots, z^{(m)},\|\cdot\|\right).
\end{align*}
\end{definition}
\begin{lemma}[Generalization Bound via Covering Number \cite{liang2016cs229t,zhang2023mathematical}]
\label{lemma:gen_thru_cn}
Suppose $\mathcal{F}$ is a class of bounded functions, and $\log \mathcal{N}_{\infty}\left(\mathcal{F} ; \varepsilon ; x^{(1)}, \ldots, x^{(m)}\right) \leq C_{\mathcal{F}} / \varepsilon^{2}$ for all $x^{(1)}, \ldots, x^{(m)} \in \mathcal{X}^{m}$. Then for any $\delta>0$, with probability at least $1-\delta$, simultaneously for all $f \in \mathcal{F}$, the generalization error $\varepsilon_{\text {gen }}$ satisfies
\begin{align*}
\varepsilon_{\mathrm{gen}}(f) \leq \tilde{O}\left(\sqrt{\frac{C_{\mathcal{F}}}{m}}+\sqrt{\frac{\log (1 / \delta)}{m}}\right) .
\end{align*}
\end{lemma}

We start by proving the Lipschitzness of $\Softmax_1$. We first introduce \cref{lemma:A.6_4.0} from \cite{edelman2022inductive}.
\begin{lemma}[Lemma A.6. of \cite{edelman2022inductive}] \label{lemma:A.6_4.0}
    Consider a function $f: \mathbb{R}^{d} \rightarrow \mathbb{R}^{d}$ such that the Jacobian $\calJ_f\coloneqq\grad f$ of the function satisfies $\|\calJ_f(\bv)\|_{1,1} \leq$ $c_{f}$ for all $\bv \in \mathbb{R}^{d}$, then for any vectors $\bv_{1}, \bv_{2} \in \mathbb{R}^{d}$,
\begin{align*}
\left\|f\left(\bv_{1}\right)-f\left(\bv_{2}\right)\right\|_{1} \leq c_{f}\left\|\bv_{1}-\bv_{2}\right\|_{\infty} .
\end{align*}
\end{lemma}
With \cref{lemma:A.6_4.0}, we obtain the Lipschitzness of $\Softmax_1$.
\begin{lemma}[Lipschitzness of $\Softmax_1$] \label{lemma:lip_softmax1_4.0}
For vectors $\bx_{1}, \bx_{2} \in \mathbb{R}^{d},$
\begin{align*}
\left\| \Softmax_1(\bx_1) - \Softmax_1(\bx_2) \right\|_{1} \leq 2\left\|\bx_{1}- \bx_{2}\right\|_{\infty} .
\end{align*}
\end{lemma}
\begin{proof}[Proof of \cref{lemma:lip_softmax1_4.0}]
    We prove that $\Softmax_1$ satisfies $\|\calJ_f(\theta)\|_{1,1} \leq c_{f}$, and use \cref{lemma:A.6_4.0} to obtain the Lipschitzness. 

    Let
    $\calJ_{\Softmax_1}(\bx)\coloneqq \grad_\bx\Softmax_1(\bx)$.

    For any $\bx\in\R^d$, we first denote the elements of $\calJ_{\Softmax_1}(\bx)$ as
    \begin{align*}
    \frac{\partial \Softmax_1(\bx)_i}{\partial x_j}, \text{ for } i,j \in[d] .
    \end{align*}
    Observe that for $i=j$:
    \begin{align*}
    & \frac{\partial \Softmax_1(\bx)_i}{\partial x_j}= 
    \Softmax_1(\bx)_i - \Softmax_1(\bx)_i\Softmax_1(\bx)_j,
    \end{align*}
    and for $i\neq j$:
    \begin{align*}
    \frac{\partial \Softmax_1(\bx)_i}{\partial x_j}=  - \Softmax_1(\bx)_i\Softmax_1(\bx)_j.
    \end{align*}
    Therefore, we have 
    \begin{align*}
    \|\calJ_{\Softmax_1}(\bx) \|_{1,1}  
    &  = \abs{\sum_{i,j=1}^d \Softmax_1(\bx)_i \one(i=j) - \Softmax_1(\bx)_i \Softmax_1(\bx)_j} 
    \\ \nonumber & = 
    \abs{\sum_{i,j=1}^d \Softmax_1(\bx)_i \left( \one(i=j) -  \Softmax_1(\bx)_j \right) }
    \\ \nonumber & <
    2 \sum_i^d \Softmax_1(\bx)_i \left( 1 - \Softmax_1(\bx)_i \right)
    \\ \nonumber & \le
    2.
    \end{align*}
    Finally, by \cref{lemma:A.6_4.0}, we have
    \begin{align*}
\left\|\Softmax_1(\bx_1)-\Softmax_1(\bx_2)\right\|_{1} \leq 2\left\| \bx_{1}- \bx_{2} \right\|_{\infty} .
\end{align*}
This completes the proof.
\end{proof}

Next, we prove the Lipschitzness of $f_{\text{hop}}$ in parameter.
\begin{lemma}[Lipschitzness of $f_{\text{hop}}$] 
\label{lemma:lip_fhop_parameter}
    For any $ \bW_{K}, \bW'_{K} \in \mathcal{W}_K$, $ \tilde{\bW}_{V}, \tilde{\bW}'_{V} \in \tilde{\mathcal{W}}_V$, $\tau \in [T]$:
    \begin{align*} 
    &~ \left\| f_{\text{hop}}(\bY,\bq_\tau;\bW_{K}, \tilde{\bW}_{V}) - f_{\text{hop}}(\bY,\bq_\tau;\bW'_{K}, \tilde{\bW}'_{V}) \right\|    
    \\  
    \leq &~   2 B_V B_Y \left\| \beta \bY \bW_K \bq_\tau - \beta \bY \bW'_K \bq_\tau \right\|_{\infty} 
    + \left\|  \left( \bY \tilde{\bW}_V \right)^\sT - \left( \bY \tilde{\bW}_V' \right)^\sT  \right\|_{2,\infty} .
    \end{align*}
\end{lemma}
\begin{proof}[Proof of \cref{lemma:lip_fhop_parameter}]
    \begin{align*}
    &~  \left\| f_{\text{hop}}(\bY,\bq_\tau;\bW_{K}, \tilde{\bW}_{V}) - f_{\text{hop}}(\bY,\bq_\tau;\bW'_{K}, \tilde{\bW}'_{V}) \right\|  
    \\ 
    \nonumber 
    = &~  
    \left\| \tilde{\bW}_V^\sT \bY^\sT  \Softmax_1 \( \beta \bY \bW_K \bq_\tau \)
    - 
   \tilde{\bW}_V'^\sT \bY^\sT  \Softmax_1 \( \beta \bY \bW_K' \bq_\tau \) \right\| 
   \\ \nonumber
    =&~   
     \norm{\tilde{\bW}_V^\sT \bY^\sT \( \Softmax_1\left( \beta \bY \bW_K \bq_\tau \right) - \Softmax_1\left(\beta \bY \bW_K' \bq_\tau \right) \)  
    + \left( \tilde{\bW}_V^\sT \bY^\sT - \tilde{\bW}_V'^\sT \bY^\sT \right) \Softmax_1\left(\beta \bY \bW_K' \bq_\tau \right)} 
     \\
     \leq &~   \left\| \tilde{\bW}_V^\sT \bY^\sT \( \Softmax_1\left( \beta \bY \bW_K \bq_\tau \right) - \Softmax_1\left(\beta \bY \bW_K' \bq_\tau \right) \) \right\| 
        + \left\| \left( \tilde{\bW}_V^\sT \bY^\sT - \tilde{\bW}_V'^\sT \bY^\sT \right) \Softmax_1\left(\beta \bY \bW_K' \bq_\tau \right)  \right\|
        \annot{By triangle inequality}
        \\
    \leq &~   \left\|  \tilde{\bW}_V^\sT \bY^\sT  \right\|_{2,\infty}   \left\|  \Softmax_1\left( \beta \bY \bW_K \bq_\tau \right) - \Softmax_1\left(\beta \bY \bW_K' \bq_\tau \right)    \right\|_1 
    \\
    &\hspace{12em}+ \left\|  \tilde{\bW}_V^\sT \bY^\sT - \tilde{\bW}_V'^\sT \bY^\sT \right\|_{2,\infty} \left\|  \Softmax_1\left(\beta \bY \bW_K' \bq_\tau \right) \right\|_1
    \annot{By $\norm{\bA \bx} \leq  \norm{\bA}_{2,\infty} \norm{\bx }_1$}
    \\
    \leq &~   
    \left\|  \tilde{\bW}_V^\sT \right\|_2 \left\| \bY^\sT  \right\|_{2,\infty}   \left\|  \Softmax_1\left( \beta \bY \bW_K \bq_\tau \right) - \Softmax_1\left(\beta \bY \bW_K' \bq_\tau \right)    \right\|_1 
    \\
    &\hspace{12em}+ 
    \left\|  \tilde{\bW}_V^\sT \bY^\sT - \tilde{\bW}_V'^\sT \bY^\sT \right\|_{2,\infty} \left\|  \Softmax_1\left(\beta \bY \bW_K' \bq_\tau \right) \right\|_1 
    \annot{By $\|\bP \bQ\|_{2, \infty} \leq\|\bP\|_{2}\|\bQ\|_{2, \infty}$}
    \\
    \leq &~   
    2 B_V B_Y \norm{ \beta \bY \bW_K \bq_\tau - \beta \bY \bW'_K \bq_\tau}_{\infty} +
    \left\|  \tilde{\bW}_V^\sT \bY^\sT - \tilde{\bW}_V'^\sT \bY^\sT \right\|_{2,\infty} \left\|  \Softmax_1\left(\beta \bY \bW_K' \bq_\tau \right) \right\|_1 
    \annot{By \cref{assumption:norm_bounds}-\hyperref[item:A4]{(A4)} and  \cref{lemma:lip_softmax1_4.0}}
    \\
    \leq &~ 
    2 B_V B_Y \left\| \beta \bY \bW_K \bq_\tau - \beta \bY \bW'_K \bq_\tau \right\|_{\infty} 
    + \left\|  \left( \bY \tilde{\bW}_V \right)^\sT - \left( \bY \tilde{\bW}_V' \right)^\sT  \right\|_{2,\infty}. 
    \end{align*}
\end{proof}
Next, with  \cref{lemma:lip_fhop_parameter}, we construct a covering number bound for a Modern Hopfield model function class using the covering number of its composing functions. 
We write the composing functions as $f_K: \mathbb{R}^{M \times a} \times \mathbb{R}^{d} \rightarrow \mathbb{R}^{M}$ as:
\begin{align*} 
      f_{K}(\bY,\bq;\bW_{K})
     = 
    \beta \bY \bW_{K} \bq,
\end{align*} 
and 
$f_V: \mathbb{R}^{M \times a} \rightarrow \mathbb{R}^{d \times M}$ as:
\begin{align*} 
      f_{V}(\bY; \tilde{\bW}_{V})
     = 
    \( \bY
\tilde{\bW}_V \)^\sT.
\end{align*}
With $f_k$ and $f_V$, we prove that the covering number of $\mathcal{F}_{\text {hop }}$ is bounded as below.
\begin{lemma} \label{lemma:cn_hop_K_V}
    Under the \cref{assumption:norm_bounds}, for any $\alpha \in [0,1]$ the covering number of $\mathcal{F}_{\text {hop }}$ satisfies
    \begin{align*}
&~  \log \mathcal{N}_{\infty}\left(\mathcal{F}_{\text {hop }} ; \varepsilon ;\left\{\left( \bY^{(i)}, \bq_\tau^{(i)} \right)\right\}^{i \in [N], }_{\tau \in [T] } ;\|\cdot\|_{2}\right) 
\\ 
\leq &~
\log \mathcal{N}_{\infty}\left( \mathcal{F}_{K} ; \varepsilon_{K} ;\left\{\left(\by_{t}^{(i)}, \bq_\tau^{(i)} \right)\right\}_{t \in[M], \tau \in [T]}^{i \in [N]} \right)
 +
\log \mathcal{N}_{\infty}\left( \mathcal{F}_{V} ; \varepsilon_{V} ;\left\{\by_{t}^{(i)}\right\}_{t \in[M]}^{i \in[N]} ;\|\cdot\|_{2} \right),
\nonumber
\end{align*}
where 
$\mathcal{F}_K=\left\{(\by, \bq) \mapsto \beta \by^\sT \bW_{K} \bq: \bW_{K}\in \mathcal{W}_{K} \right\}$ 
and 
$\mathcal{F}_{V}=\left\{ \by^\sT \mapsto \( \by^\top
\tilde{\bW}_V \)^\sT : \tilde{\bW}_V \in \tilde{\mathcal{W}}_{V} \right\}$.
\end{lemma}
\begin{proof}[Proof of \cref{lemma:cn_hop_K_V}]
We prove that for each $\varepsilon > 0$ and input sample $(\bY^{(i)}, \bQ^{(i)})$ for all $i \in[N]$, there exists a cover $\mathcal{C}_\text{hop}$ for $\mathcal{F}_\text{hop}$.
    From \cref{lemma:lip_fhop_parameter}, we see that 
    \begin{align*}  
    &~ \left\| f_{\text{hop}}(\bY,\bq_\tau;\bW_{K}, \tilde{\bW}_{V}) - f_{\text{hop}}(\bY,\bq_\tau;\bW'_{K}, \tilde{\bW}'_{V}) \right\|   
    \\
    \leq &~
     2 B_V B_Y \left\|f_K \left( \bY^{(i)}, \bq_\tau^{(i)} ; \bW_{K} \right) - f_K \left( \bY^{(i)}, \bq_\tau^{(i)} ; \bW_{K}' \right) \right\|_{\infty}  
     + 
    \left\|  f_V \left( \bY^{(i)} ; \tilde{\bW}_V \right) -  f_V \left( \bY^{(i)} ; \tilde{\bW}_V' \right)  \right\|_{2,\infty} . 
    \end{align*}
    With the property of $\ell_{\infty}$-norm, we have
    \begin{align*}
\max _{i \in[N]}\left\| f_K \left( \bY^{(i)}, \bq_\tau^{(i)} ; \bW_{K} \right) - f_K \left( \bY^{(i)}, \bq_\tau^{(i)} ; \bW_{K}' \right)
\right\|_{\infty}
=
\max _{i \in[N], t \in[M]}
\abs{f_K \left( \by_t^{(i)}, \bq_\tau^{(i)} ; \bW_{K} \right) - f_K \left( \by_t^{(i)}, \bq_\tau^{(i)} ; \bW_{K}' \right)}.
\end{align*}
Also, with the property of $\ell_{2,\infty}$-norm, 
\begin{align*}
\max _{i \in[N]} \left\|  
f_V \left( \bY^{(i)} ; \tilde{\bW}_V \right) -  f_V \left( \bY^{(i)} ; \tilde{\bW}_V' \right)
\right\|_{2,\infty} 
=\max _{i \in[N], t \in[M]} \left\| 
f_V \left( \by_t^{(i)} ; \tilde{\bW}_V \right) -  f_V \left( \by_t^{(i)} ; \tilde{\bW}_V' \right) 
\right\| .
\end{align*}
Now, we let $\mathcal{C}_{K}$ (a set of $\bW_K$) be the $\varepsilon_{K}$-cover for $\mathcal{F}_{K}$ over inputs $\left\{\left(\by_{t}^{(i)}, \bq_\tau^{(i)} \right)\right\}_{t \in[M], \tau \in [T]}^{i \in [N]}$ of size
\begin{align*}
\mathcal{N}_{\infty}\left(\mathcal{F}_{K} ; \varepsilon_{K} ;\left\{\left(\by_{t}^{(i)}, \bq_\tau^{(i)} \right)\right\}_{t \in[M], \tau \in [T]}^{i \in [N]} \right) .
\end{align*}
Also, let $\mathcal{C}_{V}$ (a set of $\tilde{\bW}_{V}$) be the $\varepsilon_{V}$-cover for $\mathcal{F}_{V}$ over inputs $\left\{y_{t}^{(i)}\right\}_{t \in[M]}^{i \in [N]}$ of size
\begin{align*}
\mathcal{N}_{\infty}\left( \mathcal{F}_{V} ; \varepsilon_{V} ;\left\{\by_{t}^{(i)}\right\}_{t \in[M]}^{i \in[N]} ;\|\cdot\|_{2}\right) .
\end{align*}
We now construct the cover for $\mathcal{F}_{\text{hop}}$. First Set 
\begin{align*}
 \mathcal{C}_{\text {hop }}
 = \left\{f_{\text {hop }}\left(\bY^{(i)},\bq_\tau^{(i)}; \bW_{K}' \tilde{\bW}_{V}' \right)_{\tau \in [T]}^{i \in [N]}:  \bW_{K}' \in \mathcal{C}_{K} , \tilde{\bW}_{V}' \in \mathcal{C}_{V}\right\} .
\end{align*}
Then for any $\bW_{K}\in \mathcal{\bW}_{K}, \tilde{\bW}'_{V} \in \tilde{\mathcal{W}}_V$,  there exists $\bW_{K}' \in \mathcal{C}_{\text {hop}}, \tilde{\bW}_{V}' \in \mathcal{C}_{V}$ (using \cref{lemma:lip_fhop_parameter}):
\begin{align*}
\left\| f_{\text{hop}}(\bY,\bq_\tau;\bW_{K}, \tilde{\bW}_{V}) - f_{\text{hop}}(\bY,\bq_\tau;\bW'_{K}, \tilde{\bW}'_{V}) \right\|  
 \leq 
 2 B_V B_Y \varepsilon_{K}+ \varepsilon_{V} .
\end{align*}
The size of the cover $\mathcal{C}_{\text {hop }}$ we have constructed is,
\begin{align*} 
&~  \log  \abs{\mathcal{C}_{\text {hop }}}
\\  
= &~
\log \abs{\mathcal{C}_{K}} + \log \abs{\mathcal{C}_{V}} \\ 
= &~
 \log \mathcal{N}_{\infty}\left(\mathcal{F}_{K} ; \varepsilon_{K} ;\left\{\left(\by_{t}^{(i)}, \bq_\tau^{(i)} \right)\right\}_{t \in[M], \tau \in [T]}^{i \in [N] } \right)
+
\log \mathcal{N}_{\infty}\left( \mathcal{F}_{V} ; \varepsilon_{V} ;\left\{\by_{t}^{(i)}\right\}_{t \in[M]}^{i \in[N]} ;\|\cdot\|_{2} \right),
\end{align*}
where $\varepsilon=2 \varepsilon_{\text{Score}}+ \varepsilon_{K}$  for $\mathcal{C}_{\text {hop }}$.
\end{proof}
Next, we introduce a useful lemma for completing the proof of \cref{lemma:gen_OutEffHop}.
\begin{lemma} \label{lemma_a.8}
    (Lemma A.8 of \cite{edelman2022inductive}) For $\alpha_{i}, \beta_{i} \geq 0$, the solution to the following optimization
\begin{align*}
&\min _{x_{1}, \ldots, x_{n}} \sum_{i=1}^{n} \frac{\alpha_{i}}{x_{i}^{2}} 
\quad
\text { subject to } \sum_{i=1}^{n} \beta_{i} x_{i}=C,
\end{align*}
is $\nicefrac{\gamma^{3}}{C^{2}}$ and is achieved at $x_{i}=\nicefrac{C}{\gamma}\left(\nicefrac{\alpha_{i}}{\beta_{i}}\right)^{1 / 3}$ where $\gamma=\sum_{i=1}^{n} \alpha_{i}^{1 / 3} \beta_{i}^{2/3}$.
\end{lemma}
\begin{proof}[Proof of \cref{lemma_a.8}]
    The proof follows by a standard Lagrangian analysis. 
    Let $f(x)$ be the objective function and $g(x)$ be the constraint function. 
    With Lagrange multiplier, we have
\begin{align*}
    \grad f \left( x \right) - \lambda \grad g \left( x \right) = 0.
\end{align*}
By plugging in $f$ and $g$ we have
\begin{align}
\label{Lagrangemultiply}
    x_i = -\(\frac{2\alpha_i}{\lambda \beta_i}\)^{\frac{1}{3}},
\end{align}
for all $i \in [n]$. 
In addition, we get $\lambda$ by plugging \eqref{Lagrangemultiply} into the constraint $g$:
\begin{align*}
    \lambda = \frac{\sum_{i=1}^n (2a_i)^{\frac{1}{3}}\beta_i^{\frac{2}{3}}}{C}.
\end{align*}
\end{proof}
To bound the covering number of $\mathcal{F}_K, \mathcal{F}_V$, we introduce the covering number bound for a linear function class:
\begin{lemma}[Covering Number Bound for Linear Function Class, Lemma 4.6 of \cite{edelman2022inductive}] \label{lemma:lin_cover_4.0}
Let $\mathcal{W}:\left\{W \in \mathbb{R}^{d_{1} \times d_{2}}:\left\|W^{\top}\right\|_{2,1} \leq B_{W}\right\}$, and consider the function class $\mathcal{F}:\{x \mapsto W x: W \in$ $\mathcal{W}\}$. For any $\varepsilon>0$ and $x^{(1)}, \ldots, x^{(N)} \in \mathbb{R}^{d_{2}}$ satisfying $\forall i \in[N],\left\|x^{(i)}\right\| \leq B_{X}$,
    \begin{align*}
    \log \mathcal{N}_{\infty}\left(\mathcal{F} ; \varepsilon ; x^{(1)}, \ldots, x^{(N)} ;\|\cdot\|_{2}\right) \lesssim \frac{\left(B_{X} B_{W}\right)^{2}}{\varepsilon^{2}} \log \left(d_{1} N\right) .
    \end{align*}
\end{lemma}
We now obtain the covering number bound of a Modern Hopfield Model explicitly by bounding the two function classes $\mathcal{F}_{V}$ and $\mathcal{F}_{K}$ using \cref{lemma:lin_cover_4.0}.
\begin{lemma}[Covering Number Bound of Outlier-Efficient Hopfield Layer] \label{lemma:hop_cn}
    \begin{align*}
     &~\log \mathcal{N}_{\infty}\left(\mathcal{F}_{\text {hop }} ; \varepsilon ;\left\{\left( \bY^{(i)}, \bQ^{(i)} \right)\right\}^{i \in [N]} ;\|\cdot\|_{2, \infty}\right) 
    \\  \nonumber 
    \leq &~
    \varepsilon^{-2} \left( \left(  4 B_V^2 B_Y^2 \left( B_{\beta K}^{2,1} \right)^{2} \log \left(d N M\right)
    \right)^\frac{1}{3} 
    +
    \left(  \left( B_V^{2,1} \right)^2 \log(d N M) \right)^\frac{1}{3}
    \right)^3,
\end{align*}
where $B_{\beta K}^{2,1}=\beta B_K^{2,1}$.
\end{lemma}
\begin{proof}[Proof of \cref{lemma:hop_cn}]
    First observe that since $\| \bq_\tau \| \leq 1$ (\cref{assumption:norm_bounds}-\hyperref[item:A1]{(A1)}), we have 
    \begin{align*}
    \abs{\beta \by_t^\sT \bW_K \bq_\tau - \beta \by_t^\sT \bW'_K \bq_\tau} 
    \leq 
    \left\| 
    \beta \by_t^\sT \bW_K  - \beta \by_t^\sT \bW'_K   
    \right\| .  
    \end{align*}
    We define the right hand side as (taking transpose to make it a column vector):
    \begin{align*}
    \hat{\mathcal{F}}_{K}:=\left\{y_t \mapsto   \bW_{\beta K}^\top y_t : \left\| \bW_{\beta K}\right\|_{2,1} \leq B_{\beta K}^{2,1}\right\},
    \end{align*}
    where $\bW_{\beta K}:=\beta  \bW_{K}$ and $B_{\beta K}^{2,1} = \beta B_{K}^{2,1}$.
    
    Since the covering number of $\mathcal{F}_{K}$ is at most the covering number of $\hat{\mathcal{F}}_{K}$, instead of discussing the covering number bound of $\mathcal{F}_{K}$ , we focus on $\hat{\mathcal{F}}_{K}$. 
    
    Now, by \cref{lemma:lin_cover_4.0}, \cref{assumption:norm_bounds}-\hyperref[item:A3]{(A3)} and
\cref{assumption:norm_bounds}-\hyperref[item:A4]{(A4)} we have
    \begin{align} \label{eq_score_cover_num_4}
     \log \mathcal{N}_{\infty}\left( \mathcal{F}_{K} ; \varepsilon_{K} ;\left\{\left(\by_{t}^{(i)}, \bq_\tau^{(i)} \right)\right\}_{t \in[M], \tau \in [T]}^{i \in [N]} \right) 
      \leq
    & \log \mathcal{N}_{\infty}\left(\hat{\mathcal{F}}_{K} ; \varepsilon_{K} ;\left\{\left( \by_{t}^{(i)} \right)\right\}_{t \in[M]}^{i \in[N]}\right) 
    \\ \nonumber   \lesssim
    & \frac{ \left( B_{\beta K}^{2,1} \right)^{2}}{\varepsilon_{K}^{2}} \log \left(d N M\right),    
    \end{align}
    and 
    \begin{align} \label{eq_K_cover_num_4}
    & \log \mathcal{N}_{\infty}\left(\mathcal{F}_{V} ; \varepsilon_V ;\left\{\by_{t}^{(i)}\right\}_{t \in[M]}^{i \in[N]} ;\|\cdot\|_{2}\right)
    \lesssim
    \frac{ \left( B_{V}^{2,1} \right)^{2}}{\varepsilon_{V}^{2}} \log \left(d  N M\right).  
    \end{align}
    Next, we find the optimal $\varepsilon_{K}$ and $\varepsilon_{V}$ to minimize the sum of \eqref{eq_score_cover_num_4} and \eqref{eq_K_cover_num_4}, subject to
    \begin{align*}
    2 B_V B_Y \varepsilon_{K}+ \varepsilon_{V} = \varepsilon .
    \end{align*}
    By \cref{lemma_a.8}, the optimal bound is
\begin{align*}
    &~\log \mathcal{N}_{\infty}\left(\mathcal{F}_{\text {hop }} ; \varepsilon ;\left\{\left( \bY^{(i)}, \bQ^{(i)} \right)\right\}^{i \in [N]} ;\|\cdot\|_{2, \infty}\right) 
    \\ 
    \leq &~
    \log \mathcal{N}_{\infty}\left(\mathcal{F}_{\text {hop }} ; \varepsilon ;\left\{\left( \bY^{(i)}, \bq_\tau^{(i)} \right)\right\}^{i \in [N]}_{\tau \in [T] } ;\|\cdot\|_{2}\right) 
    \\  
    \leq &~
    \varepsilon^{-2} \left( \left(  4 B_V^2 B_Y^2 \left( B_{\beta K}^{2,1} \right)^{2} \log \left(d N M\right)
    \right)^\frac{1}{3} 
    +
    \left(  \left( B_V^{2,1} \right)^2 \log(d N M) \right)^\frac{1}{3}
    \right)^3.
\end{align*}
\end{proof}
With the covering number bound, we arrive at the norm-based generalization bound \cref{lemma:gen_OutEffHop} through \cref{lemma:gen_thru_cn}.

\clearpage
\section{Additional Numerical Experiments}
\label{sec:add_exps}

\subsection{Supplemental Experimental Results (\cref{fig:differnet_layer} and \cref{fig:same_layer})}
\label{sec:sup_exp_details}
We conducted in-depth case studies on the BERT model. 
In Figure \ref{fig:differnet_layer}, we focus on the outlier performance in selected layers.
The figure shows that outliers become stronger in deeper layers of the vanilla model, corroborating insights from \citet{bondarenko2021understanding}. 
However, $\mathtt{OutEffHop}$ maintains a consistent maximum infinity norm $\norm{\mathbf{x}}_{\infty}$ across all layers, demonstrating its effectiveness in controlling outliers.
In Figure \ref{fig:same_layer}, we delve into the maximum infinity norm $\norm{\mathbf{x}}_{\infty}$ within the 10th layer's various tensor components. 
These tensors are after attention layer, the first residual layernorm after attention, and the first, second FFN layers. 
As mentioned in \cite{bondarenko2023quantizable}, FFN layers indeed increase the outliers heavily along the training process in vanilla attention. 
In contrast, $\mathtt{OutEffHop}$ suppresses the outliers growing in both FFN layers. 
The effectiveness of $\mathtt{OutEffHop}$ is due to its built-in no-operation (no-op) pattern which defaults queries to this pattern when updates are unnecessary.
This eliminates the need to learn outlier values in FFN layers to direct attention weights toward specific tokens for a no-op.
Additionally, we observe that 
the first residual LayerNorm after the attention mechanism tends to amplify outliers. 
This observation is also mentioned in \citet{wei2022outlier}'s finding. 
Furthermore, we note that the outliers of $\mathtt{OutEffHop}$ are larger than those of the vanilla model.
Our method, $\mathtt{OutEffHop}$, focusing solely on the attention mechanism, offers evidence of its effectiveness in mitigating outliers within our approach.

\subsection{Verifying Theoretical Results}

We also verify our theoretical findings following the settings in \cite{hu2023SparseHopfield}.

\label{sec:thm_exps}
\begin{figure*}[htb]
\begin{minipage}[t]{0.5\textwidth}
  \includegraphics[width=\linewidth]{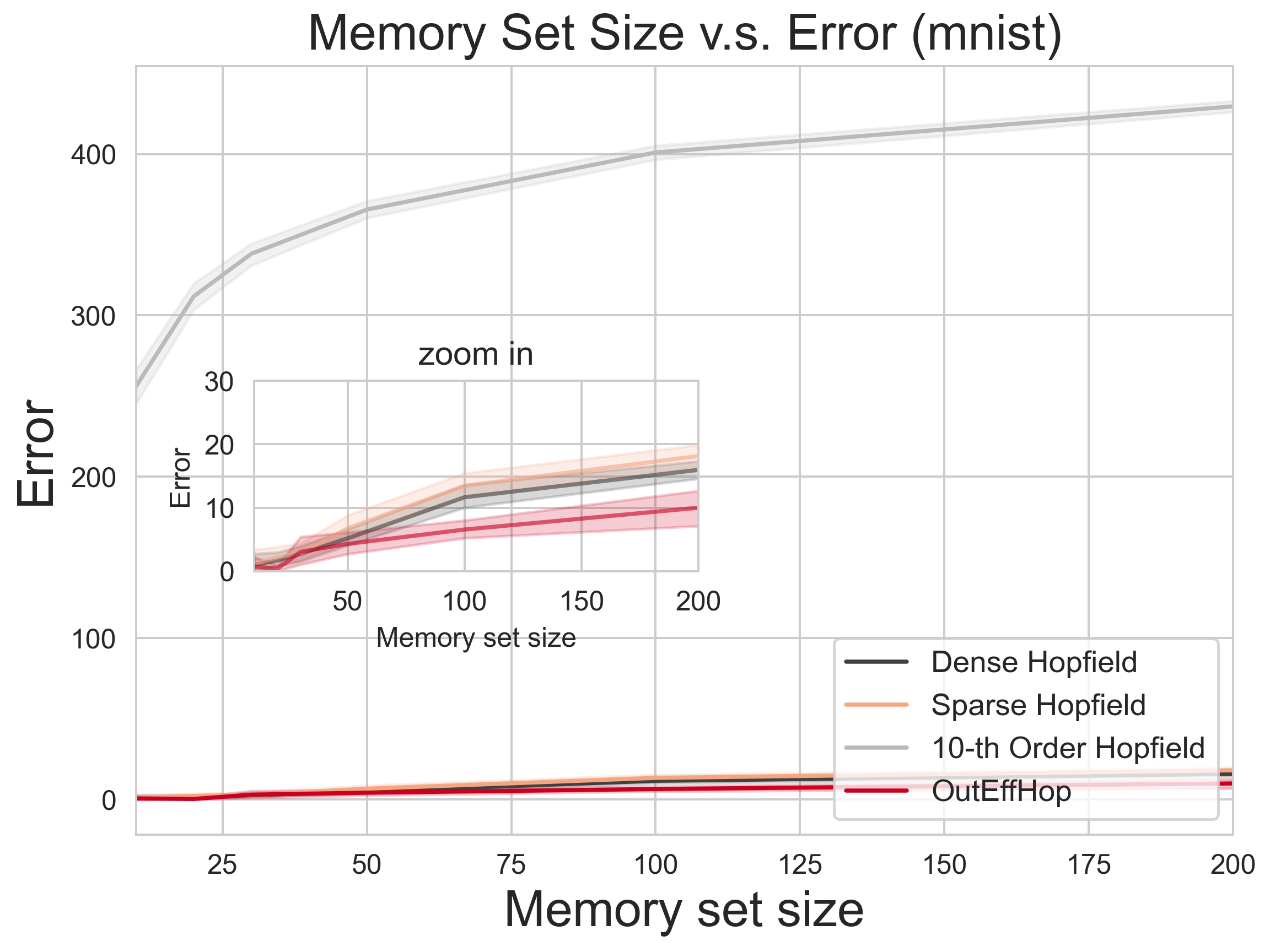}
\end{minipage}%
\hfill
\begin{minipage}[t]{0.5\textwidth}
  \includegraphics[width=\linewidth]{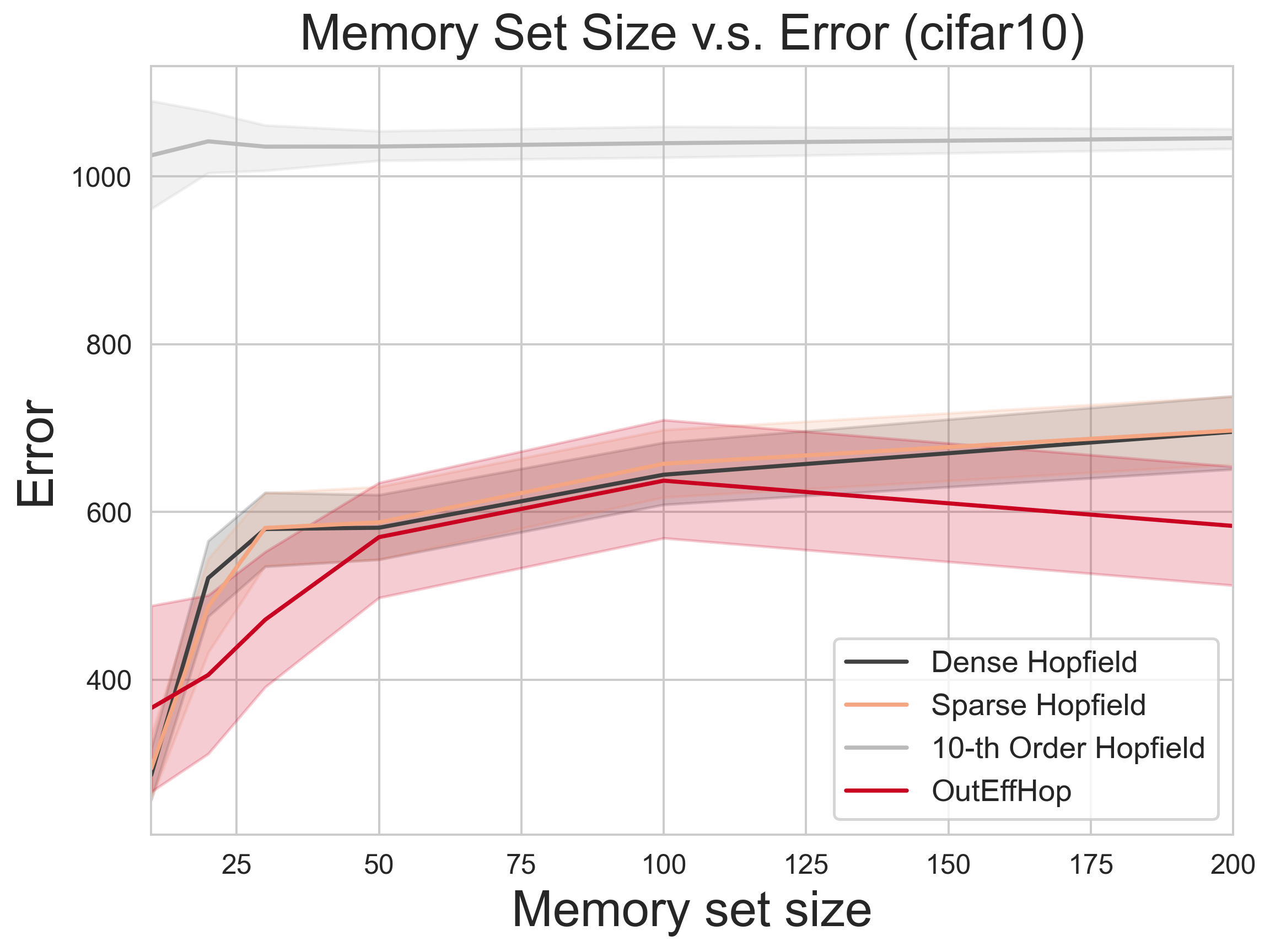}
\end{minipage}%
\caption{
\textbf{Memory Capacity.}
Our extensive evaluation of memory capacity across various Hopfield Networks, including Vanilla Modern Hopfield, Sparse Hopfield, 10th Order Hopfield, and our $\mathtt{OutEffHop}$, is conducted on two image datasets: MNIST and CIFAR10. We observe that $\mathtt{OutEffHop}$ outperforms its baselines, especially when the memory set size is large.
}
\label{fig:memory}
\end{figure*}

\begin{figure*}[htb]
\begin{minipage}[t]{0.5\textwidth}
  \includegraphics[width=\linewidth]{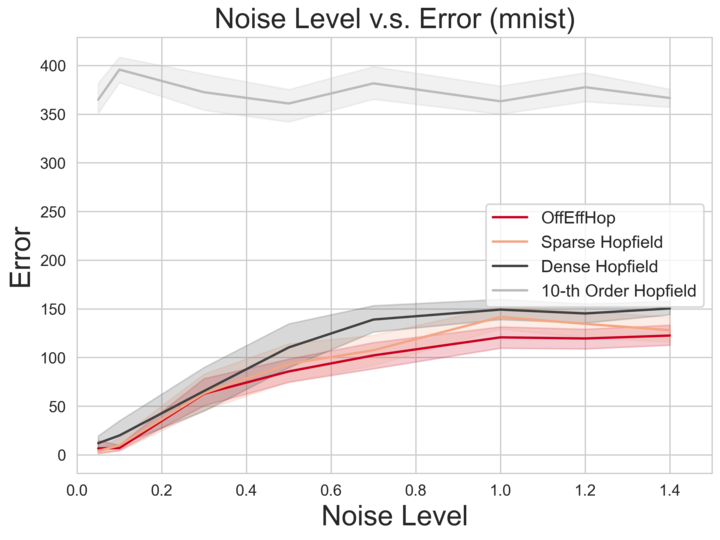}
\end{minipage}%
\hfill
\begin{minipage}[t]{0.5\textwidth}
  \includegraphics[width=\linewidth]{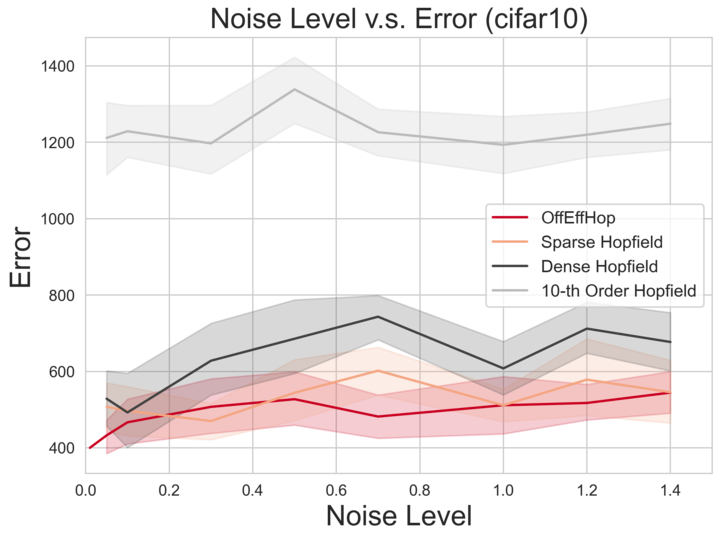}
\end{minipage}%
\caption{
\textbf{Noise-Robustness.}
Our extensive evaluation of noise robustness across various Hopfield Networks, including Vanilla Modern Hopfield, Sparse Hopfield, 10th Order Hopfield, and our $\mathtt{OutEffHop}$, is conducted on two image datasets: MNIST and CIFAR10. The results show that as the noise level rises, the impact of $\mathtt{OutEffHop}$ on the error rate is minimal.
}
\label{fig:noise}
\end{figure*}

\paragraph{Memory Capacity.}
For the memory capacity, we compare our Outlier-Efficient Modern Hopfield Model ($\mathtt{OutEffHop}$) with Dense (Softmax) \cite{,ramsauer2020hopfield}, Sparse \cite{hu2023SparseHopfield} and 10th order polynomial Hopfield model \cite{krotov2016dense} on MNIST \cite{lecun1998gradient} (high sparsity) and CIFAR10 \cite{krizhevsky2009learning} (low sparsity) datasets.
For all Hopfield models, we set $\beta = 1$. As shown in Figure \ref{fig:memory}, $\mathtt{OutEffHop}$ outperforms its baselines, especially when the memory set size is large.

\paragraph{Noise-Robustness.}
For the robustness against noise queries, we inject Gaussian noises varying variances ($\sigma$) into the images.
The results, as shown in Figure~\ref{fig:noise}, show that $\mathtt{OutEffHop}$ excels when the signal-to-noise ratio in patterns is low.

\paragraph{Faster Convergence.}
We numerically analyse the convergence of $\mathtt{OutEffHop}$, Dense and Sparse Hopfield model by evaluating their loss and accuracy in two different datasets.
We use the Vision Transfromer \cite{dosovitskiy2020image} (ViT) as the backbone and then replace the attention layer with different Hopfield layers. 
The hyperparameters used in our experiment are listed in Table \ref{table:hyper-tiny-cls}.
As shown in Figure \ref{fig:covergence}, our model surpasses its original counterpart across all datasets.

\begin{figure*}[!h]
\begin{tabular}{cccc}
\raisebox{+\height}{\rot{\tiny (a) CIFAR10}} & \includegraphics[width=0.95\linewidth]{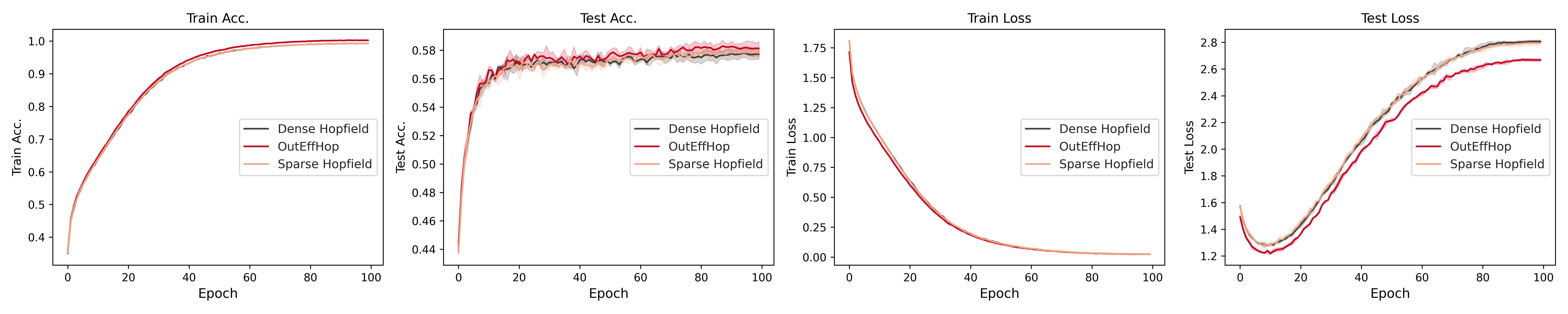} \\
\vspace{-1.4em}\\
\raisebox{+\height}{\rot{\tiny (b) CIFAR100}} & \includegraphics[width=0.95\linewidth]{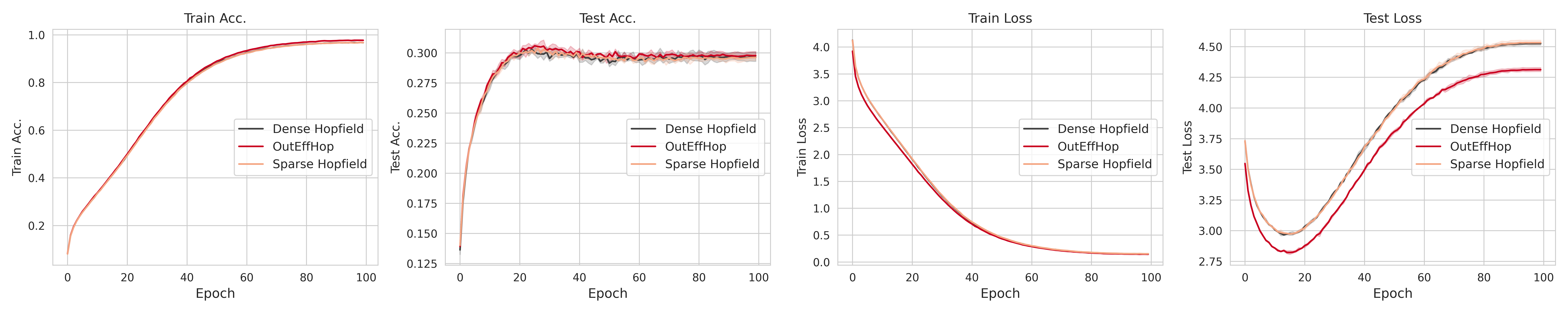} \\
\vspace{-1.5em}\\
\end{tabular}
\caption{
\textbf{Faster Convergence.} 
Our extensive evaluation of faster covergence across various Hopfield Networks, including Vanilla Modern Hopfield, Sparse Hopfield,
and our $\mathtt{OutEffHop}$, is conducted on two image datasets: CIFAR10 and CIFAR100.
The results show that  $\mathtt{OutEffHop}$ has faster convergence than baselines.
}
\label{fig:covergence}
\end{figure*}

\begin{table}[!ht]
        \centering
        \caption{Hyperparameter used in the fast convergence task.
        }
        \vspace*{0.05truein} 
        \begin{tabular}{l*{1}{c}}
        \toprule
            \bf{parameter} & \multicolumn{1}{c}{\bf{values}}\\ 
            \midrule
            learning rate & $1e-4$\\
            embedding dimension & 512 \\
            Feed forward dimension & 1024 \\
            Dropout & 0.3 \\
            activation function & GELU \\
            Epoch & 100 \\
            Batch size & 512 \\
            Model optimizer & Adam \\
            Patch size & 32 \\
            \bottomrule
        \end{tabular}
        \label{table:hyper-tiny-cls}
\end{table}

\clearpage
\subsection{Computational Cost Comparison}
{

We compare the computational resources of four different models against the vanilla $\Softmax$ and $\mathtt{OutEffHop}$, as detailed in Table~\ref{tab:resource}. 
We measure the pre-training records of all four models. 
Memory usage for OPT, BERT, and ViT is monitored using Wandb \footnote{\url{https://wandb.ai/}}, while for STanHop, it is tracked via system logs\footnote{We thank the authors of \cite{reneau2023feature} for their helpful comments on this part.}. 
The model sizes for this experiment match those described in section 4.1. 
Our experimental setup used a Slurm system with two 80G A100 GPUs and a 24-core Intel(R) Xeon(R) Gold 6338 CPU at 2.00GHz. 
We also provide the wandb diagram of the system memory usage in Figure~\ref{fig:resource}.
}

\begin{table}[htp]
    \centering
    \caption{The computational resource comparison of vanilla $\Softmax$ and $\mathtt{OutEffHop}$ in 4 models. We compare the Time and average of the Memory RAM usage in the model pre-training periods. }
    \vspace{1em}
    \begin{tabular}{ccc}
    \toprule
    Model & Method & Memory Usage (Gb) \\
    \midrule
    \multirow{2}{1em}{ViT} & Vanilla & 47.47 \\
    & OutEffHop & 49.69\\
    \midrule
    \multirow{2}{1em}{ERT} & Vanilla  & 7.56 \\
    & OutEffHop &   7.20\\
    \midrule
    \multirow{2}{1em}{OPT} & Vanilla  & 3.75 \\
    & OutEffHop & 3.75\\
     \midrule
    \multirow{2}{1em}{STN} & Vanilla &  5.30 \\
    & OutEffHop & 5.28\\
    \bottomrule
    \end{tabular}
    \label{tab:resource}
\end{table}

\begin{figure}
    \centering
    \includegraphics[width=.75\linewidth]{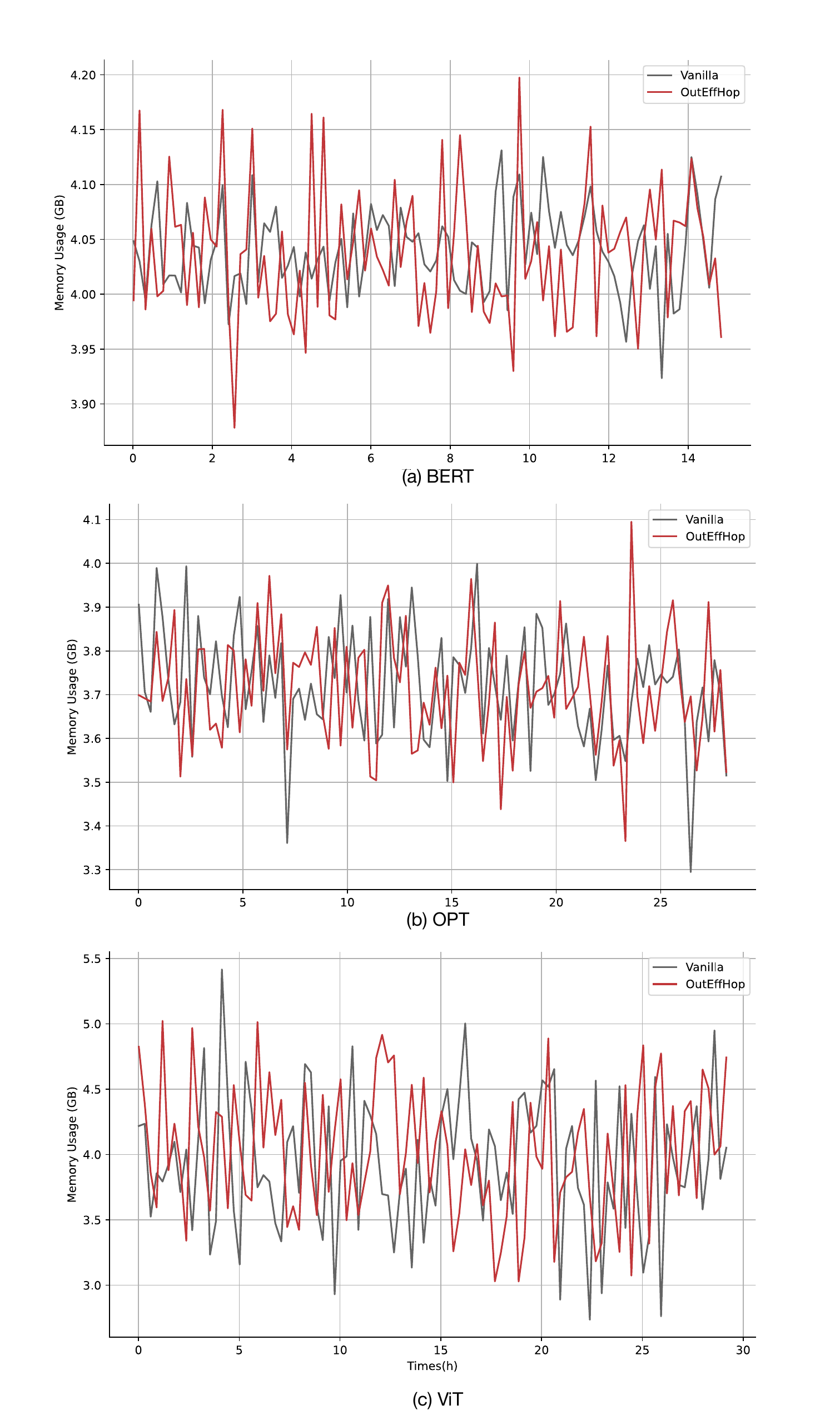}\\
     
    \caption{
The computational resource comparison between Vanilla $\Softmax$ and $\mathtt{OuTEffHop}$ involves measuring RAM usage via Wandb in a system equipped with 180G RAM under the Slurm system.}
    \label{fig:resource}
\end{figure}

\newpage
\twocolumn
\section*{Acknowledgments}
JH would like to thank Shang Wu, Yen-Ju Lu, Jing Liu, Jesus Villalba, Dino Feng and Andrew Chen for enlightening discussions, the Red Maple Family for support, and Jiayi Wang for facilitating experimental deployments.
The authors would also like to thank the anonymous reviewers and program chairs for their constructive comments.

JH is partially supported by the Walter P. Murphy Fellowship.
HL is partially supported by NIH R01LM1372201, NSF CAREER1841569, DOE DE-AC02-07CH11359, DOE LAB 20-2261 and a NSF TRIPODS1740735.
This research was supported in part through the computational resources and staff contributions provided for the Quest high performance computing facility at Northwestern University which is jointly supported by the Office of the Provost, the Office for Research, and Northwestern University Information Technology.
The content is solely the responsibility of the authors and does not necessarily represent the official
views of the funding agencies.

\def\arxivfont{\rm}
\bibliographystyle{plainnat}

\bibliography{refs}

\begin{thebibliography}{74}
\providecommand{\natexlab}[1]{#1}
\providecommand{\url}[1]{\texttt{#1}}
\expandafter\ifx\csname urlstyle\endcsname\relax
  \providecommand{\doi}[1]{doi: #1}\else
  \providecommand{\doi}{doi: \begingroup \urlstyle{rm}\Url}\fi

\bibitem[Alman and Song(2023)]{alman2023fast}
Josh Alman and Zhao Song.
\newblock Fast attention requires bounded entries.
\newblock In \emph{Thirty-seventh Conference on Neural Information Processing Systems (NeurIPS)}, 2023.
\newblock URL \url{https://openreview.net/forum?id=KOVWXcrFIK}.

\bibitem[Alman and Song(2024{\natexlab{a}})]{alman2024fine}
Josh Alman and Zhao Song.
\newblock The fine-grained complexity of gradient computation for training large language models.
\newblock \emph{arXiv preprint arXiv:2402.04497}, 2024{\natexlab{a}}.

\bibitem[Alman and Song(2024{\natexlab{b}})]{as23_tensor}
Josh Alman and Zhao Song.
\newblock How to capture higher-order correlations? generalizing matrix softmax attention to kronecker computation.
\newblock In \emph{The Twelfth International Conference on Learning Representations (ICLR)}, 2024{\natexlab{b}}.
\newblock URL \url{https://openreview.net/forum?id=v0zNCwwkaV}.

\bibitem[Auer et~al.(2023)Auer, Gauch, Klotz, and Hochreiter]{auer2023conformal}
Andreas Auer, Martin Gauch, Daniel Klotz, and Sepp Hochreiter.
\newblock Conformal prediction for time series with modern hopfield networks.
\newblock \emph{Advances in Neural Information Processing Systems}, 36, 2023.
\newblock URL \url{https://arxiv.org/abs/2303.12783}.

\bibitem[Bhandare et~al.(2019)Bhandare, Sripathi, Karkada, Menon, Choi, Datta, and Saletore]{bhandare2019efficient}
Aishwarya Bhandare, Vamsi Sripathi, Deepthi Karkada, Vivek Menon, Sun Choi, Kushal Datta, and Vikram Saletore.
\newblock Efficient 8-bit quantization of transformer neural machine language translation model.
\newblock \emph{arXiv preprint arXiv:1906.00532}, 2019.
\newblock URL \url{https://arxiv.org/abs/1906.00532}.

\bibitem[Bietti et~al.(2023)Bietti, Cabannes, Bouchacourt, Jegou, and Bottou]{bietti2024birth}
Alberto Bietti, Vivien Cabannes, Diane Bouchacourt, Herve Jegou, and Leon Bottou.
\newblock Birth of a transformer: A memory viewpoint.
\newblock \emph{Advances in Neural Information Processing Systems (NeurIPS)}, 36, 2023.
\newblock URL \url{https://arxiv.org/abs/2306.00802}.

\bibitem[Bommasani et~al.(2021)Bommasani, Hudson, Adeli, Altman, Arora, von Arx, Bernstein, Bohg, Bosselut, Brunskill, et~al.]{bommasani2021opportunities}
Rishi Bommasani, Drew~A Hudson, Ehsan Adeli, Russ Altman, Simran Arora, Sydney von Arx, Michael~S Bernstein, Jeannette Bohg, Antoine Bosselut, Emma Brunskill, et~al.
\newblock On the opportunities and risks of foundation models.
\newblock \emph{arXiv preprint arXiv:2108.07258}, 2021.
\newblock URL \url{https://arxiv.org/abs/2108.07258}.

\bibitem[Bondarenko et~al.(2021)Bondarenko, Nagel, and Blankevoort]{bondarenko2021understanding}
Yelysei Bondarenko, Markus Nagel, and Tijmen Blankevoort.
\newblock Understanding and overcoming the challenges of efficient transformer quantization, 2021.
\newblock URL \url{https://arxiv.org/abs/2109.12948}.

\bibitem[Bondarenko et~al.(2023)Bondarenko, Nagel, and Blankevoort]{bondarenko2023quantizable}
Yelysei Bondarenko, Markus Nagel, and Tijmen Blankevoort.
\newblock Quantizable transformers: Removing outliers by helping attention heads do nothing.
\newblock \emph{Advances in Neural Information Processing Systems (NeurIPS)}, 36, 2023.
\newblock URL \url{https://arxiv.org/abs/2306.12929}.

\bibitem[Brandstetter(2021)]{hopfeildblog2021}
Johannes Brandstetter.
\newblock Blog post: Hopfield networks is all you need, 2021.
\newblock URL \url{https://ml-jku.github.io/hopfield-layers/}.
\newblock Accessed: April 4, 2023.

\bibitem[Brown et~al.(2020)Brown, Mann, Ryder, Subbiah, Kaplan, Dhariwal, Neelakantan, Shyam, Sastry, Askell, et~al.]{brown2020language}
Tom Brown, Benjamin Mann, Nick Ryder, Melanie Subbiah, Jared~D Kaplan, Prafulla Dhariwal, Arvind Neelakantan, Pranav Shyam, Girish Sastry, Amanda Askell, et~al.
\newblock Language models are few-shot learners.
\newblock \emph{Advances in neural information processing systems}, 33:\penalty0 1877--1901, 2020.
\newblock URL \url{https://arxiv.org/abs/2005.14165}.

\bibitem[Burns(2024)]{burns2024semantically}
Thomas~F Burns.
\newblock Semantically-correlated memories in a dense associative model.
\newblock In \emph{Forty-first International Conference on Machine Learning (ICML)}, 2024.
\newblock URL \url{https://arxiv.org/abs/2404.07123}.

\bibitem[Burns and Fukai(2023)]{burns2023simplicial}
Thomas~F Burns and Tomoki Fukai.
\newblock Simplicial hopfield networks.
\newblock In \emph{The Eleventh International Conference on Learning Representations (ICLR)}, 2023.
\newblock URL \url{https://openreview.net/forum?id=_QLsH8gatwx}.

\bibitem[Cabannes et~al.(2024)Cabannes, Dohmatob, and Bietti]{cabannes2023scaling}
Vivien Cabannes, Elvis Dohmatob, and Alberto Bietti.
\newblock Scaling laws for associative memories.
\newblock In \emph{The Twelfth International Conference on Learning Representations (ICLR)}, 2024.
\newblock URL \url{https://openreview.net/forum?id=Tzh6xAJSll}.

\bibitem[Chaudhry et~al.(2023)Chaudhry, Zavatone-Veth, Krotov, and Pehlevan]{chaudhry2024long}
Hamza Chaudhry, Jacob Zavatone-Veth, Dmitry Krotov, and Cengiz Pehlevan.
\newblock Long sequence hopfield memory.
\newblock \emph{Advances in Neural Information Processing Systems (NeurIPS)}, 36, 2023.
\newblock URL \url{https://arxiv.org/abs/2306.04532}.

\bibitem[Clark et~al.(2019)Clark, Khandelwal, Levy, and Manning]{clark-etal-2019-bert}
Kevin Clark, Urvashi Khandelwal, Omer Levy, and Christopher~D. Manning.
\newblock revealt does {BERT} look at? an analysis of {BERT}{'}s attention.
\newblock In Tal Linzen, Grzegorz Chrupa{\l}a, Yonatan Belinkov, and Dieuwke Hupkes, editors, \emph{Proceedings of the 2019 ACL Workshop BlackboxNLP: Analyzing and Interpreting Neural Networks for NLP}, pages 276--286, Florence, Italy, August 2019. Association for Computational Linguistics.
\newblock \doi{10.18653/v1/W19-4828}.
\newblock URL \url{https://aclanthology.org/W19-4828}.

\bibitem[Demircigil et~al.(2017)Demircigil, Heusel, L{\"o}we, Upgang, and Vermet]{demircigil2017model}
Mete Demircigil, Judith Heusel, Matthias L{\"o}we, Sven Upgang, and Franck Vermet.
\newblock On a model of associative memory with huge storage capacity.
\newblock \emph{Journal of Statistical Physics}, 168:\penalty0 288--299, 2017.
\newblock URL \url{https://arxiv.org/abs/1702.01929}.

\bibitem[Dettmers et~al.(2022)Dettmers, Lewis, Belkada, and Zettlemoyer]{llm.int.8}
Tim Dettmers, Mike Lewis, Younes Belkada, and Luke Zettlemoyer.
\newblock Gpt3.int8(): 8-bit matrix multiplication for transformers at scale.
\newblock In S.~Koyejo, S.~Mohamed, A.~Agarwal, D.~Belgrave, K.~Cho, and A.~Oh, editors, \emph{Advances in Neural Information Processing Systems}, volume~35, pages 30318--30332. Curran Associates, Inc., 2022.
\newblock URL \url{https://proceedings.neurips.cc/paper_files/paper/2022/file/c3ba4962c05c49636d4c6206a97e9c8a-Paper-Conference.pdf}.

\bibitem[Devlin et~al.(2019)Devlin, Chang, Lee, and Toutanova]{devlin-etal-2019-bert}
Jacob Devlin, Ming-Wei Chang, Kenton Lee, and Kristina Toutanova.
\newblock {BERT}: Pre-training of deep bidirectional transformers for language understanding.
\newblock In Jill Burstein, Christy Doran, and Thamar Solorio, editors, \emph{Proceedings of the 2019 Conference of the North {A}merican Chapter of the Association for Computational Linguistics: Human Language Technologies, Volume 1 (Long and Short Papers)}, Minneapolis, Minnesota, June 2019. Association for Computational Linguistics.
\newblock URL \url{https://aclanthology.org/N19-1423}.

\bibitem[Dosovitskiy et~al.(2020)Dosovitskiy, Beyer, Kolesnikov, Weissenborn, Zhai, Unterthiner, Dehghani, Minderer, Heigold, Gelly, et~al.]{dosovitskiy2020image}
Alexey Dosovitskiy, Lucas Beyer, Alexander Kolesnikov, Dirk Weissenborn, Xiaohua Zhai, Thomas Unterthiner, Mostafa Dehghani, Matthias Minderer, Georg Heigold, Sylvain Gelly, et~al.
\newblock An image is worth 16x16 words: Transformers for image recognition at scale.
\newblock \emph{arXiv preprint arXiv:2010.11929}, 2020.
\newblock URL \url{https://arxiv.org/abs/2010.11929}.

\bibitem[Dudley(1978)]{dudley1978central}
Richard~M Dudley.
\newblock Central limit theorems for empirical measures.
\newblock \emph{The Annals of Probability}, pages 899--929, 1978.
\newblock URL \url{https://projecteuclid.org/journals/annals-of-probability/volume-6/issue-6/Central-Limit-Theorems-for-Empirical-Measures/10.1214/aop/1176995384.full}.

\bibitem[Edelman et~al.(2022)Edelman, Goel, Kakade, and Zhang]{edelman2022inductive}
Benjamin~L Edelman, Surbhi Goel, Sham Kakade, and Cyril Zhang.
\newblock Inductive biases and variable creation in self-attention mechanisms.
\newblock In \emph{International Conference on Machine Learning}, pages 5793--5831. PMLR, 2022.
\newblock URL \url{https://arxiv.org/abs/2110.10090}.

\bibitem[Floridi and Chiriatti(2020)]{floridi2020gpt}
Luciano Floridi and Massimo Chiriatti.
\newblock Gpt-3: Its nature, scope, limits, and consequences.
\newblock \emph{Minds and Machines}, 30:\penalty0 681--694, 2020.
\newblock URL \url{https://link.springer.com/article/10.1007/s11023-020-09548-1}.

\bibitem[F{\"u}rst et~al.(2022)F{\"u}rst, Rumetshofer, Lehner, Tran, Tang, Ramsauer, Kreil, Kopp, Klambauer, Bitto, et~al.]{furst2022cloob}
Andreas F{\"u}rst, Elisabeth Rumetshofer, Johannes Lehner, Viet~T Tran, Fei Tang, Hubert Ramsauer, David Kreil, Michael Kopp, G{\"u}nter Klambauer, Angela Bitto, et~al.
\newblock Cloob: Modern hopfield networks with infoloob outperform clip.
\newblock \emph{Advances in neural information processing systems}, 35:\penalty0 20450--20468, 2022.
\newblock URL \url{https://arxiv.org/abs/2110.11316}.

\bibitem[Gao et~al.(2023)Gao, Song, Wang, and Yin]{gao2023fast}
Yeqi Gao, Zhao Song, Weixin Wang, and Junze Yin.
\newblock A fast optimization view: Reformulating single layer attention in llm based on tensor and svm trick, and solving it in matrix multiplication time.
\newblock \emph{arXiv preprint arXiv:2309.07418}, 2023.

\bibitem[Gu et~al.(2024{\natexlab{a}})Gu, Liang, Liu, Shi, Song, and Yin]{gu2024conv}
Jiuxiang Gu, Yingyu Liang, Heshan Liu, Zhenmei Shi, Zhao Song, and Junze Yin.
\newblock Conv-basis: A new paradigm for efficient attention inference and gradient computation in transformers.
\newblock \emph{arXiv preprint arXiv:2405.05219}, 2024{\natexlab{a}}.
\newblock URL \url{https://arxiv.org/abs/2405.05219}.

\bibitem[Gu et~al.(2024{\natexlab{b}})Gu, Liang, Shi, Song, and Zhou]{gu2024tensor}
Jiuxiang Gu, Yingyu Liang, Zhenmei Shi, Zhao Song, and Yufa Zhou.
\newblock Tensor attention training: Provably efficient learning of higher-order transformers.
\newblock \emph{arXiv preprint arXiv:2405.16411}, 2024{\natexlab{b}}.
\newblock URL \url{https://arxiv.org/abs/2405.16411}.

\bibitem[Guo et~al.(2020)Guo, Dai, Vrande{\v{c}}i{\'c}, and Al-Rfou]{guo2020wiki}
Mandy Guo, Zihang Dai, Denny Vrande{\v{c}}i{\'c}, and Rami Al-Rfou.
\newblock Wiki-40b: Multilingual language model dataset.
\newblock In \emph{Proceedings of the Twelfth Language Resources and Evaluation Conference}, pages 2440--2452, 2020.
\newblock URL \url{https://aclanthology.org/2020.lrec-1.297/}.

\bibitem[Hofmann et~al.(2024)Hofmann, Schmid, Lehner, Klotz, and Hochreiter]{hofmann2024energy}
Claus Hofmann, Simon Schmid, Bernhard Lehner, Daniel Klotz, and Sepp Hochreiter.
\newblock Energy-based hopfield boosting for out-of-distribution detection.
\newblock \emph{arXiv preprint arXiv:2405.08766}, 2024.

\bibitem[Hoover et~al.(2023)Hoover, Liang, Pham, Panda, Strobelt, Chau, Zaki, and Krotov]{hoover2023energy}
Benjamin Hoover, Yuchen Liang, Bao Pham, Rameswar Panda, Hendrik Strobelt, Duen~Horng Chau, Mohammed~J Zaki, and Dmitry Krotov.
\newblock Energy transformer.
\newblock \emph{arXiv preprint arXiv:2302.07253}, 2023.
\newblock URL \url{https://arxiv.org/abs/2302.07253}.

\bibitem[Hopfield(1982)]{hopfield1982neural}
John~J Hopfield.
\newblock Neural networks and physical systems with emergent collective computational abilities.
\newblock \emph{Proceedings of the national academy of sciences}, 79\penalty0 (8):\penalty0 2554--2558, 1982.
\newblock URL \url{https://www.pnas.org/doi/10.1073/pnas.79.8.2554?trk=public_post_comment-text}.

\bibitem[Hopfield(1984)]{hopfield1984neurons}
John~J Hopfield.
\newblock Neurons with graded response have collective computational properties like those of two-state neurons.
\newblock \emph{Proceedings of the national academy of sciences}, 81\penalty0 (10):\penalty0 3088--3092, 1984.
\newblock URL \url{https://www.pnas.org/doi/10.1073/pnas.81.10.3088}.

\bibitem[Horowitz(2014)]{horowitz20141}
Mark Horowitz.
\newblock 1.1 computing's energy problem (and what we can do about it).
\newblock In \emph{2014 IEEE international solid-state circuits conference digest of technical papers (ISSCC)}, pages 10--14. IEEE, 2014.
\newblock URL \url{https://ieeexplore.ieee.org/document/6757323}.

\bibitem[Hu et~al.(2023)Hu, Yang, Wu, Xu, Chen, and Liu]{hu2023SparseHopfield}
Jerry Yao-Chieh Hu, Donglin Yang, Dennis Wu, Chenwei Xu, Bo-Yu Chen, and Han Liu.
\newblock On sparse modern hopfield model.
\newblock In \emph{Thirty-seventh Conference on Neural Information Processing Systems (NeurIPS)}, 2023.
\newblock URL \url{https://arxiv.org/abs/2309.12673}.

\bibitem[Hu et~al.(2024{\natexlab{a}})Hu, Chen, Wu, Ruan, and Liu]{hu2024nonparametric}
Jerry Yao-Chieh Hu, Bo-Yu Chen, Dennis Wu, Feng Ruan, and Han Liu.
\newblock Nonparametric modern hopfield models.
\newblock \emph{arXiv preprint arXiv:2404.03900}, 2024{\natexlab{a}}.
\newblock URL \url{https://arxiv.org/abs/2404.03900}.

\bibitem[Hu et~al.(2024{\natexlab{b}})Hu, Lin, Song, and Liu]{hu2024computational}
Jerry Yao-Chieh Hu, Thomas Lin, Zhao Song, and Han Liu.
\newblock On computational limits of modern hopfield models: A fine-grained complexity analysis.
\newblock In \emph{Forty-first International Conference on Machine Learning (ICML)}, 2024{\natexlab{b}}.
\newblock URL \url{https://arxiv.org/abs/2402.04520}.

\bibitem[Ji et~al.(2021)Ji, Zhou, Liu, and Davuluri]{ji2021dnabert}
Yanrong Ji, Zhihan Zhou, Han Liu, and Ramana~V Davuluri.
\newblock Dnabert: pre-trained bidirectional encoder representations from transformers model for dna-language in genome.
\newblock \emph{Bioinformatics}, 37\penalty0 (15):\penalty0 2112--2120, 2021.
\newblock URL \url{https://academic.oup.com/bioinformatics/article/37/15/2112/6128680}.

\bibitem[johnowhitaker(2023)]{john2023}
johnowhitaker.
\newblock Blog post: Exploring softmax1, or “community research for the win!”, 2023.
\newblock URL \url{https://datasciencecastnet.home.blog/2023/08/04/exploring-softmax1-or-community-research-for-the-win/}.
\newblock Accessed: August 4, 2023.

\bibitem[Junczys-Dowmunt et~al.(2018)Junczys-Dowmunt, Heafield, Hoang, Grundkiewicz, and Aue]{junczys2018marian}
Marcin Junczys-Dowmunt, Kenneth Heafield, Hieu Hoang, Roman Grundkiewicz, and Anthony Aue.
\newblock Marian: Cost-effective high-quality neural machine translation in c++.
\newblock \emph{arXiv preprint arXiv:1805.12096}, 2018.
\newblock URL \url{https://arxiv.org/abs/1805.12096}.

\bibitem[Kobayashi et~al.(2020)Kobayashi, Kuribayashi, Yokoi, and Inui]{kobayashi2020attention}
Goro Kobayashi, Tatsuki Kuribayashi, Sho Yokoi, and Kentaro Inui.
\newblock Attention is not only a weight: Analyzing transformers with vector norms, 2020.
\newblock URL \url{https://aclanthology.org/2020.emnlp-main.574/}.

\bibitem[Kovaleva et~al.(2019)Kovaleva, Romanov, Rogers, and Rumshisky]{kovaleva2019revealing}
Olga Kovaleva, Alexey Romanov, Anna Rogers, and Anna Rumshisky.
\newblock Revealing the dark secrets of bert, 2019.
\newblock URL \url{https://arxiv.org/abs/1908.08593}.

\bibitem[Kozachkov et~al.(2022)Kozachkov, Kastanenka, and Krotov]{kozachkov2022building}
Leo Kozachkov, Ksenia~V Kastanenka, and Dmitry Krotov.
\newblock Building transformers from neurons and astrocytes.
\newblock \emph{bioRxiv}, pages 2022--10, 2022.
\newblock URL \url{https://www.pnas.org/doi/10.1073/pnas.2219150120}.

\bibitem[Krizhevsky et~al.(2009)Krizhevsky, Hinton, et~al.]{krizhevsky2009learning}
Alex Krizhevsky, Geoffrey Hinton, et~al.
\newblock Learning multiple layers of features from tiny images.
\newblock 2009.
\newblock URL \url{https://www.cs.utoronto.ca/~kriz/learning-features-2009-TR.pdf}.

\bibitem[Krotov and Hopfield(2016)]{krotov2016dense}
Dmitry Krotov and John~J. Hopfield.
\newblock Dense associative memory for pattern recognition.
\newblock \emph{CoRR}, 2016.
\newblock URL \url{https://arxiv.org/abs/1606.01164}.

\bibitem[Krotov and Hopfield(2021)]{krotov2020large}
Dmitry Krotov and John~J. Hopfield.
\newblock Large associative memory problem in neurobiology and machine learning.
\newblock In \emph{International Conference on Learning Representations}, 2021.
\newblock URL \url{https://arxiv.org/abs/2008.06996}.

\bibitem[LeCun et~al.(1998)LeCun, Bottou, Bengio, and Haffner]{lecun1998gradient}
Yann LeCun, L{\'e}on Bottou, Yoshua Bengio, and Patrick Haffner.
\newblock Gradient-based learning applied to document recognition.
\newblock \emph{Proceedings of the IEEE}, 86\penalty0 (11):\penalty0 2278--2324, 1998.
\newblock URL \url{https://ieeexplore.ieee.org/document/726791}.

\bibitem[Liang(2016)]{liang2016cs229t}
Percy Liang.
\newblock Cs229t/stat231: Statistical learning theory (winter 2016), 2016.
\newblock URL \url{https://web.stanford.edu/class/cs229t/notes.pdf}.

\bibitem[Marchesi et~al.(1993)Marchesi, Orlandi, Piazza, and Uncini]{182695}
M.~Marchesi, G.~Orlandi, F.~Piazza, and A.~Uncini.
\newblock Fast neural networks without multipliers.
\newblock \emph{IEEE Transactions on Neural Networks}, 4\penalty0 (1):\penalty0 53--62, 1993.
\newblock \doi{10.1109/72.182695}.
\newblock URL \url{https://ieeexplore.ieee.org/document/182695}.

\bibitem[Miller(2023)]{miller2021}
Evan Miller.
\newblock Blog post: Attention is off by one, 2023.
\newblock URL \url{https://www.evanmiller.org/attention-is-off-by-one.html}.
\newblock Accessed: August 4, 2023.

\bibitem[Olver et~al.(2010)Olver, Lozier, Boisvert, and Clark]{olver2010nist}
Frank~WJ Olver, Daniel~W Lozier, Ronald~F Boisvert, and Charles~W Clark.
\newblock \emph{NIST handbook of mathematical functions hardback and CD-ROM}.
\newblock Cambridge university press, 2010.
\newblock URL \url{https://www.amazon.com/Handbook-Mathematical-Functions-Hardback-CD-ROM/dp/0521192250}.

\bibitem[Paischer et~al.(2022)Paischer, Adler, Patil, Bitto-Nemling, Holzleitner, Lehner, Eghbal-Zadeh, and Hochreiter]{paischer2022history}
Fabian Paischer, Thomas Adler, Vihang Patil, Angela Bitto-Nemling, Markus Holzleitner, Sebastian Lehner, Hamid Eghbal-Zadeh, and Sepp Hochreiter.
\newblock History compression via language models in reinforcement learning.
\newblock In \emph{International Conference on Machine Learning}, pages 17156--17185. PMLR, 2022.
\newblock URL \url{https://arxiv.org/abs/2205.12258}.

\bibitem[Ramsauer et~al.(2020)Ramsauer, Schafl, Lehner, Seidl, Widrich, Adler, Gruber, Holzleitner, Pavlovic, Sandve, et~al.]{ramsauer2020hopfield}
Hubert Ramsauer, Bernhard Schafl, Johannes Lehner, Philipp Seidl, Michael Widrich, Thomas Adler, Lukas Gruber, Markus Holzleitner, Milena Pavlovic, Geir~Kjetil Sandve, et~al.
\newblock Hopfield networks is all you need.
\newblock \emph{arXiv preprint arXiv:2008.02217}, 2020.
\newblock URL \url{https://arxiv.org/abs/2008.02217}.

\bibitem[Reneau et~al.(2023)Reneau, Hu, Xu, Li, Gilani, and Liu]{reneau2023feature}
Alex Reneau, Jerry Yao-Chieh Hu, Chenwei Xu, Weijian Li, Ammar Gilani, and Han Liu.
\newblock Feature programming for multivariate time series prediction.
\newblock In \emph{Proceedings of the 40th International Conference on Machine Learning (ICML)}, volume 202 of \emph{Proceedings of Machine Learning Research}, pages 29009--29029. PMLR, 23--29 Jul 2023.
\newblock URL \url{https://arxiv.org/abs/2306.06252}.

\bibitem[Russakovsky et~al.(2015)Russakovsky, Deng, Su, Krause, Satheesh, Ma, Huang, Karpathy, Khosla, Bernstein, Berg, and Fei-Fei]{imagenet15russakovsky}
Olga Russakovsky, Jia Deng, Hao Su, Jonathan Krause, Sanjeev Satheesh, Sean Ma, Zhiheng Huang, Andrej Karpathy, Aditya Khosla, Michael Bernstein, Alexander~C. Berg, and Li~Fei-Fei.
\newblock Imagenet large scale visual recognition challenge.
\newblock \emph{International Journal of Computer Vision (IJCV)}, 115\penalty0 (3):\penalty0 211--252, 2015.
\newblock \doi{10.1007/s11263-015-0816-y}.
\newblock URL \url{https://arxiv.org/abs/1409.0575}.

\bibitem[Schimunek et~al.(2023)Schimunek, Seidl, Friedrich, Kuhn, Rippmann, Hochreiter, and Klambauer]{schimunek2023contextenriched}
Johannes Schimunek, Philipp Seidl, Lukas Friedrich, Daniel Kuhn, Friedrich Rippmann, Sepp Hochreiter, and G{\"u}nter Klambauer.
\newblock Context-enriched molecule representations improve few-shot drug discovery.
\newblock In \emph{The Eleventh International Conference on Learning Representations}, 2023.
\newblock URL \url{https://openreview.net/forum?id=XrMWUuEevr}.

\bibitem[Seidl et~al.(2022)Seidl, Renz, Dyubankova, Neves, Verhoeven, Wegner, Segler, Hochreiter, and Klambauer]{seidl2022improving}
Philipp Seidl, Philipp Renz, Natalia Dyubankova, Paulo Neves, Jonas Verhoeven, Jorg~K Wegner, Marwin Segler, Sepp Hochreiter, and Gunter Klambauer.
\newblock Improving few-and zero-shot reaction template prediction using modern hopfield networks.
\newblock \emph{Journal of chemical information and modeling}, 62\penalty0 (9):\penalty0 2111--2120, 2022.
\newblock URL \url{https://pubs.acs.org/doi/10.1021/acs.jcim.1c01065}.

\bibitem[Shkolnik et~al.(2020)Shkolnik, Chmiel, Banner, Shomron, Nahshan, Bronstein, and Weiser]{shkolnik2020robust}
Moran Shkolnik, Brian Chmiel, Ron Banner, Gil Shomron, Yury Nahshan, Alex Bronstein, and Uri Weiser.
\newblock Robust quantization: One model to rule them all, 2020.
\newblock URL \url{https://arxiv.org/abs/2002.07686}.

\bibitem[Sriperumbudur and Lanckriet(2009)]{sriperumbudur2009convergence}
Bharath~K Sriperumbudur and Gert~RG Lanckriet.
\newblock On the convergence of the concave-convex procedure.
\newblock In \emph{Advances in neural information processing systems}, volume~9, pages 1759--1767, 2009.
\newblock URL \url{https://papers.nips.cc/paper_files/paper/2009/file/8b5040a8a5baf3e0e67386c2e3a9b903-Paper.pdf}.

\bibitem[Tang and Kwan(1993)]{229903}
C.Z. Tang and H.K. Kwan.
\newblock Multilayer feedforward neural networks with single powers-of-two weights.
\newblock \emph{IEEE Transactions on Signal Processing}, 41\penalty0 (8):\penalty0 2724--2727, 1993.
\newblock \doi{10.1109/78.229903}.
\newblock URL \url{https://ieeexplore.ieee.org/document/229903}.

\bibitem[Vaswani et~al.(2017)Vaswani, Shazeer, Parmar, Uszkoreit, Jones, Gomez, Kaiser, and Polosukhin]{vaswani2017attention}
Ashish Vaswani, Noam Shazeer, Niki Parmar, Jakob Uszkoreit, Llion Jones, Aidan~N Gomez, {\L}ukasz Kaiser, and Illia Polosukhin.
\newblock Attention is all you need.
\newblock \emph{Advances in neural information processing systems}, 30, 2017.
\newblock URL \url{https://arxiv.org/abs/1706.03762}.

\bibitem[Wei et~al.(2022)Wei, Zhang, Zhang, Gong, Zhang, Zhang, Yu, and Liu]{wei2022outlier}
Xiuying Wei, Yunchen Zhang, Xiangguo Zhang, Ruihao Gong, Shanghang Zhang, Qi~Zhang, Fengwei Yu, and Xianglong Liu.
\newblock Outlier suppression: Pushing the limit of low-bit transformer language models.
\newblock \emph{Advances in Neural Information Processing Systems}, 35:\penalty0 17402--17414, 2022.
\newblock URL \url{https://arxiv.org/abs/2209.13325}.

\bibitem[Widrich et~al.(2020)Widrich, Sch{\"a}fl, Pavlovi{\'c}, Ramsauer, Gruber, Holzleitner, Brandstetter, Sandve, Greiff, Hochreiter, et~al.]{widrich2020modern}
Michael Widrich, Bernhard Sch{\"a}fl, Milena Pavlovi{\'c}, Hubert Ramsauer, Lukas Gruber, Markus Holzleitner, Johannes Brandstetter, Geir~Kjetil Sandve, Victor Greiff, Sepp Hochreiter, et~al.
\newblock Modern hopfield networks and attention for immune repertoire classification.
\newblock \emph{Advances in Neural Information Processing Systems}, 33:\penalty0 18832--18845, 2020.
\newblock URL \url{https://arxiv.org/abs/2007.13505}.

\bibitem[Wu et~al.(2024{\natexlab{a}})Wu, Hu, Hsiao, and Liu]{wu2024uniform}
Dennis Wu, Jerry Yao-Chieh Hu, Teng-Yun Hsiao, and Han Liu.
\newblock Uniform memory retrieval with larger capacity for modern hopfield models.
\newblock In \emph{Forty-first International Conference on Machine Learning (ICML)}, 2024{\natexlab{a}}.
\newblock URL \url{https://arxiv.org/abs/2404.03827}.

\bibitem[Wu et~al.(2024{\natexlab{b}})Wu, Hu, Li, Chen, and Liu]{wu2023stanhop}
Dennis Wu, Jerry Yao-Chieh Hu, Weijian Li, Bo-Yu Chen, and Han Liu.
\newblock {ST}anhop: Sparse tandem hopfield model for memory-enhanced time series prediction.
\newblock In \emph{The Twelfth International Conference on Learning Representations (ICLR)}, 2024{\natexlab{b}}.
\newblock URL \url{https://arxiv.org/abs/2312.17346}.

\bibitem[Wu et~al.(2023)Wu, Irsoy, Lu, Dabravolski, Dredze, Gehrmann, Kambadur, Rosenberg, and Mann]{wu2023bloomberggpt}
Shijie Wu, Ozan Irsoy, Steven Lu, Vadim Dabravolski, Mark Dredze, Sebastian Gehrmann, Prabhanjan Kambadur, David Rosenberg, and Gideon Mann.
\newblock Bloomberggpt: A large language model for finance.
\newblock \emph{arXiv preprint arXiv:2303.17564}, 2023.
\newblock URL \url{https://arxiv.org/abs/2303.17564}.

\bibitem[Xu et~al.(2024)Xu, Huang, Hu, Li, Gilani, Goan, and Liu]{xu2024bishop}
Chenwei Xu, Yu-Chao Huang, Jerry Yao-Chieh Hu, Weijian Li, Ammar Gilani, Hsi-Sheng Goan, and Han Liu.
\newblock Bishop: Bi-directional cellular learning for tabular data with generalized sparse modern hopfield model.
\newblock In \emph{Forty-first International Conference on Machine Learning (ICML)}, 2024.
\newblock URL \url{https://arxiv.org/abs/2404.03830}.

\bibitem[Yuille and Rangarajan(2003)]{cccp}
A.~L. Yuille and Anand Rangarajan.
\newblock {The Concave-Convex Procedure}.
\newblock \emph{Neural Computation}, 15\penalty0 (4):\penalty0 915--936, 04 2003.
\newblock URL \url{https://doi.org/10.1162/08997660360581958}.

\bibitem[Zafrir et~al.(2019)Zafrir, Boudoukh, Izsak, and Wasserblat]{zafrir2019q8bert}
Ofir Zafrir, Guy Boudoukh, Peter Izsak, and Moshe Wasserblat.
\newblock Q8bert: Quantized 8bit bert.
\newblock In \emph{2019 Fifth Workshop on Energy Efficient Machine Learning and Cognitive Computing-NeurIPS Edition (EMC2-NIPS)}, pages 36--39. IEEE, 2019.
\newblock URL \url{https://arxiv.org/abs/1910.06188}.

\bibitem[Zhang et~al.(2022)Zhang, Roller, Goyal, Artetxe, Chen, Chen, Dewan, Diab, Li, Lin, et~al.]{zhang2022opt}
Susan Zhang, Stephen Roller, Naman Goyal, Mikel Artetxe, Moya Chen, Shuohui Chen, Christopher Dewan, Mona Diab, Xian Li, Xi~Victoria Lin, et~al.
\newblock Opt: Open pre-trained transformer language models.
\newblock \emph{arXiv preprint arXiv:2205.01068}, 2022.
\newblock URL \url{https://arxiv.org/abs/2205.01068}.

\bibitem[Zhang(2023)]{zhang2023mathematical}
Tong Zhang.
\newblock \emph{Mathematical analysis of machine learning algorithms}.
\newblock Cambridge University Press, 2023.
\newblock URL \url{https://tongzhang-ml.org/lt-book/lt-book.pdf}.

\bibitem[Zhou et~al.(2021)Zhou, Zhang, Peng, Zhang, Li, Xiong, and Zhang]{zhou2021informer}
Haoyi Zhou, Shanghang Zhang, Jieqi Peng, Shuai Zhang, Jianxin Li, Hui Xiong, and Wancai Zhang.
\newblock Informer: Beyond efficient transformer for long sequence time-series forecasting, 2021.
\newblock URL \url{https://arxiv.org/abs/2012.07436}.

\bibitem[Zhou et~al.(2023)Zhou, Ji, Li, Dutta, Davuluri, and Liu]{zhou2023dnabert}
Zhihan Zhou, Yanrong Ji, Weijian Li, Pratik Dutta, Ramana Davuluri, and Han Liu.
\newblock Dnabert-2: Efficient foundation model and benchmark for multi-species genome.
\newblock \emph{arXiv preprint arXiv:2306.15006}, 2023.
\newblock URL \url{https://arxiv.org/abs/2306.15006}.

\bibitem[Zhou et~al.(2024)Zhou, Wu, Ho, Wang, Shi, Davuluri, Wang, and Liu]{zhou2024dnabert}
Zhihan Zhou, Weimin Wu, Harrison Ho, Jiayi Wang, Lizhen Shi, Ramana~V Davuluri, Zhong Wang, and Han Liu.
\newblock Dnabert-s: Learning species-aware dna embedding with genome foundation models.
\newblock \emph{ArXiv}, 2024.
\newblock URL \url{https://arxiv.org/abs/2402.08777}.

\bibitem[Zhu et~al.(2015)Zhu, Kiros, Zemel, Salakhutdinov, Urtasun, Torralba, and Fidler]{Zhu_2015_ICCV}
Yukun Zhu, Ryan Kiros, Rich Zemel, Ruslan Salakhutdinov, Raquel Urtasun, Antonio Torralba, and Sanja Fidler.
\newblock Aligning books and movies: Towards story-like visual explanations by watching movies and reading books.
\newblock In \emph{The IEEE International Conference on Computer Vision (ICCV)}, December 2015.
\newblock URL \url{https://arxiv.org/abs/1506.06724}.

\end{thebibliography}

\end{document}